\documentclass{article}

\usepackage{microtype}
\usepackage{graphicx}
\usepackage{subfigure}
\usepackage{booktabs} 

\usepackage{algorithm}
\usepackage[noend]{algpseudocode}

\makeatletter
\newenvironment{protocol}[1][htb]{%
  \renewcommand{\ALG@name}{Protocol}
  \begin{algorithm}[#1]%
  }{\end{algorithm}
}
\makeatother

\input{Defs/mathdefs}
\usepackage{amsthm}
\usepackage{amssymb}
\usepackage{thmtools}

\declaretheoremstyle[
	    spaceabove=\topsep, 
	    spacebelow=\topsep, 
	    headfont=\normalfont\bfseries,
	    bodyfont=\normalfont\itshape,
	    notefont=\normalfont\bfseries,
	    notebraces={(}{)},
	    postheadspace=0.33em, 
	    headpunct={.},
    ]{theorem}
\declaretheorem[style=theorem]{theorem}

\declaretheoremstyle[
	    spaceabove=\topsep, 
	    spacebelow=\topsep, 
	    headfont=\normalfont\bfseries,
	    bodyfont=\normalfont,
	    notefont=\normalfont\bfseries,
	    notebraces={(}{)},
	    postheadspace=0.33em, 
	    headpunct={.},
    ]{definition}

\declaretheoremstyle[
        spaceabove=\topsep, 
        spacebelow=\topsep, 
        headfont=\normalfont\bfseries,
        bodyfont=\normalfont,
        notefont=\normalfont\bfseries,
        notebraces={}{},
        postheadspace=0.33em, 
        qed=$\blacksquare$, 
        headpunct={.},
    ]{proofstyle}
\declaretheorem[style=proofstyle,numbered=no,name=Proof]{proof}

\declaretheorem[style=theorem,sibling=theorem,name=Lemma]{lemma}

\declaretheorem[style=theorem,numbered=no,name=Theorem]{theorem*}
\declaretheorem[style=theorem,numbered=no,name=Lemma]{lemma*}
\declaretheorem[style=theorem,numbered=no,name=Corollary]{corollary*}
\declaretheorem[style=theorem,numbered=no,name=Proposition]{proposition*}
\declaretheorem[style=theorem,numbered=no,name=Claim]{claim*}
\declaretheorem[style=theorem,numbered=no,name=Fact]{fact*}
\declaretheorem[style=theorem,numbered=no,name=Observation]{observation*}
\declaretheorem[style=theorem,numbered=no,name=Conjecture]{conjecture*}

\declaretheorem[style=definition,numbered=no,name=Definition]{definition*}
\declaretheorem[style=definition,numbered=no,name=Remark]{remark*}
\declaretheorem[style=definition,numbered=no,name=Example]{example*}
\declaretheorem[style=definition,numbered=no,name=Question]{question*}

\newcommand{\piClass}{\Pi_M}

\newcommand{\Egood}{E_g}
\newcommand{\EgoodBar}{\bar{E}_g}

\newcommand{\numStates}{\abs{\X}}
\newcommand{\setEpochs}{[E]}
\newcommand{\numEpochs}{E}
\newcommand{\setEpochsE}{K_e}
\newcommand{\kEpoch}{k_e}
\newcommand{\kPrevEpoch}{k_{e-1}}

\newcommand{\covK}[1][k]{{\Lambda^{#1}}}

\newcommand{\covKH}[1][h]{{\Lambda^{k}_{#1}}}

\newcommand{\covWarmupH}[1][h]{{\Lambda^{0}_{#1}}}
\newcommand{\hatCovK}[1][k]{{\covK[\kEpoch]}}
\newcommand{\hatCovKH}[1][h]{{\Lambda^{\kEpoch}_{#1}}}

\newcommand{\covZ}[1][\kEpoch]{{\Sigma_\zeta^{#1}}}

\newcommand{\betaR}{\beta_r}
\newcommand{\betaP}{\beta_p}
\newcommand{\betaPh}[1][h]{\beta_{p,#1}}
\newcommand{\betaPtag}{\beta_{\hat{p}}}
\newcommand{\betaZ}{\beta_\zeta}
\newcommand{\betaRZ}{\beta_{r\zeta}}
\newcommand{\betaQ}{\beta_Q}
\newcommand{\betaQh}[1][h]{\beta_{Q,#1}}
\newcommand{\betaWarmup}{\beta_w}
\newcommand{\epsCov}{\epsilon_{\text{cov}}}
\newcommand{\firstEpisodeAfterWarmup}{k_0}

\newcommand{\M}{\mathcal{M}}
\newcommand{\A}{\mathcal{A}}
\newcommand{\X}{\mathcal{X}}
\newcommand{\Z}{\mathcal{Z}}

\newcommand{\Q}{\mathcal{Q}}
\newcommand{\V}{\mathcal{V}}
\newcommand{\D}{\mathcal{D}}
\newcommand{\Dh}[1][h]{\D_{#1}}
\newcommand{\lambdaR}{\lambda_r}
\newcommand{\lambdaP}{\lambda_p}

\newcommand{\etaO}{\eta_o}
\newcommand{\etaH}{\eta_x}

\newcommand{\psiKH}[1][h]{\widehat{\psi}_{#1}^{k}}

\newcommand{\psiH}[1][h]{\psi_{#1}}
\newcommand{\hatJE}[1][e]{\hat{j}_{#1}}

\newcommand{\thetaH}[1][h]{\theta_{#1}}
\newcommand{\thetaK}[1][k]{{\widehat{\theta}^{#1}}}
\newcommand{\thetaKH}[1][h]{{\widehat{\theta}^{k}_{#1}}}

\newcommand{\z}{\zeta}
\newcommand{\zh}[1][h]{\zeta_{#1}}
\newcommand{\zr}{\zeta_r}
\newcommand{\zrh}[1][h]{\zeta_{r,#1}}
\newcommand{\zp}{\zeta_p}
\newcommand{\zph}[1][h]{\zeta_{p,#1}}

\newcommand{\piK}[1][]{\pi^k_{#1}}
\newcommand{\piOpt}[1][]{\pi^\star_{#1}}

\newcommand{\rh}[1][h]{r_{#1}}

\newcommand{\bh}[1][h]{b_{#1}}

\newcommand{\regret}{\mathrm{Regret}}

\newenvironment{ack}{\subsection*{Acknowledgements}}

\usepackage{hyperref}

\usepackage[accepted]{icml2024}

\usepackage{amsmath}
\usepackage{amssymb}
\usepackage{mathtools}
\usepackage{amsthm}

\usepackage[capitalize]{cleveref}

\usepackage[disable,textsize=tiny]{todonotes}

\icmltitlerunning{Near-Optimal Regret in Linear MDPs with Aggregate Bandit Feedback}

\begin{document}

\twocolumn[
\icmltitle{Near-Optimal Regret in Linear MDPs with Aggregate Bandit Feedback}



\icmlsetsymbol{equal}{*}
\begin{icmlauthorlist}
\icmlauthor{Asaf Cassel}{equal,tau}
\icmlauthor{Haipeng Luo}{usc,amzn}
\icmlauthor{Aviv Rosenberg}{equal,google}
\icmlauthor{Dmitry Sotnikov}{amzn}
\end{icmlauthorlist}

\icmlaffiliation{tau}{Blavatnik School of Computer Science, Tel Aviv University}
\icmlaffiliation{usc}{University of Southern California}
\icmlaffiliation{google}{Google Research}
\icmlaffiliation{amzn}{Amazon Science}

\icmlcorrespondingauthor{Asaf Cassel}{acassel@mail.tau.ac.il}

\icmlkeywords{reinforcement learning, aggregate feedback, linear MDP, policy optimization}

\vskip 0.3in
]



\printAffiliationsAndNotice{\textsuperscript{*}Work partially done while at Amazon Science.} 


\begin{abstract}
In many real-world applications, it is hard to provide a reward signal in each step of a Reinforcement Learning (RL) process and more natural to give feedback when an episode ends.
To this end, we study the recently proposed model of RL with Aggregate Bandit Feedback (RL-ABF), where the agent only observes the sum of rewards at the end of an episode instead of each reward individually.
Prior work studied RL-ABF only in tabular settings, where the number of states is assumed to be small.
In this paper, we extend ABF to linear function approximation and develop two efficient algorithms with near-optimal regret guarantees: a value-based optimistic algorithm built on a new randomization technique with a Q-functions ensemble, and a policy optimization algorithm that uses a novel hedging scheme over the ensemble.
\end{abstract}

\section{Introduction}

Reinforcement Learning (RL) has demonstrated remarkable empirical success in recent years \cite{mnih2015human,haarnoja2018soft,ouyang2022training}, leading to increasing interest in understanding its theoretical guarantees. 
The standard model for RL is the Markov Decision Process (MDP) in which an agent interacts with an environment over multiple steps.
In every step, the agent takes an action based on its observation of the current state, and immediately gets a reward feedback before transitioning to the next state.
However, in many applications, it is either hard to provide accurate reward or the reward naturally arrives only at the end of an episode, e.g.,  in robotics \cite{jain2013learning} or Large Language Models (LLMs; \citet{stiennon2020learning}).

To address this issue, several works \cite{Efroni_Merlis_Mannor_2021,pmlr-v134-cohen21a,chatterji2021theory} studied the Aggregate Bandit Feedback (ABF) model in which the agent only observes the sum of rewards at the end of an episode as feedback, instead of each reward individually. 
Yet, they all focused on tabular MDPs where the number of states is assumed to be finite and small.
Although ABF is challenging even in the tabular setting, in all practical applications the state space is huge and RL algorithms use function approximation to approximate the value function efficiently.

In this paper, we tackle aggregate feedback in the presence of function approximation.
Specifically, we consider ABF in Linear MDPs \cite{jin2020provably}, developing two efficient algorithms with near-optimal regret guarantees, i.e., $\tilde{O}( \text{poly}(d H) \sqrt{K})$ regret where $K$ is the number of episodes, $d$ is the dimension of the linear MDP, and $H$ is the horizon.
Our first algorithm, called Randomized Ensemble Least Squares Value Iteration (RE-LSVI), is an optimistic LSVI algorithm where the optimism is achieved via a new randomization technique.
This algorithm is simple and gives our tightest regret bound.
Our second algorithm, called Randomized Ensemble Policy Optimization (REPO), is based on policy optimization (PO)  methods that have become extremely popular in recent years \cite{schulman2015trust,schulman2017proximal} and especially in training LLMs \cite{ouyang2022training}.
This is the first PO algorithm for aggregate feedback, which does not exist even in the tabular MDP setting.
Theoretically, ABF is substantially more challenging under function approximation since it can no longer be formulated as a linear bandits problem \cite{Efroni_Merlis_Mannor_2021} (doing so would lead to a regret bound that scales with the number of states).

Our main technical contribution is the \emph{ensemble randomization} technique.
We use independently sampled Gaussian noise to compute multiple Q-functions from the same data, allowing for an optimistic estimate of the optimal value with controlled variance.
In addition, we introduce a \emph{loose truncation} mechanism to keep our estimates bounded without creating biases in the random walks induced by the added noise.
While \citet{Efroni_Merlis_Mannor_2021} already leveraged randomization to solve ABF in the tabular setting, they use a complex Thompson Sampling (TS) based analysis \cite{agrawal2013thompson}.
In contrast, our novel use of the ensemble unlocks a very simple regret analysis based on optimism.
A concurrent work \cite{wu2023making} uses randomization to solve Preference-based RL in linear MDPs, which is closely related to ABF \cite{chen2022human}.
Applying their algorithm in our setting again requires a complex TS analysis and yields regret with a worse dependence on $d$ and $H$.

Another major contribution is a novel \emph{hedging scheme} on top of the Q-functions ensemble, which enables us to obtain the first PO algorithm for ABF.
It also demonstrates the effectiveness of our ensemble technique even for algorithms that are not based on optimism, as PO algorithms are notoriously difficult to analyze, albeit highly successful in practice.
In fact, only very recently it was proven that PO achieves $\sqrt{K}$ regret in linear MDPs with standard reward feedback \cite{sherman2023rate}.
While our PO algorithm in linear MDPs involves a warm-up stage that uniformly explores the state space (similarly to \citet{sherman2023rate}), in the tabular setting we develop an alternative analysis that avoids the need to perform this warm-up and may be of independent interest.

\paragraph{Related Work.}
There is a rich literature on regret minimization in RL, where the most popular algorithmic methods are optimism (achieved via exploration bonuses, e.g.,~\citet{jaksch2010near,azar2017minimax,jin2020provably}), policy optimization (e.g.,~\citet{cai2020provably,shani2020optimistic,luo2021policy}), and global optimization methods based on either regularization (e.g.,~\citet{zimin2013online,rosenberg2019bandit,rosenberg2019online, jin2020learning}) or optimism (e.g.,~\citet{zanette2020learning}).
They all rely on the standard assumption that each individual reward within an episode is revealed.

\citet{Efroni_Merlis_Mannor_2021} introduced RL-ABF and gave the first regret bounds in tabular MDPs.
\citet{chatterji2021theory} then generalized aggregate feedback to a more realistic setting where only certain binary feedback is available, but their algorithm is efficient only under additional assumptions.
\citet{pmlr-v134-cohen21a} studied ABF in tabular adversarial environments, but their algorithm is built on a global mirror descent approach that cannot be extended to function approximation.
Unlike all previous work (including this one) that focus on ABF in the context of online RL and exploration, \citet{xu2022provably} recently studied ABF in offline RL which poses different challenges.

A closely related model is Preference-based RL (PbRL; \citet{wirth2017survey}), where the agent receives feedback only in terms of preferences over a trajectory pair instead of absolute rewards.
\citet{saha2023dueling,chen2022human} achieved sub-linear regret for PbRL, but their algorithms are computationally intractable even in tabular MDPs.
As mentioned earlier, in a concurrent work, \citet{wu2023making} utilize randomization to solve PbRL in linear MDPs efficiently.
Additional works on PbRL study sample complexity and offline RL, and use the term RL from Human Feedback (RLHF; \citet{wang2023rlhf,zhan2023provable,zhan2023query}).

A different line of work study delayed reward feedback in RL \cite{howson2023optimism,jin2022near,lancewicki2020learning,lancewicki2023learning}.
While they also deal with reward feedback that is not provided immediately after taking an action, they still observe each reward individually, thus avoiding the need to perform accurate credit assignment that our ABF model requires. 

It is also worth mentioning that recently \citet{tiapkin2023modelfree} utilized an ensemble of Q-functions for regret minimization in tabular MDPs with standard reward feedback.
While we perturb the reward in a manner that is applicable to both value iteration and policy optimization, they perturb the learning rates in Q-learning.

\section{Problem Setup}

\paragraph{Markov Decision Process (MDP).}
A finite horizon MDP $\M$ is defined by a tuple $(\X, \A, x_1, r, P, H)$ with $\X$, a set of states, $\A$, a set of actions, $H$, decision horizon, $x_1 \in \X$, an initial state (which is assumed to be fixed for simplicity), $r = (r_h)_{h \in [H]}, r_h: \X \times \A \to [0,1]$, a horizon dependent immediate reward function for taking action $a$ at state $x$ and horizon $h$, and $P = (P_h)_{h \in [H]}, P_h: \X \times \A \to \Delta(\X)$, the transition probabilities. A single episode of an MDP is a sequence $(x_h, a_h, r_h)_{h \in [H]} \in (\X \times \A \times [0,1])^H$ such that
\begin{align*}
    \Pr[x_{h+1} = x' \mid x_h = x, a_h = a] = P_h(x' \mid x,a)
    ,
\end{align*}
and $\rh \in [0,1]$ is sampled such that $\EE\brk[s]*{\rh \mid x_h, a_h} = \rh(x_h,a_h)$. 
Notice the overloaded notation for $\rh$ where $\rh(\cdot, \cdot)$ refers to the immediate (mean) reward function, and $\rh$ (without inputs) refers to a sampled reward at horizon $h$.

\paragraph{Linear MDP.}
A linear MDP \cite{jin2020provably} satisfies all the properties of the above MDP but has the following additional structural assumptions. There is a known feature mapping $\phi : \mathcal{X} \times A \to \RR[d]$ such that
\begin{align*}
    r_h(x,a) = \phi(x,a)\tran \thetaH
    ,
    \quad
    P_h(x' \mid x,a) = \phi(x,a)\tran \psiH(x')
    ,
\end{align*}
where $\thetaH \in \RR[d]$, $\theta = (\thetaH[1]\tran, \ldots , \thetaH[H]\tran)\tran \in \RR[d H]$, and $\psiH : \mathcal{X} \to \RR[d]$ are unknown parameters.
We make the following normalization assumptions, common throughout the literature:
\begin{enumerate}
    \item $\norm{\phi(x,a)} \le 1$ for all $x \in X, a \in \A$;
    \item $\norm{\thetaH} \le \sqrt{d}$ for all $h \in [H]$;
    \item $\norm{\abs{\psiH}(\X)} = \norm{\sum_{x \in \X} \abs{\psiH(x)}} \le \sqrt{d}$ for all $h \in [H]$;
\end{enumerate}
where $\abs{\psiH(x)}$ is the entry-wise absolute value of $\psiH(x) \in \RR[d]$.
We follow the standard assumption in the literature that the action space $\A$ is finite.
In addition, for ease of mathematical exposition, we also assume that the state space $\X$ is finite.
This allows for simple matrix notation and avoids technical measure theoretic definitions.
Importantly, our results are completely independent of the state space size $\abs{\X}$, both computationally and in terms of regret.
Thus, there is no particular loss of generality.

\paragraph{Policy and Value.} A stochastic Markov policy $\pi = (\pi_h)_{h \in [H]} : [H] \times \X \mapsto \Delta(\A)$ is a mapping from a step and a state to a distribution over actions. Such a policy induces a distribution over trajectories $\iota = (x_h,a_h)_{h \in [H]}$, i.e., sequences of $H$ state-action pairs. For $f : (\X \times \A)^H \to \RR$, which maps trajectories to real values, we denote the expectation with respect to $\iota$ under dynamics $P$ and policy $\pi$ as $\EE[P, \pi] \brk[s]{f(\iota)}$. Similarly, we denote the probability under this distribution by $\PR[P, \pi] \brk[s]{\cdot}$. We denote the class of stochastic Markov policies as $\piClass$.
For each policy $\pi$ and horizon $h \in [H]$ we define its reward-to-go as $V_h^{\pi}(x) = \EE[P,\pi][\sum_{h'=h}^{H} r_{h'}(x_{h'},a_{h'}) \mid x_h = x]$, which is the expected reward if one starts from state $x$ at horizon $h$ and follows policy $\pi$ onwards.
 The performance of a policy, also called its value, is measured by its expected cumulative reward, given by $V_1^{\pi}(x_1)$.
When clear from context, we omit the $1$ and simply write $V^{\pi}(x_1)$. The optimal policy is thus given by $\piOpt \in \argmax_{\pi \in \piClass} V^{\pi}(x_1)$, known to be optimal even among the class of stochastic history-dependent policies. Finally, we denote the optimal value as $V^\star = V_1^{\piOpt}$.

\paragraph{Aggregate Feedback and Regret.}
We consider a standard episodic regret minimization setting where an algorithm performs $K$ interactions with an MDP $\M$, but limit ourselves to \emph{aggregate feedback} reward observations.
Concretely, at the start of each interaction/episode $k \in [K]$, the agent specifies a stochastic Markov policy $\piK = (\piK[h])_{h \in [H]}$. Subsequently, it observes the trajectory $\iota^k$ sampled from the distribution $\PR[P, \piK]$, and the cumulative episode reward $v^k$ (see Protocol~\ref{alg:protocol}).
This is unlike standard bandit feedback where the individual rewards $(\rh^k)_{h \in [H]}$ are observed.
%
\begin{protocol}[!ht]
	\caption{Aggregate Feedback Interaction Protocol} \label{alg:protocol}
	\begin{algorithmic}[1]
	    \For{episode $k = 1,2,\ldots, K$}
                \State Agent chooses policy $\piK$.
                \State Observes trajectory $\iota^k = (x_h^k,a_h^k)_{h \in [H]}$.
                \State Observes episode reward $v^k = \sum_{h \in [H]} \rh^k.$
	        
	    \EndFor
	\end{algorithmic}
\end{protocol}

We measure the quality of any algorithm via its \emph{regret} -- the difference between the value of the policies $\piK$ generated by the algorithm and that of the optimal policy $\piOpt$, i.e.,
\begin{align*}
    \regret
    =
    \sum_{k \in [K]} V^\star(x_1) - V^{\piK}(x_1)
    .
\end{align*}

\section{Algorithms and Main Results}
We present two algorithms for regret minimization in aggregate feedback linear MDPs. 
The first, RE-LSVI (\cref{alg:RE-LSVI}), is a value-based optimistic algorithm, and the second, REPO (\cref{alg:r-opo-for-linear-mdp}), is a policy optimization routine. 
Both algorithms achieve regret with the optimal $\sqrt{K}$ dependence on the number of episodes, but RE-LSVI has a favorable dependence on $H$.

\paragraph{Notation.} 
Throughout the paper
$
    \phi_h^k
    =
    \phi(x_h^k, a_h^k) \in \RR[d]
$
denotes the state-action features at horizon $h$ of episode $k$, and
$
    \phi^{k} 
    =
    \brk{{\phi_1^{k}}\tran, \ldots, {\phi_H^{k}}\tran}\tran
    \in \RR[dH]
$
is their concatenation. 
Following a similar convention, for any vector $\zeta \in \RR[dH]$, let $\brk[c]{\zeta_h}_{h \in [H]} \in \RR[d]$ denote its $H$ equally sized sub-vectors, i.e.,
$
    (\zeta_1\tran, \ldots, \zeta_H\tran)\tran 
    =
    \zeta
    .
$
Notice that the same does not hold for matrices, e.g., $\covKH$ is not a sub-matrix of $\covK$.
In addition, $\norm{v}_A = \sqrt{v\tran A v}$, and clipping operator $\clip_\beta[z]$ for some $\beta > 0$ is defined as $\min\brk[c]{\beta, \max\brk[c]{-\beta, z}}$.
Hyper-parameters follow the notations $\beta_z$, $\lambda_z$, and $\eta_z$ for some $z$ and $\delta \in (0,1)$ denotes a confidence parameter. Finally, in the context of an algorithm, $\gets$ signs refer to compute operations whereas $=$ signs define operators, which are evaluated at specific points as part of compute operations.

\subsection{Randomized Ensemble Least Squares Value Iteration (RE-LSVI)}

At the start of episode $k$, similarly to algorithms for standard reward feedback (e.g., \citet{jin2020provably}), RE-LSVI defines a dynamics backup operator $\psiKH$ for each $h \in [H]$, which, when given a function $V: \X \mapsto \RR$, estimates the vector $\psiH V = \sum_{x\in\X} \psiH(x)V(x)$ using least squares as follows:
\begin{equation}\label{eq:LS-estimate-psi}
\psiKH V = (\covKH)^{-1} \sum_{\tau \in \Dh^k} \phi_h^\tau V(x_{h+1}^\tau), 
\end{equation}
where $\covKH =\lambdaP I + \sum_{\tau \in \Dh^k} \phi_h^\tau (\phi_h^\tau)\tran$ for some parameter $\lambdaP > 0$ is the covariance matrix for horizon $h$, and $\Dh^k$ is a dataset. 
On the other hand, for reward estimates, while standard algorithms compute a least square estimator for each $\thetaH$ using the observed rewards at horizon $h$, this is no longer feasible in our ABF model.
Instead, given the aggregate reward $v^\tau = \sum_{h \in [H]} r_h^\tau$ for $\tau < k$, it is natural to directly estimate the aggregate reward vector $\theta$ using least squares:
\begin{equation}\label{eq:LS-estimate-theta}
   \thetaK =  (\covK)^{-1} \sum_{\tau \in \D^k} \phi^\tau v^\tau,
\end{equation}
where $\covK = \lambdaR I + \sum_{\tau \in \D^k} \phi^\tau (\phi^\tau)\tran$ for some parameter $\lambdaR > 0$ is the aggregate covariance matrix, and $\D^k$ is a dataset.
The corresponding $H$ sub-vectors $\{\thetaKH\}_{h\in [H]}$ then estimate $\{\thetaH\}_{h\in [H]}$.
For RE-LSVI, $\D^k$ and $\D^k_h$ are both simply $\{1, \ldots, k-1\}$, that is, all previous data (but they will be different for our next algorithm).

To encourage exploration, we use the standard exploration bonuses $\bh^k(x,a) = \betaP \norm{\phi(x,a)}_{(\covKH)^{-1}}$ for some parameter $\betaP > 0$ to handle uncertainty in the dynamics.
To handle uncertainty in the rewards, however, it is crucial to deploy our proposed ensemble randomization technique (reasoning deferred to the end of this subsection): we draw $m \approx \log (K/\delta)$ independent Gaussian noise vectors $\z^{k,i} \sim \mathcal{N}(0, \betaR^2 (\covK)^{-1})$ for $i\in [m]$ whose covariance is attuned to that of the aggregate reward least squares estimate (with some parameter $\betaR>0$). 
Next, we calculate $m$ value estimates $\{\hat{V}_h^{k,i}\}_{i\in[m]}$ using standard LSVI, each with a different perturbed reward vector $\thetaK + \z^{k,i}$ (\Cref{line:LSVI} in \Cref{alg:RE-LSVI}).
Finally, we find the index $i_k \in\argmax_{i \in [m]} \hat{V}_1^{{k,i}}(x_1)$ with the maximal initial value and follow the greedy policy with respect to the $i_k$-th Q-function.

One final important tweak we make to LSVI is the clipping of the value estimates. Specifically, standard algorithms truncate each $\hat{Q}_h^{k,i}(x,a)$ to $[0,H]$ (the range of the true Q-function), but we propose a looser clip that truncates these Q-estimates to $[-(H + 1 - h)\betaRZ, (H + 1 - h)\betaRZ]$ for some parameter $\betaRZ > 0$ such that $(H + 1 - h)\betaRZ \approx \sqrt{d} H\betaR$.
The symmetry in our clipping is to avoid biases in the random walk induced by $\z^{k,i}$, and the looser threshold is crucial for the analysis (more discussion to follow).
\begin{algorithm}[!ht]
	\caption{RE-LSVI with aggregate feedback} \label{alg:RE-LSVI}
	\begin{algorithmic}[1]
		\State \textbf{input}:
		$\delta, \lambdaR, \lambdaP, \betaR, \betaP, \betaRZ > 0 ; m \ge 1$.

	    \For{episode $k = 1,2,\ldots, K$}

                \State Define $\D^k = \Dh^k = \brk[c]{1, \ldots, k-1}$ for all $h \in [H].$
                
                \State Compute $\thetaK$ and define $\psiKH$ (\cref{eq:LS-estimate-theta,eq:LS-estimate-psi}).
         


	        
	       \State Sample
                $
                \z^{k,i} \sim \mathcal{N}(0, \betaR^2 (\covK)^{-1})
                $
                for all $i \in [m].$

                
                
                \State Define $\hat{V}_{H+1}^{k,i}(x) = 0$ for all $i\in[m], x \in \mathcal{X}$. 
                


                \State For every $i \in [m]$ and $h = H, \ldots, 1$:
                \begin{alignat*}{2}
                        &
                        w_h^{k, i}
                        &&
                        \gets
                        \thetaKH + \zh^{k,i}
                        +
                        \psiKH \hat{V}^{k,i}_{h+1},
                        \\
                        &
                        \hat{Q}_h^{k,i}(x,a)
                        &&
                        =
                        \clip_{(H + 1 - h)\betaRZ}\brk[s]*{
                            \phi(x,a)\tran w_h^{k,i}
                            +
                            \bh^k(x,a) 
                            },
                        \\
                        &
                        \hat{V}_h^{k,i}(x)
                        &&
                        =
                        \max_{a \in \mathcal{A}} \hat{Q}_h^{k,i}(x,a)
                        \\
                        &
                        \pi^{k,i}_h(x)
                        &&
                        \in
                        \argmax_{a \in \A} \hat{Q}^{k,i}_h(x,a)
                        .
                    \end{alignat*}
                    \label{line:LSVI}                
	       \State 
            \label{line:LSVI-ensemble}
            $i_k \gets \argmax_{i \in [m]} \hat{V}_1^{{k,i}}(x_1)$ and play $\piK = \pi^{k, i_k}.$
	       
	        
	       \State Observe episode reward $v^k$ and trajectory $\iota^k$.
        
            
	       
	    \EndFor
	\end{algorithmic}
\end{algorithm}

Most of the operations in RE-LSVI are similar to algorithms for standard reward feedback (e.g., \citet{jin2020provably}).
The only difference lays in drawing Gaussian noise vectors and calculating $m$ value functions instead of $1$.
Thus, the computational complexity is similar to previous work up to a logarithmic $\log(K/\delta)$ factor. 
The following is our main result for \cref{alg:RE-LSVI}.
\begin{theorem}
\label{thm:regret-RE-LSVI}
    Suppose that we run RE-LSVI (\cref{alg:RE-LSVI}) with the parameters defined in \cref{lemma:good-event} (in \cref{appendix-sec:RE-LSVI}).
    Then with probability at least $1 - \delta$, we have
    \begin{align*}
        \regret
        =
        O\brk*{\sqrt{ d^5 H^7 K \log^6 (dHK / \delta)}}
        .
    \end{align*}
\end{theorem}

\paragraph{Discussion.}
\cref{alg:RE-LSVI} contains elements of both LSVI-UCB \cite{jin2020provably} and UCBVI-TS \cite{Efroni_Merlis_Mannor_2021}, but also new ideas, necessary for combining them.
When constructing an optimistic planning procedure, \citet{Efroni_Merlis_Mannor_2021} noticed that the reward bonuses have to be trajectory-dependent. This breaks the MDP structure and makes efficient planning impossible. Overcoming this, they suggest drawing a random bonus whose covariance scales with the uncertainty of the aggregate feedback value estimation. They then follow a standard planning procedure over the empirical MDP.
Unfortunately, directly planning in the empirical linear MDP seems to be a difficult task since value backups are calculated via a least squares argument, which corresponds to a potentially non-valid transition kernel, i.e., one with negative entries and whose sum may exceed $1$.
Consequently, one has to contend either with value backups blowing up exponentially with $H$, or with an empirical optimal policy that is not greedy with respect to its Q-function.

This is typically overcome by truncating the value to $[0,H]$ before using it in the backup step of the dynamic program (see e.g., \citep{jin2020provably}). The resulting policy is not optimal for the empirical Linear MDP but can be shown to be optimistic with respect to the true optimal policy.
Unfortunately, this truncation introduces a bias to the aggregate reward estimate, breaking the correlation between the uncertainty of the individual estimates at each horizon $h \in [H]$.

We overcome this bias by introducing a loose truncation mechanism that clips the values of $\hat{Q}_h^{k,i} (h \in [H])$ to $[-(H+1-h)\betaRZ, (H+1-h)\betaRZ]$ where $\betaRZ$ is a high probability bound on the immediate perturbed reward estimates. This does not immediately solve the issue since the dynamics are still invalid, but through a careful (high probability) analysis, we show that one can bound the error by the unclipped, thus unbiased, process (see the analysis in \cref{sec:analysis:LSVI} from \cref{eq:analysis-lsvi-bounded-reward} to \cref{eq:optimism-term-i*}).

A second important feature of our algorithm is the randomized ensemble. In \citet{Efroni_Merlis_Mannor_2021} the equivalent noise term is sampled only once and a single value is calculated. Similar to other Thompson Sampling (TS) methods, this results in an estimator that is optimistic only with constant probability, thus requiring a careful and intricate analysis to handle the instances where it is not optimistic. In contrast, we draw $m \approx \log(K/\delta)$ noise terms and calculate their respective value functions. Each value estimate is thus optimistic with constant probability, and since the noise terms are i.i.d, it is straightforward to see that at least one of them is optimistic with high probability. We show that following the policy related to the maximal value is thus also optimistic with high ($\ge 1 - \delta$) probability. This yields a clear and easy-to-follow optimism-based analysis (see \cref{sec:analysis:LSVI}).
Moreover, in what follows we show how to extend the randomized ensemble idea to obtain a policy optimization-based algorithm for linear MDPs with ABF.

\subsection{\mbox{Randomized Ensemble Policy Optimization (REPO)}}
We now introduce our second algorithm REPO that is based on the more popular policy optimization scheme.
REPO starts with a reward-free warm-up routine by \citet{sherman2023rate}, which is in turn based on \citet{wagenmaker2022reward}, to uniformly explore the state space. 
It lasts for $\firstEpisodeAfterWarmup \approx \sqrt{K}$ episodes and outputs exploratory datasets $\brk[c]{\Dh^0}_{h \in [H]}$ that are used to initialize the least squares estimators for the value iteration backups (\cref{eq:LS-estimate-psi}).
In addition, for each horizon $h \in [H]$, it defines a set of ``known'' states:
\begin{align}
\label{eq:Zh}
    \Z_h
    =
    \brk[c]*{x \in \X \mid \max_{a \in \A} \norm{\phi(x,a)}_{(\covWarmupH)^{-1}} \le \frac{1}{2 \betaWarmup H}}
    ,
\end{align}
where $\covWarmupH = \lambdaP I + \sum_{\tau \in \Dh^0}\phi_h^\tau (\phi_h^\tau)\tran$ and $\betaWarmup>0$ is some parameter.
The algorithm then proceeds in epochs, each one ending when the uncertainty (of either the dynamics or aggregate reward) shrinks by a multiplicative factor, as expressed by the determinant of their covariance (\cref{line:repo-epoch-condition} in \cref{alg:r-opo-for-linear-mdp}).
At the start of each epoch $e$, we draw $m \approx \log(K/\delta)$ reward perturbation vectors $\z^{\kEpoch,i}, i \in [m]$, where $\kEpoch$ denotes the first episode in the epoch. We also initialize $m$ PO sub-routines and a Multiplicative Weights (MW) sub-routine that arbitrates over the PO copies.
Inside an epoch, at the start of each episode $k$, we compute the aggregate reward vector $\thetaK$ (\cref{eq:LS-estimate-theta}) and estimated dynamics backup operators $\psiKH$ (\cref{eq:LS-estimate-psi}), and run the $m$ copies of PO, each differing only by their reward perturbation (see \cref{line:REPO} in \cref{alg:r-opo-for-linear-mdp}). This involves running online mirror descent (OMD) updates over the estimated Q values. Finally, a policy is chosen by hedging over the values of each PO copy using MW (\cref{line:REPO-Hedge} in \cref{alg:r-opo-for-linear-mdp}). 

Notice that, we do not use direct reward bonuses ($b_h^k$ in \cref{alg:RE-LSVI}) to encourage exploration. Instead, we augment the covariance of the reward perturbations with an additional term that accounts for the dynamics uncertainty (\cref{line:REPO-covZ} in \cref{alg:r-opo-for-linear-mdp}). 
Additionally, we replace the clipping mechanism in the value backups with an indicator that zeroes out the Q estimates outside the ``known'' states sets $\Z_h, h \in [H]$. Similarly to \citet{sherman2023rate}, this is crucial as PO performance scales with the complexity (log covering number) of the policy class, which scales with $K$ under the clipping mechanism in \cref{alg:RE-LSVI} (more discussion to follow).

\begin{algorithm}[t]
	\caption{REPO with aggregate feedback} 
 \label{alg:r-opo-for-linear-mdp}
	\begin{algorithmic}[1]
	    \State \textbf{input}:
		$\delta, \epsCov, \betaWarmup, \etaO, \etaH, \lambdaR, \lambdaP, \betaR, \betaP > 0 ; m \ge 1$.

            \State Run reward-free warm-up \citep[Algorithm 2]{sherman2023rate} with $(\frac{\delta}{7}, \betaWarmup, \epsCov)$ to get index sets $\brk[c]{\Dh^0}_{h \in [H]}$, ``known'' states sets $\brk[c]{\Z_h}_{h \in [H]}$ (\cref{eq:Zh}), and let $\firstEpisodeAfterWarmup$ be the first episode after the warm-up.



            \State \textbf{initialize}: $e \gets -1$.
        
	    \For{episode $k = \firstEpisodeAfterWarmup,\ldots, K$}
     
                \If{\mbox{$k = \firstEpisodeAfterWarmup$
                    \textbf{ or }
                    $\exists h \in [H], \  \det\brk{\covKH} \ge 2 \det\brk{\hatCovKH}$}
                    \mbox{$\qquad \qquad \qquad \ $ \textbf{ or }
                    $ \det\brk{\covK} \ge 2 \det\brk{\hatCovK}$}
                }
                \label{line:repo-epoch-condition}

                    \State $e \gets e + 1$ and $\kEpoch \gets k$.

                    \State 
                    \label{line:REPO-covZ}
                    $
                        {\covZ
                        \gets
                        2\betaR^2 {(\hatCovK)}^{-1}
                        +
                        2 H \betaP^2 \mathrm{diag}\brk*{\hatCovKH[1], \ldots, \hatCovKH[H]}^{-1}}
                    $

                    \State Sample $\z^{\kEpoch,i} 
                    \sim 
                    \mathcal{N}\brk*{0, \covZ}$ for all $i \in [m]$.

                    \State Reset $p^k(i) \gets 1/m , \pi^{k,i}_h(a \mid x) \gets 1/\abs{\A}$.
                
                \EndIf

            \State Sample $i_k$ according to $p^k(\cdot)$ and  play $\piK = \pi^{k,i_k}$.

            \State Observe episode reward $v^k$ and trajectory $\iota^k$.

            
            

            \State Define $\D^k = \brk[c]{k_0, \ldots, k-1}$ and $\Dh^k = \Dh^0 \cup \D^k.$
            
            \State Compute $\thetaK$ and define $\psiKH$ (\cref{eq:LS-estimate-theta,eq:LS-estimate-psi}).
            
            \State Define $\hat{V}_{H+1}^{k,i}(x) = 0$ for all $i\in[m], x \in \mathcal{X}$.
            
            \State For every $i \in [m]$ and $h = H, \ldots, 1$:
            %
            \begin{alignat*}{2}
                    &
                    \qquad
                    w_h^{k,i}
                    &&
                    \gets
                    \thetaKH + \zh^{\kEpoch,i}
                    +
                    \psiKH \hat{V}^{k,i}_{h+1}
                    \\  
                    &
                    \qquad
                    \hat{Q}_h^{k,i}(x,a)
                    &&
                    =
                    \phi(x,a)\tran w_h^{k,i}
                    \cdot \indEvent{ x \in \Z_h }
                    \\
                    &
                    \qquad
                    \hat{V}_h^{k,i}(x)
                    &&
                    =
                    \sum_{a \in \mathcal{A}} \pi^{k,i}_h(a \mid x) \hat{Q}_h^{k,i}(x,a)
                    \\
                    &
                    \qquad
                    \pi^{k+1,i}_h(a \mid x)
                    &&
                    \propto
                    \pi^{k,i}_h(a \mid x) \exp (\etaO \hat{Q}_h^{k,i}(x,a))
                    .
                \end{alignat*}
            \label{line:REPO}
	    

           \State
           \label{line:REPO-Hedge}
           $p^{k+1}(i) \propto p^{k}(i) \exp(\etaH \hat{V}_1^{{k,i}}(x_1))$ for all $i \in [m]$.

	    \EndFor
	\end{algorithmic}
	
\end{algorithm}
We note that the computational complexity of \cref{alg:r-opo-for-linear-mdp} is comparable to RE-LSVI (\cref{alg:RE-LSVI}).
%
The following is our main result for \cref{alg:r-opo-for-linear-mdp} (for the full analysis see \cref{appendix-sec:REPO}).
\begin{theorem}
\label{thm:REPO-regret}
    Suppose that we run REPO (\cref{alg:r-opo-for-linear-mdp}) with the parameters defined in \cref{thm:regret-bound-PO-linear} (in \cref{appendix-sec:REPO}).
    Then with probability at least $1 - \delta$, we have
    \begin{align*}
        \regret
        =
        O\brk*{\sqrt{ d^5 H^9 K \log^9 (dH \abs{\A} K / \delta)}}
        .
    \end{align*}
\end{theorem}

\paragraph{Discussion.}
There are several conceptual differences between REPO and RE-LSVI.
In RE-LSVI, the ensemble's decision is greedy with respect to the current value estimates (\cref{line:LSVI-ensemble} in \cref{alg:RE-LSVI}). This works because the individual policies $\pi^{k,i}_h$ are greedy with respect to their values. In contrast, the PO sub-routines in REPO produce non-greedy policies $\pi^{k,i}_{h}$ whose performance is not competitive alone, but only as an entire sequence. Thus, the ensemble decisions must ensure that the chosen policies are competitive with the best contiguous sequence in hindsight. This is ensured by hedging over the values, i.e., choosing randomly with probabilities governed by the MW rule (\cref{line:REPO-Hedge}).

The second difference relating to the ensemble method is the use of an epoch schedule. 
This is a fairly standard doubling trick that allows us to keep the covariance matrix fixed in intervals (at negligible cost). We use it to draw the reward perturbations only once at the start of the epoch and keep them fixed throughout the epoch. Since the policies are only competitive as a sequence, they must also be optimistic as a sequence.\footnote{We note that the notion of optimism in PO algorithms is slightly different compared to value iteration methods.}
As the reward perturbations encourage this optimism, our analysis depends on their sum inside an epoch, which is simply a single perturbation multiplied by the length of the epoch. Had we drawn a new perturbation for each episode, their covariances would be correlated, thus their sum would be non-Gaussian, which breaks our optimistic guarantees.
We suspect that a Berry-Esseen type argument for martingales might show that this distribution would converge to Gaussian, thus obviating the need for an epoch schedule. We were unable to verify this and leave it for future work.
%

%
As we mentioned previously, REPO replaces the value clipping with an indicator mechanism by \citet{sherman2023rate}, which zeroes-out the Q values outside ``known'' states sets, obtained during a warm-up period.
This change is necessary to control the log covering number of the policy class, on which the regret has a square root dependence. Notice that our policies are a soft-max over the sum of past Q functions, i.e., $\pi^{k,i}_h (a | x) \propto \exp\brk{\etaO \sum_{k=\kEpoch}^{k-1} \hat{Q}_h^{k,i}(x,a)}$. Each $\hat{Q}_h^{k,i}$ has $d$ parameters ($w_h^{k,i}$), thus the policy class may have $d K$ parameters in the worst case. The log covering number would also scale similarly, significantly deteriorating the regret. When Q values are calculated using clipping (\cref{line:LSVI} of \cref{alg:RE-LSVI}), we cannot avoid this scenario. However, using the indicator mechanism, the policy class may be summarized as
$
\pi^{k,i}_h (a | x)
\propto
\exp\brk{\etaO 
\phi(x,a)\tran w
\cdot \indEvent{ x \in \Z_h }
}
,
$
for some $w \in \RR[d]$, which has only $d$ parameters.

Finally, we elaborate on our choice of purely stochastic exploration bonuses. Stochastic bonuses are typically larger than their deterministic counterparts by a factor of $\sqrt{d}$. While this usually makes them unfavorable, in our context they compensate for this deficit by reducing the complexity of the policy class. Concretely, it reduces its log covering number from $H d^3$ to $d$, yielding an overall improvement to the regret bound. 
We note that the same cannot be said for RE-LSVI, where our clipping mechanism necessitates the use of deterministic bonuses for the dynamics uncertainty. To better understand this issue, see the analysis in \cref{sec:analysis:LSVI}.

\paragraph{REPO for Tabular MDPs with ABF.} 
As mentioned, a policy optimization algorithm with ABF did not exist before our work even for the tabular setting.
In \cref{sec:REPO-tabular}, we show a simplification of REPO for tabular MDPs with ABF. 
There, we do not restrict the Q values at all (no indicator or clipping), and thus do not need the warm-up routine by \citet{sherman2023rate}. 
Otherwise, the algorithms are essentially identical (up to parameter settings). 
Our analysis for the tabular case is also novel and follows a regret decomposition that has not been applied in the context of policy optimization to the best of our knowledge.
It allows us to incorporate the optimal value $V^\star$ instead of the estimated value $\hat{V}^{k,i}$ in some of the terms, thus avoiding complications with bounding the covering number of the value functions class.
However, it relies on the estimated dynamics backup operators being (nearly) valid distributions, i.e., have non-negative entries and sum to less than $1$, and thus cannot be applied to our current implementation for linear MDPs unfortunately.

\section{Analysis}
In this section we sketch our main proof ideas. For full details see \cref{appendix-sec:RE-LSVI} (RE-LSVI) and \cref{appendix-sec:REPO} (REPO).

\subsection{Analysis of RE-LSVI}
\label{sec:analysis:LSVI}

\paragraph{Overview.}
We start by decomposing the regret as:
\begin{align*}
    \regret
    &
    =
    \sum_{k \in [K]} 
    \underbrace{V^\star(x_1) - \hat{V}^{k,i_k}_1(x_1)}_{(i)}
    +
    \underbrace{\hat{V}^{k,i_k}_1(x_1) - V^{\piK}(x_1)}_{(ii)}
    .
\end{align*}
Term $(i)$ (optimism) reflects the difference between the performance of the optimal policy $\piOpt$ and the optimistically estimated performance of the agent's policy $\piK$. 
Term $(ii)$ (cost of optimism) reflects the difference between the true and estimated performance of $\piK$.

Next, we show in \cref{lemma:optimism,lemma:cost-of-optimism} that, conditioned on a ``good event'' that holds with probability $1 - \delta$ (see \cref{lemma:good-event}), $(i) \le 0$, i.e., the algorithm is indeed optimistic, and 
\begin{align*}
    (ii)
    \le
    \EE[P, \piK]\brk[s]*{
    (\betaR + \betaZ) \norm{\phi^k}_{(\covK)^{-1}} 
    +
    2 \betaP \sum_{h \in [H]} \norm{\phi^k_h}_{(\covKH)^{-1}}
    }
    ,
\end{align*}
where $\betaZ$ is a high-probability bound on the magnitude of the noise $\zeta^{k,i}$.
Applying Azuma's inequality to the sum of $(ii)$ over $k$ (a part of the good event), we conclude that
\begin{align*}
    \regret
    &
    \le
    (\betaR + \betaZ) 
    \sum_{k \in [K]} 
    \norm{\phi^k}_{(\covK)^{-1}} 
    +
    2 \betaP \sum_{h \in [H]}
    \sum_{k \in [K]}
    \norm{\phi^k_h}_{(\covKH)^{-1}}
    \\
    &
    +
    (\betaR + \betaZ + 2H \betaP) \sqrt{2 K \log (5 / \delta)}
    .
\end{align*}
\mbox{The proof is concluded by bounding the terms $\sum_{k} \norm{\phi^k}_{(\covK)^{-1}}$} and $\sqrt{H}\sum_{k} \norm{\phi^k_h}_{(\covKH)^{-1}}$ as $\tilde{O} (\sqrt{dHK})$ using the Cauchy-Schwarz inequality and \cref{lemma:elliptical-potential}, a standard elliptical potential lemma \citep[see, e.g.,][Lemma 13]{cohen2019learning}.


\paragraph{Optimism.}
In the remainder of this section, we explain the main claims showing that $(i) \le 0$.
The proof for term $(ii)$ follows similar arguments.
First, we use a value difference lemma by \citet{shani2020optimistic} to get that for every $i \in [m]$:
\begin{align}
    \label{eq:LSVI-regret-term-i-decomp}
    &
    V^\star(x_1) - \hat{V}^{k,i}_1(x_1)
    \\
    \nonumber
    &
    =
    \EE[P, \piOpt]\sum_{h \in [H]} 
    \underbrace{\hat{Q}_h^{k,i}(x_h, \piOpt[h](x_h)) - \hat{Q}_h^{k,i}(x_h, \pi^{k,i}_h(x_h))
    }_{(**)}
    \\
    \nonumber
    &
    +
    \underbrace{
    \EE[P, \piOpt]\sum_{h \in [H]} \phi(x_h,a_h)\tran(\theta_h + \psiH \hat{V}_{h+1}^{k,i}) - \hat{Q}_h^{k,i}(x_h, a_h)
    }_{(*)}
    .
\end{align}
By the greedy definition of $\pi^{k,i}_h$, $(**) \le 0$. 
Next, suppose that, for all $k \in [K], h \in [H], i \in [m]$, the estimation error of the dynamics backup operator is well-concentrated
\begin{align}
\label{eq:analysis-lsvi-dynamics}
    \norm{(\psi_h - \widehat{\psi}_h^k) \hat{V}_{h+1}^{k,i}}_{\covKH} \le \betaP
    ,
\end{align}
and the perturbed estimated reward is bounded
\begin{align}
\label{eq:analysis-lsvi-bounded-reward}
    \abs{\phi(x,a)\tran(\thetaKH + \zh^{k,i})}
    \le
    \betaRZ
    ,
    \forall x \in \X, a \in \A
    ,
\end{align}
both consequences of the good event. Then we have that 
\begin{align*}
    \phi&(x,a)\tran w_h^{k,i}
    +
    \betaP \norm{\phi(x,a)}_{(\covKH)^{-1}}
    \\
    &
    =
    \phi(x,a)\tran 
    (\thetaKH + \zh^{k,i}
    +
    \psiKH \hat{V}_{h+1}^{k,i})
    +
    \betaP \norm{\phi(x,a)}_{(\covKH)^{-1}}
    \\
    \tag{Cauchy-Schwarz}
    &
    \ge
    \phi(x,a)\tran 
    (
    \thetaKH + \zh^{k,i}
    +
    \psiH \hat{V}_{h+1}^{k,i}
    )
    \\
    &
    +
    (\betaP - \norm{(\psiKH - \psiH) \hat{V}_{h+1}^{k,i}}_{\covKH})
    \norm{\phi(x,a)}_{(\covKH)^{-1}}
    \\
    \tag{\cref{eq:analysis-lsvi-dynamics}}
    &
    \ge
    \phi(x,a)\tran 
    (\thetaKH + \zh^{k,i}
    +
    \psiH \hat{V}_{h+1}^{k,i})
    .
\end{align*}
Due to the clipping of $\hat{V}_{h+1}^{k,i}$ and \cref{eq:analysis-lsvi-bounded-reward}, the last term is absolutely bounded by $(H+1-h)\betaRZ$. Combined with the fact that clipping is non-decreasing, we conclude that
\begin{align*}
    \hat{Q}_h^{k,i}(x,a)
    \ge
    \phi(x,a)\tran 
    (\thetaKH + \zh^{k,i}
    +
    \psiH \hat{V}_{h+1}^{k,i})
    ,
\end{align*}
and plugging this back into $(*)$, we get that
\begin{align}
    \nonumber
    (*)
    &
    \le
    \EE[P, \piOpt]
    \sum_{h \in [H]} 
    \phi(x_h,a_h)\tran
    (\thetaH - \thetaKH - \zh^{k,i})
    \\
    \label{eq:optimism-term-i*}
    &
    =
    {\phi^{\piOpt}}\tran
    (\theta - \thetaK - \z^{k,i})
    ,
\end{align}
where $\phi^{\piOpt} = \EE[P, \piOpt](\phi(x_1,a_1)\tran \ldots, \phi(x_H, a_H)\tran)\tran.$

Now, suppose that, for all $k \in [K]$, the aggregate reward estimation error is well concentrated
\begin{align}
    \label{eq:analysis-lsvi-ensemble-theta}
     \norm{\theta - \widehat{\theta}^k}_{\covK} \le \betaR
     ,
\end{align}
and the perturbations are optimistic in the sense that
\begin{align}
\label{eq:analysis-lsvi-ensemble}
     \max_{i \in [m]} {\phi^{\piOpt}}\tran \z^{k,i} 
     \ge
     \betaR \norm{\phi^{\piOpt}}_{(\covK)^{-1}}
     ,
\end{align}
both also a part of the good event.
Recalling the definition of $i_k$ (\cref{line:LSVI-ensemble} in \cref{alg:RE-LSVI}) and putting everything together:
\begin{align*}
    (i)
    &
    =
    \min_{i \in [m]} V^\star(x_1) - \hat{V}^{k,i}_1(x_1)
    \\
    \tag{\cref{eq:optimism-term-i*}}
    &
    \le
    {\phi^{\piOpt}}\tran (\theta - \thetaK)
    -
    \max_{i \in [m]}{\phi^{\piOpt}}\tran\z^{k,i}
    \\
    \tag{\cref{eq:analysis-lsvi-ensemble-theta}}
    &
    \le
    \betaR \norm{\phi^{\piOpt}}_{(\covK)^{-1}}
    -
    \max_{i \in [m]}{\phi^{\piOpt}}\tran\z^{k,i}
    \\
    \tag{\cref{eq:analysis-lsvi-ensemble}}
    &
    \le
    0
    .
\end{align*}
\paragraph{Proof (sketch) of \cref{eq:analysis-lsvi-ensemble}.}
We conclude this section with some intuition regarding \cref{eq:analysis-lsvi-ensemble}. Notice that its right-hand side looks like the desired deterministic bonus. On the other hand, the argument inside the $\max$ operation is the effective bonus of each member in the ensemble. \cref{eq:analysis-lsvi-ensemble} may thus be interpreted as a requirement that at least one perturbation yields the correct bonus under the optimal policy, thus will have an optimistic value. As we choose to follow the ensemble member with the largest value, it too must be optimistic. 
As for verifying that \cref{eq:analysis-lsvi-ensemble} holds with high probability, while this may appear a complex argument, it follows from the following fundamental result. Let $g_i = {\phi^{\piOpt}}\tran \z^{k,i}$ and notice that, conditioned on $\covK$, $\{g_i\}_{i \in [m]}$ are i.i.d $\mathcal{N}(0, \betaR^2 \norm{\phi^{\piOpt}}_{(\covK)^{-1}}^2)$ variables. \cref{eq:analysis-lsvi-ensemble} thus holds with high probability by the following anti-concentration result.
\begin{lemma}
    Let $\sigma, m \ge 0$. Suppose that $g_i \sim \mathcal{N}(0, \sigma^2)$, $i\in[m]$ are i.i.d Gaussian random variables. With probability at least $1 - e^{-m/9}$
    \begin{align*}
        \max_{i \in [m]} g_i
        \ge
        \sigma
        .
    \end{align*}
\end{lemma}
\begin{proof}
    Recall that for a standard Gaussian random variable $G \sim \mathcal{N}(0,1)$ we have that $\Pr[G \ge 1] \ge 1/9$.
    Since $g_i$ are independent, we conclude that
    \begin{align*}
        \Pr[
        \max_{i \in [m]} g_i
        \le
        \sigma
        ]
        =
        \Pr[G \le 1]^m
        \le
        \brk*{8/9}^m
        \le
        e^{-m/9}
        ,
    \end{align*}
    and taking the complement concludes the proof.
\end{proof}

\subsection{Analysis of REPO}
\label{sec:analysis:REPO}

\paragraph{Regret decomposition.}
We start with a coarse bound on the regret incurred during the warm-up period and decompose the remaining term as in RE-LSVI (\cref{sec:analysis:LSVI}), i.e., 
\begin{align*}
    \regret
    \le
    H \firstEpisodeAfterWarmup
    &
    +
    \underbrace{
    \sum_{e \in \setEpochs}
    \sum_{k\in \setEpochsE} V^\star_1(x_1) - \hat{V}^{k,i_k}_1(x_1)
    }_{(i)}
    \\
    &
    +
    \underbrace{\sum_{e \in \setEpochs}\sum_{k\in \setEpochsE} \hat{V}^{k,i_k}_1(x_1) - V^{\pi^k}_1(x_1)}_{(ii)}
    ,
\end{align*}
where $\numEpochs \approx d H \log K$ is the number of epochs and $\setEpochsE$ is the set of episodes within epoch $e$.
Recalling that $\firstEpisodeAfterWarmup$ is the length of the warm-up routine, a result by \citet{sherman2023rate} (see \cref{lemma:reward-free-exploration}) guarantees that $\firstEpisodeAfterWarmup \approx 1 / \epsCov \approx \sqrt{K}$.

As in \cref{sec:analysis:LSVI}, we focus on bounding $(i)$ since bounding $(ii)$ uses a subset of the necessary techniques.
We would have liked to decompose $(i)$ for each episode separately (as in \cref{eq:LSVI-regret-term-i-decomp}). However, $\pi_h^{k,i}$ are no longer greedy, thus $(**)$ in \cref{eq:LSVI-regret-term-i-decomp} would not be non-positive anymore. We thus perform the same decomposition but over the entire epoch to get that for every $e \in \setEpochs, i \in [m]$:
\begin{align*}
    \sum_{k \in \setEpochsE}
    V^\star(x_1)& - \hat{V}^{k,i}_1(x_1)
    \\
    &
    =
    \EE[P, \piOpt] \sum_{h \in [H]} R_{Q,h}^{i}(x_h)
    +
    \sum_{k \in \setEpochsE} R_{Opt}^{k,i}
    ,
\end{align*}
where $R_{Opt}^{k,i}$ is defined as $(*)$ in \cref{eq:LSVI-regret-term-i-decomp} and $R_{Q,h}^{i}(x_h)$ is
\begin{align*}
    \sum_{k \in \setEpochsE}  \sum_{a \in \A}  \hat{Q}^{k,i}_h(x_h,a) \brk{ \pi^\star_h(a \mid x_h) - \pi^{k,i}_h(a \mid x_h) }
    .
\end{align*}

\paragraph{OMD.}
Since $\pi_h^{k,i}$ are updated using OMD with learning rate $\etaO > 0$ (see \cref{line:REPO} in \cref{alg:r-opo-for-linear-mdp}), we can invoke a standard OMD argument (\cref{lemma:omd-term-PO-linear}) to get that
\begin{align*}
    R_{Q,h}^{i}(x_h)
    \le
    \frac{\log\abs{\A}}{\etaO} + \etaO \abs{\setEpochsE} \betaQ^2
    ,
    \;\;\forall x_h \in \X
    ,
\end{align*}
where $\betaQ$ is a bound on $\max_{x \in \X, a \in \A} \abs{\hat{Q}_h^{k,i}(x,a)}$.
Notice that, unlike RE-LSVI, $\hat{Q}^{k,i}_h$ are not bounded by definition in REPO. Nonetheless, we adapt arguments from \citet{sherman2023rate} to show that $\betaQ \approx \betaR H \sqrt{d}$ as part of a good event, which includes the events already defined in \cref{sec:analysis:LSVI}.

\paragraph{Hedge.}
Next, we show (at the end of this section) that on the good event, there exists $\hatJE \in [m]$ such that for $\epsCov \approx 1/\sqrt{K}$ and all $k \in \setEpochsE$
\begin{align}
\label{eq:REPO-analysis-Ropt-Bound}
    R_{Opt}^{k,\hatJE}
    \le
    2 \epsCov H \betaQ
    .
\end{align}
The connection between $i_k$ and $\hatJE$ is done through the MW update rule (\cref{line:REPO-Hedge} in \cref{alg:r-opo-for-linear-mdp}). For learning rate $\etaH > 0$, a standard result (\cref{lemma:hedge-term-PO-linear}) implies that 
\begin{align*}
    \sum_{e \in \setEpochs} \sum_{k \in \setEpochsE}&
    \hat{V}^{k,j_e}_1(x_1) -  \hat{V}^{k,i_k}_1(x_1)
    \\
    &
    \le
    \frac{\numEpochs \log m}{\etaH} + \etaH K \betaQ^2 + 2 \betaQ  \sqrt{2K \log\frac{7}{\delta}}
    .
\end{align*}
\paragraph{Bounding term (i).}
Putting everything together, we have
\begin{align*}
    (i)
    &
    =
    \sum_{e \in \setEpochs} \sum_{k \in \setEpochsE} \hat{V}^{k,\hatJE}_1(x_1) -  \hat{V}^{k,i_k}_1(x_1)
    \\
    &
    \quad
    +
    \sum_{e \in \setEpochs} \sum_{k \in \setEpochsE} V^\star_1(x_1) -  \hat{V}^{k,\hatJE}_1(x_1)
    \\
    &
    \le
    \frac{\numEpochs \log m}{\etaH} + \etaH K \betaQ^2 + 2 \betaQ  \sqrt{2K \log\frac{7}{\delta}}
    \\
    &
    \quad
    +
    \frac{E H\log\abs{\A}}{\etaO} + \etaO H \betaQ^2 K
    +
    2 \epsCov H \betaQ K
    ,
\end{align*}
and plugging the parameter choices concludes the desired $O(\sqrt{K})$ bound.
Notice that calculating $\hatJE$ can only be done at the end of an epoch as it essentially maximizes the sum of estimated values inside an epoch. However, in every episode we need to choose a member of the ensemble whose policy we follow. This demonstrates the necessity of hedging over the ensemble in REPO (\cref{line:REPO-Hedge} in \cref{alg:r-opo-for-linear-mdp}), as opposed to the greedy method in RE-LSVI (\cref{line:LSVI-ensemble} in \cref{alg:RE-LSVI}).

\paragraph{Proof (sketch) of \cref{eq:REPO-analysis-Ropt-Bound}.}
To begin, we show in \cref{lemma:cost-of-truncated-features} that the reward-free warm-up explores the state space well in the sense that, for any policy $\pi \in \piClass$ and vector $v \in \RR[d]$
\begin{align*}
        \abs*{
        \EE[P, \pi]
        (\phi(x_h,a_h) - \bar{\phi}_h(x_h,a_h))\tran v
        }
        \le
        \epsCov \max_{x,a} \abs{\phi(x,a)\tran v}
        ,
    \end{align*}
    where $\bar{\phi}_h(x,a) = \phi(x,a) \indEvent{x \in \Z_h}$ are truncated features according to the known states set $\Z_h$.
    We then rewrite $R_{Opt}^{k,i}$ in the following way using the definition of $\hat{Q}_h^{k,i}(x_h, a_h)$:
    \begin{align*}
        &
        R_{Opt}^{k,i}
        =
        \underbrace{
        \sum_{h \in [H]} \brk{\bar{\phi}^{\piOpt}_h}\tran \brk*{ \thetaH - \thetaKH - \zh^{\kEpoch,i} + (\psiH -\psiKH)\hat{V}^{k,i}_{h+1} }
        }_{(a)}
        \\
        &
        \;
        +
        \underbrace{
        \EE[P, \piOpt] \sum_{h \in [H]} (\phi(x_h,a_h) - \bar{\phi}_h(x_h, a_h))\tran \brk{ \thetaH + \psiH \hat{V}^{k,i}_{h+1} }
        }_{(b)}
        ,
    \end{align*}
    where $\bar{\phi}^{\piOpt}_h = \EE[P, \piOpt]{\bar{\phi}_h(x_h,a_h)}$.
    Recall that, on the good event, $\abs{\hat{V}^{k,i}_h(x)} \le \betaQ$, thus the above argument bounds $(b)$ by $2 \epsCov H \betaQ$.
    Letting $\bar{\phi}^{\piOpt} = ((\bar{\phi}^{\piOpt}_1)\tran, \ldots, (\bar{\phi}^{\piOpt}_H)\tran)\tran$, we apply the Cauchy-Schwarz inequality together with the least squares estimation bounds (\cref{eq:analysis-lsvi-ensemble-theta,eq:analysis-lsvi-dynamics}) to get that
    \begin{align*}
        (a)
        &
        \le
        \betaR \norm{\bar{\phi}^{\piOpt}}_{(\covK)^{-1}}
        +
        \betaP \hspace{-0.4em} \sum_{h \in [H]} \hspace{-0.3em}\norm{\bar{\phi}^{\piOpt}_h}_{(\covKH)^{-1}}
        -
        \brk{\bar{\phi}^{\piOpt}}\tran \z^{\kEpoch,i}
        \\
        &
        \le
        \norm{\bar{\phi}^{\piOpt}}_{\covZ}
        -
        \brk{\bar{\phi}^{\piOpt}}\tran \z^{\kEpoch,i}
        ,
    \end{align*}
    where the last inequality also uses $\hatCovKH \preceq \covKH, \hatCovK \preceq \covK$, the definition of $\covZ$, and the Cauchy-Schwarz inequality.
    Finally, similarly to \cref{eq:analysis-lsvi-ensemble}, we have that with high probability there exists $\hatJE \in [m]$ such that the above is at most zero.

\section*{Broader Impact} This paper presents work whose goal is to advance the field of Machine Learning. There are many potential societal consequences of our work, none of which we feel must be specifically highlighted here.

\begin{ack}
    This project has received funding from the European Research Council (ERC) under the European Union’s Horizon 2020 research and innovation program (grant agreement No. 101078075).
    Views and opinions expressed are however those of the author(s) only and do not necessarily reflect those of the European Union or the European Research Council. Neither the European Union nor the granting authority can be held responsible for them. 
    This work received additional support from the Israel Science Foundation (ISF, grant number 2549/19), the Len Blavatnik and the Blavatnik Family Foundation, and the Israeli VATAT data science scholarship.
\end{ack}

\bibliography{bibliography}
\bibliographystyle{icml2024}

\newpage
\appendix
\onecolumn

\section{Proofs of Randomized Ensemble Least Squares Value Iteration (RE-LSVI)}
\label{appendix-sec:RE-LSVI}

\subsection{Proof of \cref{thm:regret-RE-LSVI}}
We begin by defining a so-called ``good event'', followed by optimism and cost of optimism, and conclude with the proof of \cref{thm:regret-RE-LSVI}.

\paragraph{Good event.}
We define the following good event $\Egood = E_1 \cap E_2 \cap E_3 \cap E_4 \cap E_5$, over which the regret is deterministically bounded:
\begin{flalign}
    \label{eq:goodTheta}
     &
     E_1 = \brk[c]*{\forall k \in [K] : \norm{\theta - \widehat{\theta}^k}_{\covK} \le \betaR};
     &
     \\
     \label{eq:goodPsi}
     &
     E_2 = \brk[c]*{\forall k \in [K], h \in [H], i \in [m] : \norm{(\psi_h - \widehat{\psi}_h^k) \hat{V}_{h+1}^{k,i}}_{\covKH} \le \betaP};
     &
     \\
     \label{eq:goodZetaConcentration}
     &
     E_3 = \brk[c]*{\forall k \in [K], i \in [m] : \norm{\z^{k,i}}_{\covK} \le \betaZ \betaR};
     &
     \\
     \label{eq:goodZetaAntiConcentration}
     &
     E_4 = \brk[c]*{\forall k \in [K] : \max_{i \in [m]} \brk{\phi^{\piOpt}}\tran \z^{k,i} \ge \betaR \norm{\phi^{\piOpt}}_{(\covK)^{-1}}};
     &
     \\
     \label{eq:goodBernstein}
    &
    E_5 
    =
    \brk[c]*{
    \sum_{k \in [K]} \EE[P, \piK] \brk[s]*{Y_k} \le \sum_{k \in [K]} Y_k + (\betaR (1 + \betaZ) + 2H \betaP) \sqrt{2K \log\frac{5}{\delta}}
    }
    ;
    &
\end{flalign}
where
$
    \phi^{\piOpt}
    =
    \EE[P, \piOpt]\brk[s]{\phi(x_{1:H}, a_{1:H})}
    ,
    \phi(x_{1:H}, a_{1:H})
    =
    (\phi(x_1, a_1)\tran, \ldots, \phi(x_H, a_H)\tran)\tran
    ,
$
and
$Y_k = (\betaR + \betaZ) \norm{\phi^k}_{(\covK)^{-1}} 
+
2 \betaP
\sum_{h \in [H]} 
\norm{\phi^k_h}_{(\covKH)^{-1}}
.
$

\begin{lemma}[Good event]
\label{lemma:good-event}
    Consider the following parameter setting:
    \begin{gather*}
        \lambdaP = 1
        ,
        \lambdaR = H
        ,
        m = 9 \log (5K / \delta)
        ,
        \betaR
        =
        2 H \sqrt{2dH \log (10K/\delta)}
        ,
        \\
        \betaZ
        =
        \sqrt{
        \frac{11 d H}{2} \log \frac{5 m K}{\delta}
        }
        ,
        \betaRZ
        =
        2 \betaZ \betaR / \sqrt{H}
        ,
        \betaP
        =
        40 \betaZ \betaR d \sqrt{H
            \log \brk*{163 K d H / \delta}
        }
    \end{gather*}
    Then $\Pr[\Egood] \ge 1 - \delta.$
\end{lemma}

\begin{lemma}
\label{lemma:bounded-emprical-reward}
    Under the parameter choices in \cref{lemma:good-event} and the good event $\Egood$ (\cref{eq:goodTheta,eq:goodPsi,eq:goodZetaConcentration,eq:goodZetaAntiConcentration,eq:goodBernstein}), we have
    \begin{align*}
        \abs{
            \phi(x,a)\tran 
            (\thetaKH + \zh^{k,i}
            +
            \psiH \hat{V}_{h+1}^{k,i})
        }
        &
        \le
        (H+1-h)\betaRZ
        \qquad
        , \forall h \in [H]
        .
    \end{align*}
\end{lemma}
This second result is a straightforward calculation that follows from the first (proofs in \cref{sec:good-event-proofs}).

\paragraph{Optimism and its cost.}
The following two results show that our value construction is optimistic concerning the true performance of the optimal policy but not overly optimistic compared with the performance of its induced policy.
\begin{lemma}[Optimism]
\label{lemma:optimism}
    Suppose that the good event $\Egood$ holds (\cref{eq:goodTheta,eq:goodPsi,eq:goodZetaConcentration,eq:goodZetaAntiConcentration,eq:goodBernstein}), then
    \begin{align*}
        V^\star(x_1)
        -
        \hat{V}^{k, i_k}_1(x_1)
        \le 
        0
        \quad
        ,
        \forall k \in [K]
        .
    \end{align*}
\end{lemma}
\begin{proof}
    First, we use \cref{lemma:extended-value-difference}, a value difference lemma by \citet{shani2020optimistic}, to get that for every $i \in [m]$
    \begin{align*}
        V^\star(x_1) - \hat{V}^{k,i}_1(x_1)
        &
        =
        \EE[P, \piOpt]\sum_{h \in [H]} 
        \hat{Q}_h^{k,i}(x_h, \piOpt[h](x_h)) - \hat{Q}_h^{k,i}(x_h, \pi^{k,i}_h(x_h))
        \\
        &
        \quad
        +
        \EE[P, \piOpt]\sum_{h \in [H]} \phi(x_h,a_h)\tran(\theta_h + \psiH \hat{V}_{h+1}^{k,i}) - \hat{Q}_h^{k,i}(x_h, a_h)
        \\
        &
        \le
        \EE[P, \piOpt]\sum_{h \in [H]} \phi(x_h,a_h)\tran(\theta_h + \psiH \hat{V}_{h+1}^{k,i}) - \hat{Q}_h^{k,i}(x_h, a_h)
        ,
    \end{align*}
    where the inequality is by the greedy definition of $\pi^{k,i}_h$.
    Next, because $\clip$ is a non-decreasing operator, we have that 
    \begin{align*}
        \hat{Q}_h^{k,i}(x,a)
        &
        =
        \clip_{(H+1-h)\betaRZ}\brk[s]*{
        \phi(x,a)\tran w_h^{k,i}
        +
        \betaP \norm{\phi(x,a)}_{(\covKH)^{-1}}
        }
        \\
        &
        =
        \clip_{(H+1-h)\betaRZ}\brk[s]*{
        \phi(x,a)\tran 
        (\thetaKH + \zh^{k,i}
        +
        \psiKH \hat{V}_{h+1}^{k,i})
        +
        \betaP \norm{\phi(x,a)}_{(\covKH)^{-1}}
        }
        \\
        \tag{Cauchy-Schwarz}
        &
        \ge
        \clip_{(H+1-h)\betaRZ}\brk[s]*{
        \phi(x,a)\tran 
        (
        \thetaKH + \zh^{k,i}
        +
        \psiH \hat{V}_{h+1}^{k,i}
        )
        +
        (\betaP - \norm{(\psiKH - \psiH) \hat{V}_{h+1}^{k,i}}_{\covKH})
        \norm{\phi(x,a)}_{(\covKH)^{-1}}
        }
        \\
        \tag{\cref{eq:goodPsi}}
        &
        \ge
        \clip_{(H+1-h)\betaRZ}\brk[s]*{
        \phi(x,a)\tran 
        (\thetaKH + \zh^{k,i}
        +
        \psiH \hat{V}_{h+1}^{k,i})
        }
        \\
        \tag{\cref{lemma:bounded-emprical-reward}}
        &
        =
        \phi(x,a)\tran 
        (\thetaKH + \zh^{k,i}
        +
        \psiH \hat{V}_{h+1}^{k,i})
        .
    \end{align*}
    Plugging this back into the above, we get that
    \begin{align*}
        V^\star(x_1) - \hat{V}^{k,i}_1(x_1)
        \le
        \EE[P, \piOpt]
        \sum_{h \in [H]} 
        \phi(x_h,a_h)\tran
        (\thetaH - \thetaKH - \zh^{k,i})
        =
        {\phi^{\piOpt}}\tran
        (\theta - \thetaK - \z^{k,i})
        ,
    \end{align*}
    where $\phi^{\piOpt} = \EE[P, \piOpt](\phi(x_1,a_1)\tran \ldots, \phi(x_H, a_H)\tran)\tran.$
    Finally, recalling the definition of $i_k$ (\cref{line:LSVI-ensemble} in \cref{alg:RE-LSVI}), we get
    \begin{align*}
        V^\star(x_1)
        -
        \hat{V}^{k, i_k}_1(x_1)
        =
        \min_{i \in [m]} V^\star(x_1) - \hat{V}^{k,i}_1(x_1)
        &
        \le
        {\phi^{\piOpt}}\tran (\theta - \thetaK)
        -
        \max_{i \in [m]}{\phi^{\piOpt}}\tran\z^{k,i}
        \\
        \tag{Cauchy-Schwarz, \cref{eq:goodTheta}}
        &
        \le
        \betaR \norm{\phi^{\piOpt}}_{\covK^{-1}}
        -
        \max_{i \in [m]}{\phi^{\piOpt}}\tran\z^{k,i}
        \\
        \tag{\cref{eq:goodZetaAntiConcentration}}
        &
        \le
        0
        .
    \end{align*}
\end{proof}

\begin{lemma}[Cost of optimism]
\label{lemma:cost-of-optimism}
    Suppose that the good event $\Egood$ holds (\cref{eq:goodTheta,eq:goodPsi,eq:goodZetaConcentration,eq:goodZetaAntiConcentration,eq:goodBernstein}), then
    \begin{align*}
        \hat{V}^{k, i_k}_1(x_1)
        -
        V_1^{\piK}(x_1)
        \le
        \EE[P, \piK]\brk[s]*{
        \betaR (1 + \betaZ) \norm{\phi(x_{1:H}, a_{1:H})}_{(\covK)^{-1}} 
        +
        2 \betaP \sum_{h \in [H]} \norm{\phi(x_h, a_h)}_{(\covKH)^{-1}}
        }
        \quad
        ,
        \forall k \in [K]
        .
    \end{align*}
\end{lemma}
\begin{proof}
    First, because $\clip$ is a non-decreasing operator, we have that 
    \begin{align*}
        \hat{Q}_h^{k,i}(x,a)
        &
        =
        \clip_{(H+1-h)\betaRZ}\brk[s]*{
        \phi(x,a)\tran w_h^{k,i}
        +
        \betaP \norm{\phi(x,a)}_{(\covKH)^{-1}}
        }
        \\
        &
        =
        \clip_{(H+1-h)\betaRZ}\brk[s]*{
        \phi(x,a)\tran 
        (\thetaKH + \zh^{k,i}
        +
        \psiKH \hat{V}_{h+1}^{k,i})
        +
        \betaP \norm{\phi(x,a)}_{(\covKH)^{-1}}
        }
        \\
        \tag{Cauchy-Schwarz}
        &
        \le
        \clip_{(H+1-h)\betaRZ}\brk[s]*{
        \phi(x,a)\tran 
        (
        \thetaKH + \zh^{k,i}
        +
        \psiH \hat{V}_{h+1}^{k,i}
        )
        +
        (\betaP + \norm{(\psiKH - \psiH) \hat{V}_{h+1}^{k,i}}_{\covKH})
        \norm{\phi(x,a)}_{(\covKH)^{-1}}
        }
        \\
        \tag{\cref{eq:goodPsi}}
        &
        \le
        \clip_{(H+1-h)\betaRZ}\brk[s]*{
        \phi(x,a)\tran 
        (\thetaKH + \zh^{k,i}
        +
        \psiH \hat{V}_{h+1}^{k,i})
        +
        2\betaP \norm{\phi(x,a)}_{(\covKH)^{-1}}
        }
        \\
        \tag{\cref{lemma:bounded-emprical-reward}, $\clip_\beta\brk[s]{z} \le z, \forall z \ge -\beta$}
        &
        \le
        \phi(x,a)\tran 
        (\thetaKH + \zh^{k,i}
        +
        \psiH \hat{V}_{h+1}^{k,i})
        +
        2\betaP \norm{\phi(x,a)}_{(\covKH)^{-1}}
        .
    \end{align*}
    Next, we use \cref{lemma:extended-value-difference}, a value difference lemma by \citet{shani2020optimistic}, to get that
    \begin{align*}
        \hat{V}^{k, i_k}_1(x_1)
        -
        V_1^{\piK}(x_1)
        &
        =
        \EE[P, \piOpt]\brk[s]*{
        \sum_{h \in [H]} 
        \hat{Q}_h^{k,i}(x_h, a_h)
        -
        \phi(x_h,a_h)\tran(\theta_h + \psiH \hat{V}_{h+1}^{k,i})
        }
        \\
        &
        \le
        \EE[P, \piOpt]\brk[s]*{
        \sum_{h \in [H]} 
        \phi(x_h,a_h)\tran
        (\thetaH - \thetaKH - \zh^{k,i})
        +
        2\betaP \norm{\phi(x_h,a_h)}_{(\covKH)^{-1}}
        }
        \\
        &
        =
        \EE[P, \piOpt]\brk[s]*{
        \phi(x_{1:H},a_{1:H})\tran
        (\theta - \thetaK - \z^{k,i})
        +
        2\betaP \sum_{h \in [H]} \norm{\phi(x_h,a_h)}_{(\covKH)^{-1}}
        }
        \\
        \tag{Cauchy-Schwarz, \cref{eq:goodTheta,eq:goodZetaConcentration}}
        &
        \le
        \EE[P, \piOpt]\brk[s]*{
        \betaR (1 + \betaZ)\norm{\phi(x_{1:H},a_{1:H})}_{(\covK)^{-1}}
        +
        2\betaP \sum_{h \in [H]} \norm{\phi(x_h,a_h)}_{(\covKH)^{-1}}
        }
        .
    \end{align*}
\end{proof}

\paragraph{Regret bound.}
The following is our main result for \cref{alg:RE-LSVI}.
\begin{theorem*}[restatement of \cref{thm:regret-RE-LSVI}]
    Suppose that we run RE-LSVI (\cref{alg:RE-LSVI}) with the parameters defined in \cref{lemma:good-event}.
    Then with probability at least $1 - \delta$, we have
    \begin{align*}
        \regret
        \le
        1088 \sqrt{ d^5 H^7 K} \log^2 \brk*{1467 K d H \log(5K/\delta) \big/ \delta}
        =
        \tilde{O}(\sqrt{ d^5 H^7 K})
        .
    \end{align*}
\end{theorem*}
\begin{proof}
    Suppose that the good event $\Egood$ holds (\cref{eq:goodTheta,eq:goodPsi,eq:goodZetaConcentration,eq:goodZetaAntiConcentration,eq:goodBernstein}). By \cref{lemma:good-event}, this holds with probability at least $1-\delta$.
    We conclude that
    \begin{align*}
        \regret
        &
        =
        \sum_{k \in [K]} V^\star(x_1) - \hat{V}_1^{k, i_k}(x_1)
        +
        \sum_{k \in [K]} \hat{V}_1^{k, i_k}(x_1) - V^{\piK}(x_1)
        \\
        \tag{\cref{lemma:optimism,lemma:cost-of-optimism}}
        &
        \le
        \sum_{k \in [K]}
        \EE[P, \piK]\brk[s]*{
        \betaR (1 + \betaZ) \norm{\phi^k}_{(\covK)^{-1}} 
        +
        2 \betaP \sum_{h \in [H]} \norm{\phi^k_h}_{(\covKH)^{-1}}
        }
        \\
        \tag{\cref{eq:goodBernstein}}
        &
        \le
        \betaR (1 + \betaZ) \sum_{k \in [K]} \norm{\phi^k}_{(\covK)^{-1}} 
        +
        2 \betaP \sum_{h \in [H]} \sum_{k \in [K]} \norm{\phi^k_h}_{(\covKH)^{-1}}
        +
        (\betaR(1 + \betaZ) + 2H \betaP) \sqrt{2K \log\frac{5}{\delta}}
        \\
        \tag{\cref{lemma:elliptical-potential}, $\norm{\phi^k}^2 \le H = \lambdaR, \norm{\phi_h^k}^2 \le 1 = \lambdaP$}
        &
        \le
        \betaR (1 + \betaZ) \sqrt{2 d H K \log(2K)}
        +
        2 \betaP H \sqrt{2 d K \log(2K)}
        +
        (\betaR (1 + \betaZ) + 2H \betaP) \sqrt{2K \log\frac{5}{\delta}}
        \\
        \tag{$\sqrt{x} + \sqrt{y} \le \sqrt{2x + 2y}$}
        &
        \le
        2 \betaR (1 + \betaZ) \sqrt{ d H K \log\brk{10K/\delta}}
        +
        4 \betaP H \sqrt{ d K \log\brk{10K/\delta}}
        \\
        &
        \le
        4 \betaZ \betaR \sqrt{ d H K \log\brk{10K/\delta}}
        +
        160 \betaZ \betaR \sqrt{ d^3 H^3 K} \log \brk*{163 K d H / \delta}
        \\
        &
        \le
        164 \betaZ \betaR \sqrt{ d^3 H^3 K} \log \brk*{163 K d H / \delta}
        \\
        &
        \le
        1088 \sqrt{ d^5 H^7 K} \log^2 \brk*{163 m K d H / \delta}
        \\
        &
        \le
        1088 \sqrt{ d^5 H^7 K} \log^2 \brk*{1467 K d H \log(5K/\delta) \big/ \delta}
        ,
    \end{align*}
    where the last four transitions used our parameter choices.
\end{proof}

\subsection{Proofs of good event (RE-LSVI)}
\label{sec:good-event-proofs}

\begin{lemma*}[restatement of \cref{lemma:bounded-emprical-reward}]
    Under the parameter choices in \cref{lemma:good-event} and the good event $\Egood$ (\cref{eq:goodTheta,eq:goodPsi,eq:goodZetaConcentration,eq:goodZetaAntiConcentration,eq:goodBernstein}), we have
    \begin{align*}
        \abs{
            \phi(x,a)\tran 
            (\thetaKH + \zh^{k,i}
            +
            \psiH \hat{V}_{h+1}^{k,i})
        }
        &
        \le
        (H+1-h)\betaRZ
        \qquad
        , \forall h \in [H]
        .
    \end{align*}
\end{lemma*}
\begin{proof}
    Under $\Egood$ we have for all $k \in [K], h \in [H], i \in [m], a \in \A, x \in \X$:
    \begin{align*}
        \tag{$\phi(x,a)\tran \thetaH \in [0,1]$, $\norm{\phi(x,a)} \le 1$}
        \abs{\phi(x,a)\tran (\thetaKH + \zh^{k,i})}
        &
        \le
        1
        +
        \norm{\thetaKH - \thetaH + \zh^{k,i}}
        \\
        \tag{triangle inequality}
        &
        \le
        1 + \norm{\thetaK - \theta} + \norm{\z^{k,i}}
        \\
        \tag{$\covK \succeq \lambdaR I$}
        &
        \le
        1
        +
        \brk{\norm{\thetaK - \theta}_{\covK} + \norm{\z^{k,i}}_{\covK}} / \sqrt{\lambdaR} 
        \\
        \tag{\cref{eq:goodTheta,eq:goodZetaConcentration}}
        &
        \le
        1
        +
        \betaR (1 + \betaZ) / \sqrt{\lambdaR} 
        \\
        \tag{$\betaZ \ge 2, \lambdaR = H$}
        &
        \le
        \betaRZ,
    \end{align*}
    and thus we also get that
    \begin{align*}
        \abs{
            \phi(x,a)\tran 
            (\thetaKH + \zh^{k,i}
            +
            \psiH \hat{V}_{h+1}^{k,i})
        }
        &
        \le
        \betaRZ
        +
        \abs{
            \phi(x,a)\tran 
            \psiH \hat{V}_{h+1}^{k,i}
        }
        \\
        &
        =
        \betaRZ
        +
        \abs{
            \EE[x' \sim P_h(\cdot \mid x,a)] \hat{V}_{h+1}^{k,i}(x')
        }
        \\
        &
        \le
        \betaRZ
        +
        \max_{x' \in \X, a' \in \A} \abs{\hat{Q}_{h+1}^{k,i}(x',a')}
        \le
        (H+1-h)\betaRZ
        ,
    \end{align*}
    where the last inequality is due to the clipping of $\hat{Q}_{h+1}^{k,i}$ (\cref{line:LSVI,alg:RE-LSVI}).
\end{proof}

\begin{lemma*}[restatement of \cref{lemma:good-event}]
    Consider the following parameter setting:
    \begin{gather*}
        \lambdaP = 1
        ,
        \lambdaR = H
        ,
        m = 9 \log (5K / \delta)
        ,
        \betaR
        =
        2 H \sqrt{2dH \log (10K/\delta)}
        ,
        \\
        \betaZ
        =
        \sqrt{
        \frac{11 d H}{2} \log \frac{5 m K}{\delta}
        }
        ,
        \betaRZ
        =
        2 \betaZ \betaR / \sqrt{H}
        ,
        \betaP
        =
        40 \betaZ \betaR d \sqrt{H
            \log \brk*{163 K d H / \delta}
        }
    \end{gather*}
    Then $\Pr[\Egood] \ge 1 - \delta.$
\end{lemma*}
\begin{proof}
    First, notice that $\norm{\phi^k}_{(\covK)^{-1}}, \norm{\phi_h^k}_{(\covKH)^{-1}} \le 1$, thus $0 \le Y_k \le \betaR (1 + \betaZ) + 2H \betaP$. Using Azuma's inequality, we conclude that $E_5$ (\cref{eq:goodBernstein}) holds with probability at least $1-\delta/5$.
    Next, by \cref{lemma:reward-error} and our choice of parameters, $E_1$ (\cref{eq:goodTheta}) holds with probability at least $1- \delta/5$.
    Now, suppose that the noise is generated such that 
    $
    \z^{k,i} = \betaR (\covKH)^{-1/2} g^{k,i}
    $ 
    where $g^{k,i} \sim \mathcal{N}(0, I_dH)$ are i.i.d for all $k \in [K], i \in [m]$.
    Indeed, notice that
    \begin{align*}
        \EE (\z^{k,i})(\z^{k,i})\tran
        &
        =
        \betaR^2 (\covKH)^{-1/2} \EE \brk[s]*{(g^{k,i}) (g^{k,i})\tran} (\covKH)^{-1/2}
        =
        \betaR^2 (\covKH)^{-1}
        .
    \end{align*}
    Taking a union bound over \cref{lemma:gaussian-norm-concentration} with $\delta / 5 m K$, we have that with probability at least $1-\delta/5$, simultaneously for all $i \in [m], k \in [K]$
    \begin{align*}
        \norm{\z^{k,i}}_{\covKH}
        =
        \betaR \norm{g^{k,i}}
        \le
        \betaR
        \sqrt{\frac{11 d H}{2} \log \frac{5 m K}{\delta}}
        =
        \betaR\betaZ
        ,
    \end{align*}
    thus establishing $E_3$ (\cref{eq:goodZetaConcentration}).
    Next, notice that conditioned on $\covK$, $(\phi^{\piOpt})\tran \z^{k,i}, i \in [m]$ are i.i.d $\mathcal{N}(0, \betaR^2 \norm{\phi^{\piOpt}}_{(\covK)^{-1}}^2)$.
    Applying \cref{lemma:gaussian-anti-concentration} with $m = 9 \log (5K / \delta)$ and taking a union bound, we have that $E_4$ (\cref{eq:goodZetaAntiConcentration}) holds with probability at least $1-\delta/5$.
    %
    %
    
    Now, for any $h \in [H]$ consider the function class $\mathcal{V}_h \subseteq \RR[\X]$ of functions mapping from $\mathcal{X}$ to $\RR$ with the following parametric form
    \begin{align*}
        V(\cdot)
        =
        \clip_{(H+1-h)\betaRZ}\brk[s]*{\max_a w\tran \phi(\cdot,a) + \beta \sqrt{\phi(\cdot, a)\tran \Lambda^{-1} \phi(\cdot, a)}}
    \end{align*}
    where $(w,\beta,\Lambda)$ are parameters satisfying $\norm{w} \le 4 K H \betaRZ, \beta \in [0, \bar{\beta}]$ where 
    $
    \bar{\beta} = 876 \betaRZ d^{7/4} H K \log (5H^2 / \delta)
    ,
    $
    and
    $\lambda_{\min}(\Lambda) \ge 1$
    where $\lambda_{\min}$ denotes the minimal eigenvalue. Let $\mathcal{N}_{\epsilon,h}$ be the $\epsilon$-covering number of $\mathcal{V}_h$ with respect to the supremum distance. Then for $\epsilon = \betaRZ H \sqrt{d} / 2K$, \cref{lemma:covering-number} says that
    \begin{align*}
        \log \mathcal{N}_{\epsilon,h}
        &
        \le
        d \log\brk*{1 + \frac{16 K H \betaRZ}{\epsilon}}
        +
        d^2 \log \brk*{1 + \frac{8 \sqrt{d} \bar{\beta}^2}{\epsilon^2}}
        \\
        &
        =
        d \log\brk*{1 + \frac{32 K^2}{\sqrt{d}}}
        +
        d^2 \log \brk*{1 + \frac{32 K^2  \bar{\beta}^2}{\sqrt{d} H^2\betaRZ^2}}
        \\
        &
        \le
        4 d^2 \log \brk*{\frac{6 K \bar{\beta}}{{d}^{1/4} H \betaRZ}}
        \\
        &
        =
        4 d^2 \log \brk*{5256 K^2 d^{3/2} \log \frac{5H^2}{\delta}}
        \\
        &
        \le
        8 d^2 \log \brk*{163 K d H / \delta}
        .
    \end{align*}
    Applying \cref{lemma:dynamics-error-set-v} to $\mathcal{V}_h, h \in [H]$, we have that with probability at least $1-\delta/5$ simultaneously for all $k \in [K], h \in [H], V \in \mathcal{V}_{h+1}$
    \begin{align*}
        \norm{(\psi - \psiKH)V}_{\covKH}
        &
        \le
        4 \betaRZ H \sqrt{d \log (K+1) + 2\log (5H \mathcal{N}_\epsilon /\delta)}
        \\
        &
        \le
        4 \betaRZ H \sqrt{
        d \log (K+1)
        +
        2\log (5H^2 /\delta) 
        +
        16 d^2 \log \brk*{163 K d H / \delta}
        }
        \\
        &
        \le
        20 \betaRZ d H \sqrt{
        \log \brk*{163 K d H / \delta}
        }
        \\
        &
        =
        40 \betaZ \betaR d \sqrt{
        H
        \log \brk*{163 K d H / \delta}
        }
        \\
        &
        =
        \betaP
        .
    \end{align*}
    Taking a union bound, all of the events so far hold with probability at least $1-\delta$. 
    Finally, we show that these events also imply that $\hat{V}_{h+1}^{k,i} \in \mathcal{V}_{h+1}$, thus $E_2$ (\cref{eq:goodPsi}) holds. $\hat{V}_{h+1}^{k,i}$ has the correct functional form. It remains to show that its parameters are within the range of $\mathcal{V}_{h+1}$. First,
    \begin{align*}
        \norm{w_h^{k,i}}
        &
        =
        \norm*{
        \thetaKH + \zh^{k,i}
        +
        \covKH^{-1} \sum_{\tau = 1}^{k-1} \phi_h^\tau \hat{V}^{k,i}_{h+1}(x_{h+1}^\tau)
        }
        \\
        &
        \le
        \norm*{(\covK)^{-1}} \sum_{\tau = 1}^{k-1} \norm{\phi^\tau} \abs{v^\tau}
        +
        \norm{\z^{k,i}}
        +
        \norm*{(\covKH)^{-1}} \sum_{\tau = 1}^{k-1} \norm{\phi_h^\tau} 
        \abs{\hat{V}^{k,i}_{h+1}(x_{h+1}^\tau)}
        \\
        \tag{$\norm{\phi_h^\tau} \le 1, \covKH \succeq I, \covK \succeq H I, \norm{\hat{V}^{k,i}_{h+1}}_\infty \le H \betaRZ$}
        &
        \le
        H K
        +
        (\norm{\z^{k,i}}_{\covK} / \sqrt{H})
        +
        K H \betaRZ
        \\
        \tag{\cref{eq:goodZetaConcentration}}
        &
        \le
        H K
        +
        (\betaR \betaZ / \sqrt{H})
        +
        K H \betaRZ
        \\
        &
        \le
        4 K H \betaRZ
        ,
    \end{align*}
    where the last transition is due to our parameter choices.
    Finally, we have that
    \begin{align*}
        \betaP
        &
        =
        20 \betaRZ d H \sqrt{
        \log \brk*{163 K d H / \delta}
        }
        \\
        \tag{$\log(x) \le x$}
        &
        \le
        20 \betaRZ d H \sqrt{
        2K d + \log \brk*{82 H / \delta}
        }
        \\
        \tag{$a+b \le ab, \forall a,b \ge 2$}
        &
        \le
        20 \betaRZ d H \sqrt{
        2K d \log \brk*{82 H / \delta}
        }
        \\
        \tag{$H,d,K \ge 1$}
        &
        \le
        20 \betaRZ d^{7/4} H K \sqrt{
        8 \log \brk*{4 H / \delta}
        }
        \\
        &
        \le
        876 \betaRZ d^{7/4} H K \log (5H^2 / \delta)
        \\
        &
        =
        \bar{\beta}
        ,
    \end{align*}
    as desired.
\end{proof}

\newpage
\section{Randomized Ensemble Policy Optimization (REPO) with Reward-Free Warm-Up}
\label{appendix-sec:REPO}

\subsection{Proof of \cref{thm:REPO-regret}}

We begin by defining a so-called ``good event'', followed by optimism, cost of optimism, Ensemble Hedging cost, and Policy Optimization cost. We conclude with the proof of \cref{thm:REPO-regret}.

\paragraph{Good event.}
Define the truncated features $\bar{\phi}_h(x,a) = \indEvent{x \in \Z_h}\phi(x,a)$ and their concatenation $\bar{\phi}(x_{1:H},a_{1:H}) = (\bar{\phi}_1(x_1, a_1)\tran, \ldots, \bar{\phi}_H(x_H, a_H)\tran)\tran$. We also define the expected truncated feature occupancy of a policy $\pi$ as $\bar{\phi}^{\pi} = \EE[P, \pi] \bar{\phi}(x_{1:H}, a_{1:H})$.
In addition, to simplify presentation, we denote $\zeta^{k,i} = \zr^{k,i} + \zp^{k,i}$ where $\zr^{k,i} \sim \mathcal{N}\brk*{0, 2\betaR^2 {(\covK)}^{-1}}$ and $\zp^{k,i} \sim \mathcal{N}\brk*{0, 2 H \betaP^2 \mathrm{diag}\brk*{\covKH[1], \ldots, \covKH[H]}^{-1}}$.
We define the following good event $\Egood = \bigcap_{i=1}^7 E_i$, over which the regret is deterministically bounded:
\begin{flalign}
    \label{eq:goodRewardPOLinear}
     &
     E_1 = \brk[c]*{\forall e \in \setEpochs, k \in \setEpochsE : \norm{\theta - \widehat{\theta}^k}_{\covK} \le \betaR};
     &
     \\
     \label{eq:goodTransitionPOLinear}
     &
     E_2 = \brk[c]*{\forall e \in \setEpochs, k \in \setEpochsE , i \in [m] , h \in [H] : \norm{(\psiH - \psiKH) \hat{V}_{h+1}^{k,i}}_{\covKH} \le \betaP, \norm{\hat{Q}_{h+1}^{k,i}}_\infty \le \betaQ};
     &
     \\
     \label{eq:goodZetaConcentrationPOLinear}
     &
     E_3 = \brk[c]*{\forall e \in \setEpochs, i \in [m], h\in [H] : \norm{\zr^{\kEpoch,i}}_{\hatCovK} \le \betaZ\betaR, \norm{\zph^{\kEpoch,i}}_{\hatCovKH} \le \betaZ\betaP};
     &
     \\
     \label{eq:goodZetaAntiConcentrationPOLinear}
     &
     E_4 = \brk[c]*{\forall e \in \setEpochs : \max_{i \in [m]} (\bar{\phi}^{\piOpt})\tran \z^{\kEpoch,i} \ge \norm{\bar{\phi}^{\piOpt}}_{\covZ}};
     &
     \\
     \label{eq:goodWarmupPOLinear}
     &
     E_5 = \brk[c]*{\forall \pi \in \piClass, h \in [H] : \PR[P, \pi] \brk[s]*{  x_h \not\in \Z_h } \le \epsCov }; 
     &
    \\
    \label{eq:goodBernsteinPOLinear}
    &
    E_6 
    =
    \brk[c]*{
    \sum_{k=\firstEpisodeAfterWarmup}^{K} \EE[P, \piK] \brk[s]*{Y_k} \le \sum_{k=\firstEpisodeAfterWarmup}^{K} Y_k 
    +
    (1+ \sqrt{2}\betaZ)(\betaR + H \betaP) \sqrt{2K \log\frac{7}{\delta}}
    }.
    &
    \\
    \label{eq:goodAzumaPOLinear}
    &
    E_7 
    =
    \brk[c]*{
    \sum_{k=\firstEpisodeAfterWarmup}^{K} \sum_{i=1}^m p^k(i) \hat{V}^{k,i}_1(x_1)
    \le \sum_{k=\firstEpisodeAfterWarmup}^{K} \hat{V}^{k,i_k}_1(x_1) + 2 \betaQ  \sqrt{2K \log\frac{7}{\delta}}
    };
    &
\end{flalign}
where
$
    Y_k
    =
    (1+\sqrt{2}\betaZ) \brk[s]*{\betaR\norm{\phi^k}_{(\covK)^{-1}} 
    +
    \betaP \sum_{h \in [H]} \norm{\phi_h^k}_{(\covKH)^{-1}}}
    .
$

\begin{lemma}[Good event]
\label{lemma:good-event-PO-linear}
    Consider the following parameter setting:
    \begin{gather*}
        \lambdaP = 1
        ,
        \lambdaR = H
        ,
        m = 9 \log (7K / \delta)
        ,
        \betaR
        =
        2 H \sqrt{2dH \log (14K/\delta)}
        ,
        \betaZ
        =
        \sqrt{11 d H \log (14 m H K/\delta)}
        ,
        \\
        \betaP
        =
        216 \betaZ \betaR \sqrt{d H \log \frac{60 K^3 H \betaQ}{\delta \sqrt{d}}}
        ,
        \betaQ
        =
        18 \betaZ \betaR \sqrt{H}
        ,
        \betaWarmup 
        =
        36 \betaZ \sqrt{d \log \frac{60 K^3 H \betaQ}{\delta \sqrt{d}}}
        ,
        \etaO \le 1
        ,
        \epsCov \ge 1/K
        .
    \end{gather*}
    Then $\Pr[\Egood] \ge 1 - \delta.$
\end{lemma}
Proof in \cref{sec:good-event-proofs-PO-linear}.


\paragraph{Optimism and its cost.}
We start with a result that bounds the cost of transitioning to the truncated state-action features.
\begin{lemma}
\label{lemma:cost-of-truncated-features}
    Suppose that the good event $\Egood$ holds (\cref{eq:goodRewardPOLinear,eq:goodTransitionPOLinear,eq:goodZetaConcentrationPOLinear,eq:goodZetaAntiConcentrationPOLinear,eq:goodWarmupPOLinear,eq:goodBernsteinPOLinear,eq:goodAzumaPOLinear}), then for all $v \in \RR[d], h \in [H], \pi \in \Pi_M$, we have
    \begin{align*}
        \abs*{
        \EE[P, \pi]\brk[s]*{ 
        (\phi(x_h,a_h) - \bar{\phi}_h(x_h,a_h))\tran v}
        }
        \le
        C \epsCov
        ,
    \end{align*}
    where $C = \max_{x,a} \abs{\phi(x,a)\tran v}$. Additionally, if $v = \psiH \hat{V}_{h+1}^{k,i}$ then $C \le \betaQ$, and if $v = \thetaKH + \zrh^{\kEpoch,i}$ then $C \le 2\betaR\betaZ / \sqrt{H}$.
\end{lemma}
\begin{proof}
    We have that
    \begin{align*}
        \abs*{
        \EE[P, \pi]\brk[s]*{ 
        (\phi(x_h,a_h) - \bar{\phi}_h(x_h,a_h))\tran v}
        }
        &
        =
        \abs*{
        \EE[P, \pi]\brk[s]*{ 
        (1 - \indEvent{x_h \in \Z_h})\phi(x_h,a_h)\tran v}
        }
        \le
        C
        \PR[P,\pi]\brk[s]*{ 
        x_h \notin \Z_h}
        \le
        C \epsCov
        ,
    \end{align*}
    where the last inequality is by \cref{eq:goodWarmupPOLinear}.
    Now, if $v = \psiH \hat{V}_{h+1}^{k,i}$ then
    \begin{align*}
        C = \max_{x,a} \abs{\phi(x,a)\tran \psiH \hat{V}_{h+1}^{k,i}}
        \le
        \max_{x,a} \norm{\phi(x,a)\tran \psiH}_1 \norm{\hat{V}_{h+1}^{k,i}}
        =
        \norm{\hat{V}_{h+1}^{k,i}}
        \le
        \betaQ
        ,
    \end{align*}
    where the last equality used that $\phi(x,a)\tran \psiH$ is a distribution over $\X$, thus has $\ell_1-$norm of $1$ and the last inequality used \cref{eq:goodTransitionPOLinear}.
    Finally, if $v = \thetaKH + \zrh^{\kEpoch,i}$ then using Cauchy-Schwarz
    \begin{align*}
        C
        \le
        \max_{x,a} \norm{\phi(x,a)}
        \norm{\thetaKH + \zrh^{\kEpoch,i}}
        \le
        \norm{\thetaH}
        +
        \norm{\thetaK - \theta}
        +
        \norm{\zr^{\kEpoch,i}}
        \le
        \sqrt{d}
        +
        \frac{\betaR(1+\betaZ)}{\sqrt{H}}
        \le
        \frac{2\betaR\betaZ}{\sqrt{H}}
        ,
    \end{align*}
    where the second inequality used that $\norm{\phi(x,a)}\le 1$ and the triangle-inequality, the third used $\covK \succeq H I, \norm{\thetaH} \le \sqrt{d}$ and \cref{eq:goodRewardPOLinear,eq:goodZetaConcentrationPOLinear}, and the fourth used $\betaR \ge \sqrt{dH}, \betaZ \ge 3$.
\end{proof}

\begin{lemma}[Optimism]
\label{lemma:optimism-PO-linear}
    Suppose that the good event $\Egood$ holds (\cref{eq:goodRewardPOLinear,eq:goodTransitionPOLinear,eq:goodZetaConcentrationPOLinear,eq:goodZetaAntiConcentrationPOLinear,eq:goodWarmupPOLinear,eq:goodBernsteinPOLinear,eq:goodAzumaPOLinear}) and define $\hatJE = \argmax_{i \in [m]} \brk{\bar{\phi}^{\piOpt}}\tran \z^{\kEpoch,i}$, $e \in \setEpochs$, then for all $e \in \setEpochs, k \in \setEpochsE$
    \begin{align*}
        \sum_{h \in [H]} \EE[P, \piOpt]\brk[s]*{ \phi(x_h,a_h)\tran \brk{ \thetaH + \psiH \hat{V}^{k,\hatJE}_{h+1}} - \hat{Q}^{k,\hatJE}_h(x_h,a_h)}
        \le
         \frac{9}{8}\epsCov H \betaQ
        \quad
        ,
        \forall k \in [K]
        .
    \end{align*}
\end{lemma}

\begin{proof}
    Define $\bar{\phi}_h^\pi = \EE[P, \pi] \bar{\phi}_h(x_h,a_h)$ and recall that $\bar{\phi}^\pi = ((\bar{\phi}^\pi_1)\tran, \ldots, (\bar{\phi}^\pi_H)\tran)\tran$
    Then we have that
    \begin{align*}
        &
        \sum_{h \in [H]} \EE[P, \piOpt]\brk[s]*{\phi(x_h,a_h)\tran \brk{ \thetaH + \psiH \hat{V}^{k,\hatJE}_{h+1} } - \hat{Q}^{k,\hatJE}_h(x_h,a_h)}
        \\
        \tag{\cref{lemma:cost-of-truncated-features}}
        &
        \le
        \frac{9}{8}\epsCov H \betaQ
        +
        \sum_{h \in [H]} \EE[P, \piOpt]\brk[s]*{\bar{\phi}_h(x_h,a_h)\tran \brk{ \thetaH + \psiH \hat{V}^{k,\hatJE}_{h+1} } - \hat{Q}^{k,\hatJE}_h(x_h,a_h)}
        \\
        &
        =
        \frac{9}{8}\epsCov H \betaQ
        +
        \sum_{h \in [H]}
        \EE[P, \piOpt]\brk[s]*{\bar{\phi}_h(x_h,a_h)}\tran \brk*{ \thetaH - \thetaKH - \zh^{\kEpoch,\hatJE} + (\psiH -\psiKH)\hat{V}^{k,\hatJE}_{h+1} }
        \\
        \tag{\cref{eq:goodRewardPOLinear,eq:goodTransitionPOLinear}, $\hatCovK \preceq \covK, \hatCovKH \preceq \covKH$}
        &
        \le
        \frac{9}{8}\epsCov H \betaQ
        +
        \betaR \norm{\bar{\phi}^{\piOpt}}_{(\hatCovK)^{-1}}
        +
        \sum_{h \in [H]} \betaP \norm{\bar{\phi}^{\piOpt}_{h}}_{(\hatCovKH)^{-1}}
        -
        \brk{\bar{\phi}^{\piOpt}}\tran \z^{\kEpoch,\hatJE}
        \\
        \tag{Cauchy-Schwarz}
        &
        \le
        \frac{9}{8}\epsCov H \betaQ
        +
        \norm{\bar{\phi}^{\piOpt}}_{\betaR^2(\hatCovK)^{-1}}
        +
        \norm{\bar{\phi}^{\piOpt}}_{{H \betaP^2 \mathrm{diag}\brk*{\hatCovKH[1], \ldots, \hatCovKH[H]}^{-1}}}
        -
        \brk{\bar{\phi}^{\piOpt}}\tran \z^{\kEpoch,\hatJE}
        \\
        \tag{Cauchy-Schwarz}
        &
        \le
        \frac{9}{8}\epsCov H \betaQ
        +
        \norm{\bar{\phi}^{\piOpt}}_{2\betaR^2 (\hatCovK)^{-1} + 2 H \betaP^2 \mathrm{diag}\brk*{\hatCovKH[1], \ldots, \hatCovKH[H]}^{-1}}
        -
        \brk{\bar{\phi}^{\piOpt}}\tran \z^{\kEpoch,\hatJE}
        \\
        &
        =
        \frac{9}{8}\epsCov H \betaQ
        +
        \norm{\bar{\phi}^{\piOpt}}_{\covZ}
        -
        \brk{\bar{\phi}^{\piOpt}}\tran \z^{\kEpoch,\hatJE}
        \\
        \tag{\cref{eq:goodZetaAntiConcentrationPOLinear}}
        &
        \le
        \frac{9}{8}\epsCov H \betaQ
        ,
    \end{align*}
    as desired.
\end{proof}

\begin{lemma}[Cost of optimism]
\label{lemma:cost-of-optimism-PO-linear}
    Suppose that the good event $\Egood$ holds (\cref{eq:goodRewardPOLinear,eq:goodTransitionPOLinear,eq:goodZetaConcentrationPOLinear,eq:goodZetaAntiConcentrationPOLinear,eq:goodWarmupPOLinear,eq:goodBernsteinPOLinear,eq:goodAzumaPOLinear}), then $\forall e \in \setEpochs  ,k \in \setEpochsE$
    \begin{align*}
        \hat{V}^{k, i_k}_1(x_1)
        -
        V_1^{\piK}(x_1)
        \le
        \frac{9}{8}\epsCov H \betaQ
        +
        (1+\sqrt{2}\betaZ) \EE[P, \piK]\brk[s]*{
        \betaR \norm{\phi(x_{1:H},a_{1:H})}_{(\covK)^{-1}} 
        +
        \betaP \sum_{h \in [H]} \norm{\phi(x_h,a_h)}_{(\covKH)^{-1}}
        }
        .
    \end{align*}
\end{lemma}

\begin{proof}
    By \cref{lemma:extended-value-difference}, a value difference lemma by \citet{shani2020optimistic},
    \begin{align*}
        &
        \hat{V}^{k, i_k}_1(x_1)
        -
        V_1^{\piK}(x_1)
        =
        \sum_{h \in [H]} \EE[P, \piK]\brk[s]*{ \hat{Q}_k^{k,i_k}(x_h,a_h) -  \phi(x_h,a_h)\tran \brk*{ \thetaH  + \psiH\hat{V}_{h+1}^{k,i_k} } }
        \\
        &
        =
        \sum_{h \in [H]} \EE[P, \piK]\brk[s]*{ \bar{\phi}_h(x_h,a_h)\tran \brk*{\thetaKH + \zrh^{\kEpoch,i_k} + \zph^{\kEpoch,i_k} + \psiKH\hat{V}_{h+1}^{k,i_k} } -  \phi(x_h,a_h)\tran \brk*{ \thetaH  + \psiH\hat{V}_{h+1}^{k,i_k} } }
        \\
        \tag{\cref{lemma:cost-of-truncated-features}}
        &
        \le
        \frac{9}{8}\epsCov H \betaQ
        +
        \EE[P, \piK]\brk[s]*{
        \sum_{h \in [H]} 
        \phi(x_h,a_h)\tran \brk*{ \thetaKH - \thetaH  + \zrh^{\kEpoch,i_k}} 
        +
        \sum_{h \in [H]}\bar{\phi}_h(x_h,a_h)\tran \brk*{ \zph^{\kEpoch,i_k} + (\psiKH - \psiH)\hat{V}_{h+1}^{k,i_k} } 
        }
        \\
        &
        =
        \frac{9}{8}\epsCov H \betaQ
        +
        \EE[P, \piK]\brk[s]*{
        \underbrace{
        \phi(x_{1:H},a_{1:H})\tran(\thetaK - \theta + \zr^{\kEpoch,i_k})
        }_{(i)}
        +
        \sum_{h \in [H]} 
        \underbrace{
        \bar{\phi}_h(x_h,a_h)\tran \brk*{ \zph^{\kEpoch,i_k} + (\psiKH - \psiH)\hat{V}_{h+1}^{k,i_k} }
        }_{(ii)}
        }
        .
    \end{align*}
    We conclude the proof by bounding $(i), (ii)$.
    \begin{align*}
        (i)
        \tag{\cref{eq:goodRewardPOLinear,eq:goodZetaConcentrationPOLinear}}
        &
        \le
        \betaR \norm{\phi(x_{1:H},a_{1:H})}_{(\covK)^{-1}}
        +
        \betaR\betaZ\norm{\phi(x_{1:H},a_{1:H})}_{(\hatCovK)^{-1}}
        \\
        \tag{\cref{lemma:matrix-norm-inequality}}
        &
        \le
        \betaR (1+\sqrt{2}\betaZ) \norm{\phi(x_{1:H},a_{1:H})}_{(\covK)^{-1}}
        .
    \end{align*}
    \begin{align*}
        (ii)
        \tag{\cref{eq:goodTransitionPOLinear,eq:goodZetaConcentrationPOLinear}}
        &
        \le
        \betaP \norm{\bar{\phi}_h(x_h,a_h)}_{(\covKH)^{-1}}
        +
        \betaP\betaZ \norm{\bar{\phi}_h(x_h,a_h)}_{(\hatCovKH)^{-1}}
        \\
        \tag{\cref{lemma:matrix-norm-inequality}}
        &
        \le
        \betaP(1+\sqrt{2}\betaZ)\norm{\bar{\phi}_h(x_h,a_h)}_{(\covKH)^{-1}}
        \\
        &
        \le
        \betaP(1+\sqrt{2}\betaZ)\norm{\phi_h(x_h,a_h)}_{(\covKH)^{-1}}
        ,
    \end{align*}
    where the last inequality used that for $x_h \in \Z_h,$ $\phi(x_h, a_h) = \bar{\phi}_h(x_h, a_h)$ and for $x_h \notin \Z_h$ $\norm{\bar{\phi}_h(x_h,a_h)}_{(\covKH)^{-1}} = 0 \le \norm{\phi(x_h,a_h)}_{(\covKH)^{-1}}$.
\end{proof}

\paragraph{Hedge over the ensemble.}
We use standard online mirror arguments to bound the regret against each single $j \in [m]$ in every epoch $e \in \setEpochs$.
\begin{lemma}[Hedge]
\label{lemma:hedge-term-PO-linear}
    For every epoch $e \in \setEpochs$, let $j_e \in [m]$.
    Suppose that the good event $\Egood$ holds (\cref{eq:goodRewardPOLinear,eq:goodTransitionPOLinear,eq:goodZetaConcentrationPOLinear,eq:goodZetaAntiConcentrationPOLinear,eq:goodWarmupPOLinear,eq:goodBernsteinPOLinear,eq:goodAzumaPOLinear}) and set $\etaH \le 1/\betaQ$, then
    \begin{align*}
        \sum_{e \in \setEpochs} \sum_{k \in \setEpochsE} \hat{V}^{k,j_e}_1(x_1) -  \hat{V}^{k,i_k}_1(x_1)
        &
        \le
        \frac{\numEpochs \log m}{\etaH} + \etaH K \betaQ^2 + 2 \betaQ  \sqrt{2K \log\frac{7}{\delta}}
        .
    \end{align*}
\end{lemma}

\begin{proof}
    Note that the distribution $p^k(\cdot)$ is reset in the beginning of every epoch.
    Then, the lemma follows by first using \cref{eq:goodAzumaPOLinear} and then applying \cref{lem:omd-regret} for each epoch individually with $y_t(a) = - \hat{V}^{k,i}_1(x_1) , x_t(a)  = p^{k}(i)$ and noting that $| \hat{V}^{k,i}_1(x_1) | \le \betaQ$ by \cref{eq:goodTransitionPOLinear}.
\end{proof}

\paragraph{Policy online mirror descent.}
We use standard online mirror descent arguments to bound the local regret in each state.
\begin{lemma}[OMD]
\label{lemma:omd-term-PO-linear}
    Suppose that the good event $\Egood$ holds (\cref{eq:goodRewardPOLinear,eq:goodTransitionPOLinear,eq:goodZetaConcentrationPOLinear,eq:goodZetaAntiConcentrationPOLinear,eq:goodWarmupPOLinear,eq:goodBernsteinPOLinear,eq:goodAzumaPOLinear}) and set $\etaO \le 1/\betaQ$, then 
    \begin{align*}
        \sum_{k \in \setEpochsE} \sum_{a \in \A} \hat{Q}_h^{k,i}(x,a) (\pi^\star_h(a \mid x) - \pi^{k,i}_h(a \mid x))
        &
        \le
        \frac{\log \abs{\A}}{\etaO} + \etaO \sum_{k \in \setEpochsE} \betaQ^2
        \quad
        ,
        \forall e \in \setEpochs, i \in [m], h \in [H], x \in \X
        .
    \end{align*}
\end{lemma}

\begin{proof}
    Note that the policy $\pi^{k,i}$ is reset in the beginning of every epoch.
    Then, the lemma follows directly by \cref{lem:omd-regret} with $y_t(a) = - \hat{Q}_h^{k,i}(x,a) , x_t(a)  = \pi^{k,i}_h(a \mid x)$ and noting that $| \hat{Q}_h^{k,i}(x,a) | \le \betaQ$ by \cref{eq:goodTransitionPOLinear}.
\end{proof}

\paragraph{Epoch schedule.}
The algorithm operates in epochs.
At the beginning of each epoch, the noise is re-sampled and the policies are reset to be uniformly random.
We denote the total number of epochs by $\numEpochs$, the first episode within epoch $e$ by $\kEpoch$, and the set of episodes within epoch $e$ by $\setEpochsE$.
The following lemma bounds the number of epochs.

\begin{lemma}
\label{lem:num-epochs}
    The number of epochs $E$ is bounded by $3 d H \log (2K)$.
\end{lemma}

\begin{proof}
    Let $\mathcal{T}_h = \brk[c]{e_h^1, e_h^2, \ldots}$ be the epochs where the condition $\det(\covKH) \ge 2 \det(\hatCovKH)$ was triggered in \cref{line:repo-epoch-condition} of \cref{alg:r-opo-for-linear-mdp}. Then we have that
    \begin{align*}
        \det(\Lambda_h^{\kEpoch})
        \ge
        \begin{cases}
            2 \det(\Lambda_h^{\kPrevEpoch})&, e \in \mathcal{T}_h
            \\
            \det(\Lambda_h^{\kPrevEpoch})&, \text{otherwise}
            .
        \end{cases}
    \end{align*}
    Unrolling this relation, we get that
    \begin{align*}
    \det(\Lambda_h^K)
    \ge
    2^{\abs{\mathcal{T}_h}-1} \det{I}
    =
    2^{\abs{\mathcal{T}_h}-1}
    ,
    \end{align*}
    and changing sides, and taking the logarithm we get that
    \begin{align*}
    \abs{\mathcal{T}_h}
    &
    \le
    1
    +
    \log_2 \det\brk*{\Lambda_h^K}
    \\
    \tag{$\det(A) \le \norm{A}^d$}
    &
    \le
    1
    +
    d \log_2 \norm{\Lambda_h^K}
    \\
    \tag{triangle inequality}
    &
    \le
    1
    +
    d \log_2 \brk*{1 + \sum_{k=1}^{K-1} \norm{\phi_h^k}^2}
    \\
    \tag{$\norm{\phi_h^k} \le 1$}
    &
    \le
    1
    +
    d \log_2 K
    \\
    &
    \le
    (3/2) d \log 2 K
    .
\end{align*}
Similarly, if $\bar{\mathcal{T}}$ are the epochs where the condition $\det(\covK) \ge 2 \det(\hatCovK)$ was triggered, the $\abs{\bar{\mathcal{T}}} \le (3/2) d H \log 2K$. We conclude that
\begin{align*}
    E
    =
    \abs{
    \bar{\mathcal{T}}
    \cup
    \brk*{
    \cup_{h \in [H]} {\mathcal{T}_h}
    }}
    \le
    \abs{\bar{\mathcal{T}}}
    +
    \sum_{h \in [H]} \abs{\mathcal{T}_h}
    \le
    3 d H \log (2K)
    .
    &
    \qedhere
\end{align*}
\end{proof}

\paragraph{Regret bound.}
\begin{theorem}[restatement of \cref{thm:REPO-regret}]
\label{thm:regret-bound-PO-linear}
    Suppose that we run \cref{alg:r-opo-for-linear-mdp} with
    \begin{align*}
        \etaH
        =
        \sqrt{\frac{3 d H \log (2K) \log m}{K \betaQ^2}}
        ,
        \etaO
        =
        \sqrt{\frac{3 d H \log (2K) \log \abs{\A}}{K \betaQ^2}}
        ,
        \epsCov
        =
        \frac{2 d^2 H^2} {3 \sqrt{C \betaQ K}}\log^4 \brk*{\frac{28 H^2 K \betaWarmup^2}{\delta}}
        ,
    \end{align*}
    where $C > 0$ is a universal constant defined in \cref{lemma:reward-free-exploration},
    and the other parameters as in \cref{lemma:good-event-PO-linear}. Then with probability at least $1 - \delta$ we incur regret at most
    \begin{align*}
        \regret
        \le
        39
        \sqrt{C  d^5 H^9 K  \log^9 \brk*{\frac{28 H^2 K \betaWarmup^2}{\delta}}}
        +
        67204 \sqrt{d^5 H^8 K \log^5 \frac{60 K^3 \abs{\A} H m \betaQ}{\delta}} 
        =
        \tilde{O}\brk*{
        \sqrt{d^5 H^9 K \log^9\frac{\abs{\A}}{\delta}}
        }
        .
    \end{align*}
\end{theorem}
\begin{proof}
    Suppose that the good event $\Egood$ holds (\cref{eq:goodRewardPOLinear,eq:goodTransitionPOLinear,eq:goodZetaConcentrationPOLinear,eq:goodZetaAntiConcentrationPOLinear,eq:goodWarmupPOLinear,eq:goodBernsteinPOLinear,eq:goodAzumaPOLinear}). By \cref{lemma:good-event-PO-linear}, this holds with probability at least $1-\delta$.
    For every epoch $e \in \setEpochs$, let $\hatJE = \argmax_{i \in [m]} (\bar{\phi}^{\piOpt})\tran \z^{\kEpoch,i}$.
    Now decompose the regret using \cref{lemma:extended-value-difference}, an extended value difference lemma by \citet{shani2020optimistic}.
    \begin{align*}
        \regret
        & =
        \sum_{k \in [K]} V^{\piOpt}_1(x_1) - V^{\pi^k}_1(x_1)
        \\
        & \le
        H \firstEpisodeAfterWarmup + \sum_{k=\firstEpisodeAfterWarmup}^K V^{\piOpt}_1(x_1) - V^{\pi^k}_1(x_1)
        \\
        & =
        H \firstEpisodeAfterWarmup + \underbrace{\sum_{k=\firstEpisodeAfterWarmup}^K \hat{V}^{k,i_k}_1(x_1) - V^{\pi^k}_1(x_1)}_{(i)} 
        +
        \underbrace{\sum_{e \in \setEpochs} \sum_{k \in \setEpochsE} \hat{V}^{k,\hatJE}_1(x_1) -  \hat{V}^{k,i_k}_1(x_1)}_{(ii)}
        \\
        & \qquad + 
        \underbrace{\sum_{e \in \setEpochs} \sum_{k \in \setEpochsE} \sum_{h \in [H]} \sum_{a \in \A} \EE[P, \piOpt]\brk[s]*{ \hat{Q}^{k,\hatJE}_h(x_h,a) \brk{ \pi^\star_h(a \mid x_h) - \pi^{k,\hatJE}_h(a \mid x_h) } } }_{(iii)}
        \\
        & \qquad +
        \underbrace{\sum_{e \in \setEpochs} \sum_{k \in \setEpochsE} \sum_{h \in [H]} \EE[P, \piOpt]\brk[s]*{ \phi(x_h,a_h)\tran \brk{ \thetaH + \psiH \hat{V}^{k,\hatJE}_{h+1} } - \hat{Q}^{k,\hatJE}_h(x_h,a_h) } }_{(iv)}
        .
    \end{align*}
    By \cref{lemma:reward-free-exploration}, for $\betaWarmup = \tilde{O}(d\sqrt{H})$ and $\epsCov \ge 1/K$, we have that $\firstEpisodeAfterWarmup \le C \cdot \brk*{
    \frac{d^4 H^4}{\epsCov} \log^8 \brk*{\frac{28 H^2 K \betaWarmup^2}{\delta}}
    }
    $.
    By \cref{lemma:optimism-PO-linear} $(iv) \le \frac{9}{8}\epsCov H \betaQ K$.
    By \cref{lemma:hedge-term-PO-linear,lem:num-epochs} (with our choice of $\etaH$), we have that 
    \begin{align*}
        (ii)
        \le
        2 \betaQ \sqrt{3 K d H \log (m) \log(2K)} + 2 \betaQ  \sqrt{2K \log\frac{7}{\delta}}
        \le
        8 \betaQ \sqrt{K d H} \log \brk*{\frac{7 K m}{\delta}}
        .
    \end{align*}
    Similarly, by \cref{lemma:omd-term-PO-linear,lem:num-epochs} (with our choice of $\etaO$) we have
    \begin{align*}
        (iii)
        \le
        \sum_{h \in [H]} \sum_{e \in \setEpochs} \EE[P, \piOpt]\brk[s]*{ \frac{\log A}{\etaO} + \etaO \sum_{k \in \setEpochsE} \betaQ^2 }
        \le
        4 H \betaQ \sqrt{K d H \log (2K) \log \abs{\A}}
        .
    \end{align*}
    For term $(i)$, we use \cref{lemma:cost-of-optimism-PO-linear} as follows.
    \begin{align*}
        (i)
        & 
        \le
        \frac{9}{8}\epsCov H \betaQ K
        +
        (1+\sqrt{2}\betaZ)\sum_{k \in [K]} \EE[P, \piK]\brk[s]*{
        \betaR \norm{\phi(x_{1:H},a_{1:H})}_{(\covK)^{-1}} 
        +
        \betaP \sum_{h \in [H]} \norm{\phi(x_h,a_h)}_{(\covKH)^{-1}}
        }
        \\
        \tag{\cref{eq:goodBernsteinPOLinear}}
        &
        \le 
        \frac{9}{8}\epsCov H \betaQ K
        +
        \betaR(1+\sqrt{2}\betaZ) \sum_{k \in [K]} \norm{\phi^k}_{(\covK)^{-1}} 
        +
        \betaP(1+\sqrt{2}\betaZ) \sum_{h \in [H]} \sum_{k \in [K]} \norm{\phi^k_h}_{(\covKH)^{-1}}
        \\
        & \qquad
        +
        (1+\sqrt{2}\betaZ)(\betaR + H \betaP) \sqrt{2 K \log\frac{7}{\delta}}
        \\
        \tag{\cref{lemma:elliptical-potential}}
        &
        \le
        \frac{9}{8}\epsCov H \betaQ K
        +
        (1+\sqrt{2}\betaZ)
        \brk{
        \betaR \sqrt{H}
        +
        \betaP H } \sqrt{ 2 K d \log(2K) }
        +
        (1+\sqrt{2}\betaZ)(\betaR + H \betaP) \sqrt{2 K \log\frac{7}{\delta}}
        \\
        \tag{$\betaZ \ge 3, \betaP \ge 216 \betaR$}
        &
        \le
        \frac{9}{8}\epsCov H \betaQ K
        +
        8 \betaP \betaZ H \sqrt{K d \log\frac{7K}{\delta}}
        .
    \end{align*}
    Putting all bounds together, we get that
    \begin{align*}
        \regret
        &
        \le
        C \cdot \brk*{
        \frac{d^4 H^5}{\epsCov} \log^8 \brk*{\frac{28 H^2 K \betaWarmup^2}{\delta}}
        }
        +
        \frac{9}{4}\epsCov H \betaQ K
        +
        8 \betaP \betaZ H \sqrt{K d \log\frac{7K}{\delta}}
        \\
        &
        +
        8 \betaQ \sqrt{K d H} \log \brk*{\frac{7 K m}{\delta}}
        +
        4 H \betaQ \sqrt{K d H \log (2K) \log \abs{\A}}
        \\
        &
        \le
        3 d^2 H^3 \log^4 \brk*{\frac{28 H^2 K \betaWarmup^2}{\delta}}
        \sqrt{C \betaQ K}
        +
        12 H \betaQ \sqrt{K d H} \log \brk*{\frac{7 K \abs{\A} m}{\delta}}
        \\
        &
        +
        108 \betaQ \betaZ H d \sqrt{K}
        \log \frac{60 K^3 H \betaQ}{\delta \sqrt{d}}
        \\
        &
        \le
        3 d^2 H^3 \log^4 \brk*{\frac{28 H^2 K \betaWarmup^2}{\delta}}
        \sqrt{C \betaQ K}
        +
        120 \betaQ \betaZ H d \sqrt{K}
        \log \frac{60 K^3 \abs{\A} H m \betaQ}{\delta}
        \\
        &
        \le
        3 d^2 H^3 \log^4 \brk*{\frac{28 H^2 K \betaWarmup^2}{\delta}}
        \sqrt{18 C \betaR \betaZ H K}
        +
        2160 \betaR \betaZ^2 d \sqrt{H^3 K}
        \log \frac{60 K^3 \abs{\A} H m \betaQ}{\delta}
        \\
        &
        \le
        39
        \sqrt{C  d^5 H^9 K  \log^9 \brk*{\frac{28 H^2 K \betaWarmup^2}{\delta}}}
        +
        67204 \sqrt{d^5 H^8 K \log^5 \frac{60 K^3 \abs{\A} H m \betaQ}{\delta}} 
        ,
    \end{align*}
    where the last transition also used that
    \begin{align*}
        \betaR\betaZ^2
        \le
        22 \sqrt{2 d^3 H^5 \log^3 (14m H K/\delta)}
        ,
        \betaR\betaZ
        \le
        \sqrt{88} d H^2  \log (14 m H K/\delta)
        .
        &
        \qedhere
    \end{align*}
\end{proof}

\subsection{Proofs of good event (REPO)}
\label{sec:good-event-proofs-PO-linear}

We begin by defining function classes and properties necessary for the uniform convergence arguments over the value functions. We then proceed to define a proxy good event, whose high probability occurrence is straightforward to prove. We then show that the proxy event implies the desired good event.

\paragraph{Value and policy classes.}
We define the following class of restricted Q-functions:
\begin{align*}
    \widehat{\Q}(\mathcal{Z}, W, C)
    =
    \brk[c]*{ \hat{Q}(\cdot,\cdot ; w,\mathcal{Z}) \mid \norm{w} \le W , \norm{ \hat{Q}(\cdot,\cdot ; w,\mathcal{Z}) }_\infty \le C }
    ,
\end{align*}
where $\hat{Q}(x,a ; w,\mathcal{Z}) = w\tran \phi(x,a) \cdot \indEvent{x \in \mathcal{Z}}$.
Next, we define the following class of soft-max policies:
\begin{align*}
    \Pi (\mathcal{Z}, W)
    =
    \brk[c]*{ \pi(\cdot \mid \cdot ; w, \mathcal{Z}) \mid \norm{w} \le W }
    ,
\end{align*}
where $\pi(a \mid x ; w, \mathcal{Z}) = \frac{\exp \brk{\hat{Q}(x,a ; w,\mathcal{Z})}}{\sum_{a' \in \A} \exp \brk{\hat{Q}(x,a' ; w,\mathcal{Z})}}$.
Finally, we define the following class of restricted value functions:
\begin{align}
\label{eq:value-function-class-def}
    \widehat{\V}(\mathcal{Z}, W, C)
    =
    \brk[c]*{ \hat{V}(\cdot ; \pi,\hat{Q}) \mid \pi \in \Pi(\mathcal{Z}, W K) , \hat{Q} \in \widehat{\Q}(\mathcal{Z}, W, C)}
    ,
\end{align}
where $\hat{V}(x ; \pi,\hat{Q}) = \sum_{a \in \A} \pi(a \mid x) \hat{Q}(x,a)$.
The following lemma provides the bound on the covering number of the value function class defined above (see proof in \cref{sec:good-event-proofs-PO-linear}).
\begin{lemma}
\label{lemma:function-class-covering-number}
    For any $\epsilon, W > 0, C \ge 1$ and $\mathcal{Z} \subseteq \X$ we have $\log \mathcal{N}_\epsilon \brk*{ \widehat{\V}(\mathcal{Z}, W, C) } \le 2 d \log (1 + 7 W C K / \epsilon)$ where $\mathcal{N}_\epsilon$ is the covering number of a class in supremum distance.
\end{lemma}

\begin{proof}
    First, notice that our policy class $\Pi(\mathcal{Z}, W)$ fits Lemma 12 of \cite{sherman2023rate} with $f_\theta(y) = y\tran \theta \cdot \indEvent{y \in \mathcal{Z}}$. Since $\norm{y} = \norm{\phi(x,a)} \le 1$, $f_\theta$ is $1-$Lipschitz with respect to $\theta$ and thus the policy class is $2-$Lipschitz, in $\ell_1-$norm, i.e.,
    \begin{align*}
        \norm{
        \pi(\cdot \mid x; w)
        -
        \pi(\cdot \mid x; w')
        }_1
        \le
        2 \norm{w - w'}
        .
    \end{align*}
    Similarly, $\widehat{\Q}(\mathcal{Z}, W, C)$ is $1-$Lipschitz in supremum norm, i.e.,
    \begin{align*}
        \norm{\hat{Q}(\cdot, \cdot; w) - \hat{Q}'(\cdot, \cdot; w')}_\infty
        \le
        \norm{w - w'}
        .
    \end{align*}
    Now, let $V, V' \in \widehat{\V}(\mathcal{Z}, W, C)$ and $w=(w_1,w_2), w'=(w_1', w_2') \in \RR[2d]$ be their respective parameters. We have that
    \begin{align*}
        \abs{V(x; \pi, \hat{Q}) - V(x; \pi', \hat{Q}')}
        \le
        \underbrace{
        \abs{V(x; \pi, \hat{Q}) - V(x; \pi, \hat{Q}')}
        }_{(i)}
        +
        \underbrace{
        \abs{V(x; \pi, \hat{Q}') - V(x; \pi', \hat{Q}')}
        }_{(ii)}
        .
    \end{align*}
    For the first term
    \begin{align*}
        (i)
        &
        =
        \abs*{\sum_{a \in \A} \pi(a\mid x) (\hat{Q}(x,a; w_2) - \hat{Q}(x,a; w_2'))}
        \\
        \tag{triangle inequality}
        &
        \le
        \sum_{a \in \A} \pi(a\mid x) \abs*{\hat{Q}(x,a; w_2) - \hat{Q}(x,a; w_2')}
        \\
        \tag{$\hat{Q}$ is $1$-Lipschitz}
        &
        \le
        \sum_{a \in \A} \pi(a\mid x) \norm{w_2 - w_2'}
        \\
        &
        =
        \norm{w_2 - w_2'}
        .
    \end{align*}
    For the second term
    \begin{align*}
        (ii)
        =
        \abs*{\sum_{a \in \A} \hat{Q}'(x,a) (\pi(a \mid x; w_1) - \pi(a \mid x; w_1')}
        \le
        C \norm{\pi(\cdot \mid x; w_1) - \pi(\cdot \mid x; w_1')}_1
        \le
        2C \norm{w_1 - w_1'}
        ,
    \end{align*}
    where the first transition used that $\norm{Q}_\infty \le C$ for all $Q \in \widehat{\Q}(\mathcal{Z}, W, C)$ and the second used the Lipschitz property of the policy class shown above.
    Combining the terms we get that
    \begin{align*}
        \abs{V(x; \pi, \hat{Q}) - V(x; \pi', \hat{Q}')}
        \le
        \norm{w_2 - w_2'}
        +
        2C \norm{w_1 - w_1'}
        \le
        \sqrt{1 + 4C^2} \norm{w - w'}
        ,
    \end{align*}
    implying that $\widehat{\V}(\mathcal{Z}, W, C)$ is $\sqrt{1 + 4C^2}-$Lipschitz in supremum norm.
    Finally, notice that $\norm{w} = \sqrt{\norm{w_1}^2 + \norm{w_2}^2} \le \sqrt{2} K W$. Applying \cref{lemma:lipschitz-cover} together with our assumption that $C \ge 1$ concludes the proof.
\end{proof}

\paragraph{Proxy good event.}
We first need the following result regarding the reward free exploration.
\begin{lemma}[Lemma 16 in \cite{sherman2023rate}]
\label{lemma:reward-free-exploration}
    Assume we run Algorithm 2 of \cite{sherman2023rate} with $\epsCov \ge 1/K$. Then it will terminate after
    $
    C \cdot \brk*{
    \frac{d H^3}{\epsCov}\max\brk[c]{\betaWarmup^2, d^3 H} \log^8 \brk*{\frac{28 H^2 K \betaWarmup^2}{\delta}}
    }
    $
    episodes where $C > 0$ is a numerical constant, and with probability at least $1 - \delta/7$, outputs $\covWarmupH, h \in [H]$ such that
    \begin{align*}
        \forall \pi \in \piClass, h \in [H] : \PR[P, \pi] \brk[s]*{  x_h \not\in \Z_h } \le \epsCov
    \end{align*}
    where
    $
    \mathcal{Z}_h
    =
    \brk[c]*{
    x \in \X \mid \forall a, \norm{\phi(x,a)}_{(\covWarmupH)^{-1}} \le 1 / (2 \betaWarmup H)
    }
    .
    $
\end{lemma}

Next, we define a proxy good event $\EgoodBar = E_1 \cap \bar{E}_2 \cap E_3 \cap E_4 \cap E_5 \cap E_6 \cap \bar{E}_7$ where
\begin{flalign}
    &
    \label{eq:goodTransitionProxPOLinear}
    \bar{E}_2 = \brk[c]*{\forall k \in [K], h \in [H], V \in \widehat{\V}(\mathcal{Z}_{h+1}, W, \betaQh[h+1]) : \norm{(\psiH - \psiKH) V}_{\covKH} \le \betaPh}
    &
    \\
    &
    \label{eq:goodAzumaProxPOLinear}
    \bar{E}_7
    =
    \brk[c]*{
    \sum_{k=\firstEpisodeAfterWarmup}^{K} \sum_{i=1}^m p^k(i) \clip_{\betaQ}\brk[s]*{\hat{V}^{k,i}_1(x_1)}
    \le
    \sum_{k=\firstEpisodeAfterWarmup}^{K} \clip_{\betaQ}\brk[s]*{\hat{V}^{k,i_k}_1(x_1)} + 2 \betaQ  \sqrt{2K \log\frac{7}{\delta}}
    }
    ,
    &
\end{flalign}
and $\betaQh, \betaPh, h \in [H]$ are such that $\betaPh = 12 \betaQh[h+1] \sqrt{d \log \frac{15 K^2 H W}{\delta \sqrt{d}}}, W = 4 K \betaQ$ and
\begin{align}
\label{eq:betaQh-def}
    \betaQh = \betaQh[H] + \brk*{1 + \frac{1}{H}}\betaQh[h+1],
    \quad
    \betaQh[H]
    =
    \frac{6 \betaZ \betaR}{\sqrt{H}}
    ,
    \quad
    \betaQh[H+1] = 0
    .
\end{align}
Then we have the following result.
\begin{lemma}[Proxy good event]
\label{lemma:good-proxy-event-PO-linear}
    Consider the parameter setting of \cref{lemma:good-event-PO-linear}. Then $\Pr[\EgoodBar] \ge 1 - \delta.$
\end{lemma}
\begin{proof}
    %
    First, by \cref{lemma:reward-error} and our choice of parameters, $E_1$ (\cref{eq:goodRewardPOLinear}) holds with probability at least $1 - \delta/7$.
    %
    Next, applying \cref{lemma:dynamics-error-set-v,lemma:function-class-covering-number}, we get that with probability at least $1-\delta / 7$ simultaneously for all $k \in [K], h \in [H], V \in \widehat{\V}(\mathcal{Z}_{h+1}, W, \betaQh[h+1])$
    \todo{fix explanation of conditional uniform convergence on $\Z_{h+1}$}
    \begin{align*}
        \norm{(\psiH - \psiKH)V}_{\covKH}
        &
        \le
        4 \betaQh[h+1] \sqrt{d \log (K+1) + 2\log (5H /\delta)
        +
        4 d \log (1 + 14 K^2 W \betaQ / \betaQ \sqrt{d})}
        \\
        &
        \le
        4 \betaQh[h+1] \sqrt{d \log (K+1) + 2\log (5H /\delta)
        +
        4 d \log \frac{15 K^2 W}{\sqrt{d}}}
        \\
        &
        \le
        12 \betaQh[h+1] \sqrt{d \log \frac{15 K^2 H W}{\delta \sqrt{d}}}
        \\
        &
        =
        \betaPh
        ,
    \end{align*}
    implying $\bar{E}_2$ (\cref{eq:goodTransitionProxPOLinear}).
    %
    %
    Now, suppose that the noise is generated such that 
    $
    \zph^{\kEpoch,i} = \sqrt{2 H \betaP^2} (\hatCovKH)^{-1/2} g_h^{e,i}
    $ 
    where $g_{p,h}^{e,i} \sim \mathcal{N}(0, I_d)$ are i.i.d for all $e \in \setEpochs, i \in [m], h \in [H]$.
    Indeed, notice that
    \begin{align*}
        \EE (\zph^{\kEpoch,i})(\zph^{\kEpoch,i})\tran
        &
        =
        2 H \betaP^2 (\hatCovKH)^{-1/2} \EE \brk[s]*{(g_{p,h}^{e,i}) (g_{p,h}^{e,i})\tran} (\hatCovKH)^{-1/2}
        =
        2 H \betaP^2 (\hatCovKH)^{-1}
        .
    \end{align*}
    Taking a union bound over \cref{lemma:gaussian-norm-concentration} with $\delta / 14 m H K$, we have that with probability at least $1-\delta/14$, simultaneously for all $i \in [m], h \in [H], e \in \setEpochs$
    \begin{align*}
        \norm{\zph^{\kEpoch,i}}_{\hatCovKH}
        =
        \sqrt{2 H}\betaP \norm{g_{p,h}^{e,i}}
        \le
        \betaP
        \sqrt{11 d H \log (14 m H K/\delta)}
        =
        \betaP\betaZ
        .
    \end{align*}
    Similarly, defining $\zr^{\kEpoch,i} = \sqrt{2\betaR^2} (\covK)^{-1/2} g_r^{e,i},$ with $g_{r}^{e,i} \sim \mathcal{N}(0, I_{dH})$,
    and taking a union bound over \cref{lemma:gaussian-norm-concentration} with $\delta / 14 m K$, we have that with probability at least $1-\delta/14$, simultaneously for all $i \in [m], k \in [K]$
    \begin{align*}
        \norm{\zr^{\kEpoch,i}}_{\hatCovK}
        =
        \sqrt{2}\betaR \norm{g_{r}^{e,i}}
        \le
        \betaR
        \sqrt{11 d H \log (14 m H K/\delta)}
        =
        \betaR\betaZ
        .
    \end{align*}
    Taking a union bound over the last two events shows that $E_3$ (\cref{eq:goodZetaConcentrationPOLinear}) holds with probability at least $1-\delta/7$.
    
    %
    %
    Next, for any $e \in \setEpochs$ notice that conditioned on $\Z_h, h \in [H]$ and $\covZ$ we have that $(\bar{\phi}^{\piOpt})\tran \z^{\kEpoch,i}, i \in [m]$ are i.i.d $\mathcal{N}(0, \sigma^2)$ variables with
    $
        \sigma^2
        =
        \norm*{\bar{\phi}^{\piOpt}}_{\covZ}^2
        .
    $
    Applying \cref{lemma:gaussian-anti-concentration} with $m = 9 \log (7K / \delta)$ and taking a union bound, we have that with probability at least $1-\delta/7$
    $E_4$ (\cref{eq:goodZetaAntiConcentrationPOLinear}) holds. 
    %
    %
    
    Next, by \cref{lemma:reward-free-exploration}, $E_5$ (\cref{eq:goodWarmupPOLinear}) holds with probability at least $1-\delta/7$.
    %
    %
    %
    Finally, notice that 
    $
    \norm{\phi^k}_{(\covK)^{-1}}, \norm{\phi_h^k}_{(\covKH)^{-1}} \le 1
    ,
    $
    thus 
    $
    0
    \le
    Y_k
    \le
    (1+ \sqrt{2}\betaZ)(\betaR + H \betaP)
    .
    $
    Using Azuma's inequality, we conclude that  $E_6$ (\cref{eq:goodBernsteinPOLinear}) holds with probability at least $1-\delta/7$.
    %
    %
    %
    $\bar{E}_7$ (\cref{eq:goodAzumaProxPOLinear}) holds with probability at least $1 - \delta/7$ by a similar argument.
\end{proof}

\paragraph{The good event.}
The following results show that the proxy good event implies the good event.

\begin{lemma}
\label{lemma:value-in-class}
    Suppose that $\EgoodBar$ holds. If $\pi^{k,i}_h \in \Pi(\Z_h, W K)$ for all $h \in [H]$ then $\hat{Q}^{k,i}_h \in \widehat{\Q}(\Z_h, W, \betaQh), \hat{V}^{k,i}_h \in \widehat{\V}(\Z_h, W, \betaQh)$ for all $h \in [H+1]$.
\end{lemma}
\begin{proof} 
    We show that the claim holds for by backwards induction on $h \in [H+1]$.
    \\
    \textbf{Base case $h = H+1$:}
    Since $V_{H+1}^{k,i} = 0$ it is also implied that $\hat{Q}^{k,i}_{H+1} = 0$. Because $w = 0 \in \widehat{\Q}(\Z_h, W, \betaQh[H+1] = 0)$ we have that $\hat{Q}^{k,i}_{H+1} \in \widehat{\Q}(\Z_h, W, \betaQh[H+1] = 0)$, and similarly $V_{H+1}^{k,i} \in \hat{V}^{k,i}_h \in \widehat{\V}(\Z_h, W, \betaQh)$.

    \textbf{Induction step:}
    Now, suppose the claim holds for $h+1$ and we show it also holds for $h$. We have that
    \begin{align*}
        \abs{\hat{Q}_h^{k,i}(x,a)}
        &
        =
        \abs{
        \phi(x,a)\tran w_h^{k,i} \indEvent{x \in \Z_h}
        }
        \\
        &
        =
        \indEvent{x \in \Z_h} \cdot 
        \abs{
        \phi(x,a)\tran (
        \theta_h + (\thetaKH - \theta_h) + \zrh^{k,i} + \zph^{k,i}
        +
        (\psiKH - \psiH) \hat{V}^{k,i}_{h+1}
        +
        \psiH \hat{V}^{k,i}_{h+1}
        )
        }
        \\
        \tag{triangle inequality, Cauchy-Schwarz, $\norm{\phi(x,a)} \le 1$}
        &
        \le
        1
        +
        \norm{\thetaK - \theta}
        +
        \norm{\zr^{k,i}}
        +
        \norm{\hat{V}_{h+1}^{k,i}}_\infty
        \\
        &
        \qquad
        +
        \indEvent{x \in \Z_h} \cdot
        \norm{\phi(x,a)}_{(\hatCovKH)^{-1}} 
        \brk[s]*{
        \norm{\zph^{k,i}}_{\hatCovKH}
        +
        \norm{(\psiKH - \psiH)\hat{V}^{k,i}_{h+1}}_{\hatCovKH}
        }
        \\
        \tag{induction hypothesis, \cref{eq:goodRewardPOLinear,eq:goodZetaConcentrationPOLinear,eq:goodTransitionProxPOLinear,eq:goodWarmupPOLinear}}
        &
        \le
        1
        +
        \frac{\betaR}{\sqrt{H}}
        +
        \frac{\betaZ \betaR}{\sqrt{H}}
        +
        \betaQh[h+1]
        +
        \frac{1}{2 \betaWarmup H}
        \brk[s]*{\betaZ \betaP + 12 \betaQh[h+1] \sqrt{d \log \frac{15 K^2 H W}{\delta \sqrt{d}}}}
        \\
        \tag{$\betaWarmup \ge 36 \betaZ \sqrt{d \log \frac{15 K^2 H W}{\delta \sqrt{d}}}$}
        &
        \le
        \betaQh[H]
        +
        \brk*{1 + \frac{1}{H}}\betaQh[h+1]
        \\
        \tag{\cref{eq:betaQh-def}}
        &
        =
        \betaQh
        ,
    \end{align*}
    and
    \begin{align*}
        \norm{w_h^{k,i}}
        &
        =
        \norm{
        \thetaKH + \zrh^{k,i} + \zph^{k,i}
        +
        \psiKH  \hat{V}^{k,i}_{h+1}
        }
        \le
        H K + \frac{\betaR \betaZ}{\sqrt{H}} + \betaP \betaZ + \betaQ K
        \le
        4 \betaQ K
        =
        W
        .
    \end{align*}
    Since $\pi_h^{k,i} \in \Pi(\Z_h, W)$, this proves the induction step and concludes the proof.
\end{proof}

\begin{lemma*}[restatement of \cref{lemma:good-event-PO-linear}]
    Consider the following parameter setting:
    \begin{gather*}
        \lambdaP = 1
        ,
        \lambdaR = H
        ,
        m = 9 \log (7K / \delta)
        ,
        \betaR
        =
        2 H \sqrt{2dH \log (14K/\delta)}
        ,
        \betaZ
        =
        \sqrt{11 d H \log (14 m H K/\delta)}
        ,
        \\
        \betaP
        =
        216 \betaZ \betaR \sqrt{d H \log \frac{60 K^3 H \betaQ}{\delta \sqrt{d}}}
        ,
        \betaQ
        =
        16 \betaZ \betaR \sqrt{H}
        ,
        \betaWarmup 
        =
        36 \betaZ \sqrt{d \log \frac{60 K^3 H \betaQ }{\delta \sqrt{d}}}
        ,
        \etaO \le 1
        ,
        \epsCov \ge 1/K
        .
    \end{gather*}
    Then $\Pr[\Egood] \ge 1 - \delta.$
\end{lemma*}

\begin{proof}
    Suppose that $\EgoodBar$ holds. By \cref{lemma:good-proxy-event-PO-linear}, this occurs with probability at least $1-\delta$. We show that $\EgoodBar$ implies $\Egood$, thus concluding the proof.
    Notice that 
    \begin{align*}
        \pi_h^{k,i}(a | x)
        \propto
        \exp\brk*{\eta \sum_{k' = k_e}^{k-1} \hat{Q}_h^{k',i}(x,a)}
        &
        =
        \exp\brk*{\indEvent{x \in \Z_h} \phi(x,a)\tran \eta \sum_{k' = k_e}^{k-1} w_h^{k',i}}
        \\
        &
        =
        \exp \brk*{\indEvent{x \in \Z_h} \phi(x,a)\tran q_h^{k,i}}
        ,
    \end{align*}
    where $q_h^{k,i} = \eta \sum_{k' = k_e}^{k-1} w_h^{k',i}.$ We thus have that $\norm{q_h^{k,i}} \le W K$ implies $\pi_h^{k,i} \in \Pi(\Z_h, W K).$ 
    We show by induction on $k \ge k_0$ that $\pi_h^{k,i} \in \Pi(\Z_h, W K)$ for all $h \in [H]$. For the base case, $k = k_0$, $\pi_h^{k,i}$ are uniform, implying that $q_h^{k_0,i} = 0$, thus $\pi_h^{k_0,i} \in \Pi(\Z_h, WK)$.
    Now, suppose the claim holds for all $k' < k$. Then by \cref{lemma:value-in-class} we have that $\hat{Q}_h^{k',i} 
    \in \widehat{\Q}(\Z_h, W, \betaQh[h+1])$ for all $k' < k$ and $h \in [H]$. This implies that $\norm{w_h^{k',i}} \le W$ for all $k' < k$ and $h \in [H]$, thus $\norm{q_h^{k,i}} \le W K \eta \le W K$ for all $h \in [H]$, concluding the induction step.

    Now, since $\pi_h^{k,i} \in \Pi(\Z_h, W K)$ for all $k \ge k_0, h \in [H], i \in [m]$, we can apply \cref{lemma:value-in-class} to get that $\hat{Q}_h^{k,i} \in \widehat{\Q}(\Z_h, W, \betaQh[h+1]), \hat{V}_h^{k,i} \in \widehat{\V}(\Z_h, W, \betaQh[h+1])$ for all $k \ge k_0, h \in [H], i \in [m]$.
    Notice that for any $h \in [H]$, 
    $
    \betaQh
    \le
    3H \betaQh[H] 
    =
    18 \sqrt{H} \betaZ \betaR
    =
    \betaQ
    ,
    $
    and
    \begin{align*}
        \betaPh
        =
        12 \betaQh[h+1] \sqrt{d \log \frac{15 K^2 H W}{\delta \sqrt{d}}}
        &
        \le
        36 \betaQh[H] H \sqrt{d \log \frac{15 K^2 H W}{\delta \sqrt{d}}}
        \\
        &
        =
        216 \betaZ \betaR \sqrt{d H \log \frac{15 K^2 H W}{\delta \sqrt{d}}}
        =
        \betaP
        .
    \end{align*}
    Using $\bar{E}_2$ (\cref{eq:goodTransitionProxPOLinear}) we conclude that $E_2$ (\cref{eq:goodTransitionPOLinear}) holds.
    This also implies that $\clip_{\betaQ}\brk[s]*{\hat{V}_1^{k,i}(x_1)} = \hat{V}_1^{k,i}(x_1)$, thus $\bar{E}_7$ (\cref{eq:goodAzumaProxPOLinear}) implies $E_7$ (\cref{eq:goodAzumaPOLinear}).
\end{proof}

\newpage
\section{Randomized Ensemble Policy Optimization (REPO) for Tabular MDPs}
\label{sec:REPO-tabular}

In this section, we present a simplified version of REPO (\cref{alg:r-opo-for-linear-mdp}) for tabular MDPs. For consistency of the presentation, we encode the tabular MDP as a linear MDP of dimension $d = \abs{\X} \abs{\A}$ using a standard one-hot encoding (see Example 2.1 in \citet{jin2020provably}).

\subsection{The Randomized Ensemble Policy Optimization Algorithm}
We present REPO for tabular MDPs in \cref{alg:r-opo}. The algorithm is identical to \cref{alg:r-opo-for-linear-mdp} up to the following differences. First, we remove the reward-free warm-up step. Second, we remove the indicator mechanism used to keep the Q values bounded. Third, we affix the dynamics backup estimate at the start of each epoch (see the choice of dataset $\Dh^k$ in \cref{line:REPO-dataset-tabular} of \cref{alg:r-opo}). Otherwise, the algorithm is unchanged. As will be seen in the analysis, this simplification is made possible since the empirical tabular MDP, i.e., the one induced by the observed samples, has a sub-stochastic transition kernel, i.e., one with non-negative values that sum to at most one.
\begin{algorithm}[!ht]
	\caption{REPO with aggregate feedback for tabular MDPs} \label{alg:r-opo}
	\begin{algorithmic}[1]
	    \State \textbf{input}:
		$\delta, \etaO, \etaH, \lambdaR, \lambdaP, \betaR, \betaP > 0 ; m \ge 1$.

            \State \textbf{initialize}: $e \gets -1$.
        
	    \For{episode $k = 1,2,\ldots, K$}
     
                \If{\mbox{$k = 1$
                    \textbf{ or }
                    $\exists h \in [H], \  \det\brk{\covKH} \ge 2 \det\brk{\hatCovKH}$
                    \textbf{ or }
                    $ \det\brk{\covK} \ge 2 \det\brk{\hatCovK}$}
                }
                \label{line:repo-epoch-condition-tabular}

                    \State $e \gets e + 1$ and $\kEpoch \gets k$.

                    \State 
                    \label{line:REPO-covZ-tabular}
                    $
                        {\covZ
                        \gets
                        2\betaR^2 {(\hatCovK)}^{-1}
                        +
                        2 H \betaP^2 \mathrm{diag}\brk*{\hatCovKH[1], \ldots, \hatCovKH[H]}^{-1}}
                    $

                    \State Sample $\z^{\kEpoch,i} 
                    \sim 
                    \mathcal{N}\brk*{0, \covZ}$ for all $i \in [m]$.

                    \State Reset $p^k(i) \gets 1/m , \pi^{k,i}_h(a \mid x) \gets 1/\abs{\A}$.
                
                \EndIf

            \State Sample $i_k$ according to $p^k(\cdot)$ and  play $\piK = \pi^{k,i_k}$.

            \State Observe episode reward $v^k$ and trajectory $\iota^k$.

            \State Define $\D^k = \brk[c]{1, \ldots, k-1}$ and $\Dh^k = \brk[c]{1, \ldots, \kEpoch - 1}$ for all $h \in [H]$.
            \label{line:REPO-dataset-tabular}
            
            \State Compute $\thetaK$ and define $\psiKH$ (\cref{eq:LS-estimate-theta,eq:LS-estimate-psi}).
            
            \State Define $\hat{V}_{H+1}^{k,i}(x) = 0$ for all $i\in[m], x \in \mathcal{X}$.
            
            \State For every $i \in [m]$ and $h = H, \ldots, 1$:
            %
            \begin{alignat*}{2}
                    &
                    \qquad
                    w_h^{k,i}
                    &&
                    \gets
                    \thetaKH + \zh^{\kEpoch,i}
                    +
                    \psiKH \hat{V}^{k,i}_{h+1}
                    \\  
                    &
                    \qquad
                    \hat{Q}_h^{k,i}(x,a)
                    &&
                    =
                    \phi(x,a)\tran w_h^{k,i}
                    \\
                    &
                    \qquad
                    \hat{V}_h^{k,i}(x)
                    &&
                    =
                    \sum_{a \in \mathcal{A}} \pi^{k,i}_h(a \mid x) \hat{Q}_h^{k,i}(x,a)
                    \\
                    &
                    \qquad
                    \pi^{k+1,i}_h(a \mid x)
                    &&
                    \propto
                    \pi^{k,i}_h(a \mid x) \exp (\etaO \hat{Q}_h^{k,i}(x,a))
                    .
                \end{alignat*}
            \label{line:REPO-tabular}
	    

           \State
           \label{line:REPO-Hedge-tabular}
           $p^{k+1}(i) \propto p^{k}(i) \exp(\etaH \hat{V}_1^{{k,i}}(x_1))$ for all $i \in [m]$.

	    \EndFor
	\end{algorithmic}
	
\end{algorithm}

The following is our main result for \cref{alg:r-opo} (proof in the following \cref{sec:analysis-repo-tabular}).
\begin{theorem}
\label{thm:regret-bound-PO}
    Suppose that we run \cref{alg:r-opo} with $\etaH = \sqrt{\frac{3 d H \log (2K) \log m}{K \betaQ^2}} , \etaO = \sqrt{\frac{3 d H \log (2K) \log \abs{\A}}{K \betaQ^2}}$ and the other parameters as in \cref{lemma:good-event-PO}. Then with probability at least $1 - \delta$ we incur regret at most
    \begin{align*}
        \regret
        & 
        \le
        1747 \sqrt{ K d^3 H^7 \max\brk[c]{H, \numStates}}\log^2 \frac{20 K H \numStates \abs{\A} m}{\delta}
        \\
        &
        =
        1747 \sqrt{ K \numStates^3 \abs{\A}^3 H^7 \max\brk[c]{H, \numStates}}\log^2 \frac{20 K H \numStates \abs{\A} m}{\delta}
        \\
        &
        =
        \tilde{O}(\sqrt{K \numStates^3 \abs{\A}^3 H^7 \max \{H, \numStates\} })
        .
    \end{align*}
\end{theorem}

\subsection{Analysis (REPO for tabular MDPs)}
\label{sec:analysis-repo-tabular}

\paragraph{Good event.}
Let $\widehat{P}^k = (\widehat{P}^k_1, \ldots, \widehat{P}^k_H)$ be the empirical (sub-probability) transition kernel defined such that $\widehat{P}^k_h (\cdot | x,a) = \phi(x,a) \psiKH$ for all $h \in [H]$. Recall that we keep the kernel fixed throughout each epoch, thus $\widehat{P}^k = \widehat{P}^{\kEpoch}$ for all $e \in \setEpochs, k \in \setEpochsE$. As in standard MDPs, the empirical transition kernel together with a policy $\pi$ defines a (sub) probability measure over trajectories $\iota = (x_h,a_h)_{h \in [H]}$. We can therefore define an expectation operator $\EE[\widehat{P}^k, \pi]$ as we did for standard MDPs. While linearity of the expectation holds for any measure, the fact that it is a sub-probability measure implies that $\EE[\widehat{P}^k, \pi] f(\iota) \le \max_\iota f(\iota).$

Finally, for a policy $\pi$ and $k \in [K], h \in [H]$, we define the expected empirical feature occupancy as
$
\hat{\phi}_h^{k, \pi}
=
\EE[\widehat{P}^k, \pi] \phi(x_h, a_h)
$
and their concatenation as
$
\hat{\phi}^{k,\pi}
=
((\hat{\phi}^{k,\pi}_1)\tran, \ldots, (\hat{\phi}^{k,\pi}_{H})\tran)\tran
.
$
In addition, to simplify presentation, we denote $\zeta^{k,i} = \zr^{k,i} + \zp^{k,i}$ where $\zr^{k,i} \sim \mathcal{N}\brk*{0, 2\betaR^2 {(\covK)}^{-1}}$ and $\zp^{k,i} \sim \mathcal{N}\brk*{0, 2 H \betaP^2 \mathrm{diag}\brk*{\covKH[1], \ldots, \covKH[H]}^{-1}}$.
We define the following good event $\Egood = \bigcap_{i=1}^7 E_i$, over which the regret is deterministically bounded:
\begin{flalign}
    \label{eq:goodRewardPO}
     &
     E_1 = \brk[c]*{\forall e \in \setEpochs, k \in \setEpochsE : \norm{\theta - \widehat{\theta}^k}_{\covK} \le \betaR};
     &
     \\
     \label{eq:goodTransitionPO}
     &
     E_2 = \brk[c]*{\forall e \in \setEpochs, k \in \setEpochsE , i \in [m] , h \in [H] : \norm{(\psiH - \psiKH){V}_{h+1}^{\star}}_{\hatCovKH} \le \betaP, \norm{\hat{Q}_{h+1}^{k,i}}_\infty \le \betaQ};
     &
     \\
     \label{eq:goodZetaConcentrationPO}
     &
     E_3 = \brk[c]*{\forall e \in \setEpochs, k \in \setEpochsE, i \in [m], h\in [H] : \norm{\zr^{\kEpoch,i}}_{\hatCovK} \le \betaZ\betaR, \norm{\zph^{\kEpoch,i}}_{\hatCovKH} \le \betaZ\betaP};
     &
     \\
     \label{eq:goodZetaAntiConcentrationPO}
     &
     E_4 = \brk[c]*{\forall e \in \setEpochs : \max_{i \in [m]} (\hat{\phi}^{\kEpoch,\piOpt})\tran \z^{\kEpoch,i} \ge \norm{\hat{\phi}^{\kEpoch,\piOpt}}_{\covZ}};
     &
     \\
     \label{eq:goodTransitionPOuniform}
     &
     E_5 = \brk[c]*{\forall e \in \setEpochs, k \in \setEpochsE , h \in [H], x \in \X, a \in \A : \norm{\phi(x,a)\tran (\psiH - \psiKH)}_{1} \le \betaPtag \norm{\phi(x,a)}_{\hatCovKH^{-1}}};
     &
    \\
    \label{eq:goodBernsteinPO}
    &
    E_6 
    =
    \brk[c]*{
    \sum_{k \in [K]} \EE[P, \piK] \brk[s]*{Y_k} \le \sum_{k \in [K]} Y_k 
    +
    ((1+ \sqrt{2}\betaZ)\betaR + \sqrt{2}(\betaP \betaZ + \betaQ \betaPtag)H) \sqrt{2K \log\frac{7}{\delta}}
    }.
    &
    \\
    \label{eq:goodAzumaPO}
    &
    E_7 
    =
    \brk[c]*{
    \sum_{k \in [K]} \sum_{i \in [m]} p^k(i) \hat{V}^{k,i}_1(x_1)
    \le \sum_{k \in [K]} \hat{V}^{k,i_k}_1(x_1) + 2 \betaQ  \sqrt{2K \log\frac{7}{\delta}}
    };
    &
\end{flalign}
where
$
    Y_k
    =
    \betaR (1 + \sqrt{2}\betaZ) \norm{\phi^k}_{(\covK)^{-1}}
        +
        \sqrt{2}(\betaP \betaZ + \betaQ \betaPtag) \sum_{h \in [H]} \norm{\phi^k_h}_{(\covKH)^{-1}} 
    .
$

\begin{lemma}[Good event]
\label{lemma:good-event-PO}
    Consider the following parameter setting:
    \begin{gather*}
        \lambdaP = 1
        ,
        \lambdaR = H
        ,
        m = 9 \log (7K / \delta)
        ,
        \betaR
        =
        2 H \sqrt{2dH \log (14K/\delta)}
        ,
        \betaZ
        =
        \sqrt{11 d H \log (14 m K/\delta)}
        ,
        \\
        \betaP
        =
        4 H \sqrt{3 d \log (14 KH / \delta)}
        ,
        \betaPtag
        =
        2 \sqrt{5 \numStates \log (20 K H \numStates \abs{\A} / \delta)}
        ,
        \betaQ
        =
        2 H \betaP \betaZ
         .
    \end{gather*}
    Then $\Pr[\Egood] \ge 1 - \delta.$
\end{lemma}
Proof in \cref{sec:good-event-proofs-PO}.

\paragraph{Optimism and its cost.}

\begin{lemma}[Optimism]
\label{lemma:optimism-PO}
    Suppose that the good event $\Egood$ holds (\cref{eq:goodRewardPO,eq:goodTransitionPOuniform,eq:goodTransitionPO,eq:goodZetaConcentrationPO,eq:goodZetaAntiConcentrationPO,eq:goodBernsteinPO,eq:goodAzumaPO}) and define $\hatJE = \argmax_{i \in [m]} (\hat{\phi}^{\kEpoch,\piOpt})\tran \z^{\kEpoch,i}$, then 
    \begin{align*}
        \sum_{h \in [H]} \EE[\widehat{P}^k, \piOpt]\brk[s]*{ Q^{\star}_h(x_h,a_h) - \phi(x_h,a_h)\tran \brk{ \thetaKH + \z^{\kEpoch,\hatJE} + \psiKH V^{\star}_{h+1} } }
        \le
        0
        \quad
        ,
        \forall e \in \setEpochs, k \in \setEpochsE
        .
    \end{align*}
\end{lemma}

\begin{proof}
    We have that
    \begin{align*}
        \sum_{h \in [H]}
        &
        \EE[\widehat{P}^k, \piOpt]\brk[s]*{ Q^{\star}_h(x_h,a_h) - \phi(x_h,a_h)\tran \brk{ \thetaKH + \zh^{\kEpoch,\hatJE} + \psiKH V^{\star}_{h+1} } }
        \\
        &
        =
        \sum_{h \in [H]}
        \EE[\widehat{P}^k, \piOpt]\brk[s]*{ \phi(x_h,a_h)\tran \brk{ \thetaH - \thetaKH - \zh^{\kEpoch,\hatJE} + (\psiH - \psiKH) V^{\star}_{h+1} }}
        \\
        &
        =
        \sum_{h \in [H]}
        (\hat{\phi}^{\kEpoch,\piOpt}_h)\tran \brk{ \thetaH - \thetaKH -  \zh^{\kEpoch,\hatJE} + (\psiH - \psiKH) V^{\star}_{h+1} }
        \\
        &
        =
        -(\hat{\phi}^{\kEpoch,\piOpt})\tran \z^{\kEpoch,\hatJE}
        +
        (\hat{\phi}^{\kEpoch,\piOpt})\tran (\theta - \thetaK)
        +
        \sum_{h \in [H]} 
        (\hat{\phi}^{\kEpoch,\piOpt}_{h})\tran 
        (\psiH - \psiKH) V^{\star}_{h+1}
        \\
        \tag{\cref{eq:goodRewardPO,eq:goodTransitionPO}}
        &
        \le
        -(\hat{\phi}^{\kEpoch,\piOpt})\tran \z^{\kEpoch,\hatJE}
        +
        \betaR \norm{\hat{\phi}^{\kEpoch,\piOpt}}_{(\covK)^{-1}}
        +
        \betaP \sum_{h \in [H]} \norm{\hat{\phi}^{\kEpoch,\piOpt}_{h}}_{(\hatCovKH)^{-1}}
        \\
        \tag{$\hatCovK \preceq \covK$}
        &
        \le
        -(\hat{\phi}^{\kEpoch,\piOpt})\tran \z^{\kEpoch,\hatJE}
        +
        \betaR \norm{\hat{\phi}^{\kEpoch,\piOpt}}_{(\hatCovK)^{-1}}
        +
        \betaP \sum_{h \in [H]} \norm{\hat{\phi}^{\kEpoch,\piOpt}_{h}}_{(\hatCovKH)^{-1}}
        \\
        \tag{Cauchy-Schwarz}
        &
        \le
        -(\hat{\phi}^{\kEpoch,\piOpt})\tran \z^{\kEpoch,\hatJE}
        +
        \norm{\hat{\phi}^{\kEpoch,\piOpt}}_{2 \betaR^2 (\hatCovKH)^{-1} + 2 H \betaP^2 \mathrm{diag}\brk{\hatCovKH[1], \ldots, \hatCovKH[H]}^{-1}}
        \\
        \tag{Definitions of $\hatJE , \covZ$}
        & 
        =
        \norm{\hat{\phi}^{\kEpoch,\piOpt}}_{\covZ}
        -
        \max_{i \in [m]} (\hat{\phi}^{\kEpoch,\piOpt})\tran \z^{\kEpoch,i} 
        \\
        \tag{\cref{eq:goodZetaAntiConcentrationPO}}
        & 
        \le
        0
        .
    \end{align*}
\end{proof}

\begin{lemma}[Cost of optimism]
\label{lemma:cost-of-optimism-PO}
    Suppose that the good event $\Egood$ holds (\cref{eq:goodRewardPO,eq:goodTransitionPO,eq:goodTransitionPOuniform,eq:goodZetaConcentrationPO,eq:goodZetaAntiConcentrationPO,eq:goodBernsteinPO,eq:goodAzumaPO}), then for all $ e \in \setEpochs ,k \in \setEpochsE$
    \begin{align*}
        \hat{V}^{k, i_k}_1(x_1)
        -
        V_1^{\piK}(x_1)
        \le
        \EE[P, \piK]\brk[s]*{ 
        \betaR (1 + \sqrt{2}\betaZ) \norm{\phi(x_{1:H},a_{1:H})}_{(\covK)^{-1}}
        +
        \sqrt{2}(\betaP \betaZ + \betaQ \betaPtag) \sum_{h \in [H]} \norm{\phi(x_h, a_h)}_{(\covKH)^{-1}} 
        }
        .
    \end{align*}
\end{lemma}

\begin{proof}
    By \cref{lemma:extended-value-difference}, a value difference lemma by \citet{shani2020optimistic},
    \begin{align*}
        \hat{V}^{k, i_k}_1(x_1)
        &
        -
        V_1^{\piK}(x_1)
        =
        \EE[P, \piK]\brk[s]*{
        \sum_{h \in [H]}  \phi(x_h,a_h)\tran \brk*{ \thetaKH - \thetaH + \zrh^{\kEpoch,i_k} + \zph^{\kEpoch,i_k} + (\psiKH - \psiH)\hat{V}_{h+1}^{k,i_k} }}
        \\
        &
        =
        \EE[P, \piK]\brk[s]*{\phi(x_{1:H},a_{1:H})\tran (\thetaK - \theta + \zr^{\kEpoch, i_k}) + 
        \sum_{h \in [H]}  \phi(x_h,a_h)\tran \brk*{ \zph^{\kEpoch,i_k} + (\psiKH - \psiH)\hat{V}_{h+1}^{k,i_k} }}
        \\
        \tag{Cauchy-Schwarz, \cref{eq:goodRewardPO,eq:goodZetaConcentrationPO}}
        & \le
        \EE[P, \piK]
        \Bigg[
        \betaR \norm{\phi(x_{1:H},a_{1:H})}_{(\covK)^{-1}}
        +
        \betaR\betaZ \norm{\phi(x_{1:H},a_{1:H})}_{(\hatCovK)^{-1}} 
        \\
        &
        \qquad\qquad
        +
        \sum_{h \in [H]} \betaP \betaZ \norm{\phi(x_h, a_h)}_{(\hatCovKH)^{-1}} 
        +
        \norm{(\phi(x_h, a_h))\tran (\psiKH - \psiH)}_1 \norm{\hat{V}_{h+1}^{k,i_k}}_\infty
        \Bigg]
        \\
        \tag{\cref{lemma:matrix-norm-inequality,eq:goodTransitionPO,eq:goodTransitionPOuniform}}
        &
        \le
        \EE[P, \piK]\brk[s]*{ 
        \betaR (1 + \sqrt{2}\betaZ) \norm{\phi(x_{1:H},a_{1:H})}_{(\covK)^{-1}}
        +
        \sqrt{2}(\betaP \betaZ + \betaQ \betaPtag) \sum_{h \in [H]} \norm{\phi(x_h, a_h)}_{(\covKH)^{-1}} 
        }
        .
    \end{align*}
\end{proof}

\paragraph{Regret bound.}

%
%
\begin{proof}[of \cref{thm:regret-bound-PO}]
    Suppose that the good event $\Egood$ holds (\cref{eq:goodRewardPO,eq:goodTransitionPOuniform,eq:goodTransitionPO,eq:goodZetaConcentrationPO,eq:goodZetaAntiConcentrationPO,eq:goodBernsteinPO,eq:goodAzumaPO}). By \cref{lemma:good-event-PO}, this holds with probability at least $1-\delta$.
    For every epoch $e \in \setEpochs$, let $\hatJE = \argmax_{i \in [m]} \sum_{k \in e} \hat{\phi}^{k,\piOpt}(x_1)\tran \z^{k,i}$. Additionally, for every $e \in \setEpochs, k \in \setEpochsE, i \in [m]$ let $\hat{V}^{k,i,\piOpt}_h$ be the value function of the true optimal policy $\piOpt$ in the empirical MDP-like structure whose rewards and dynamics are defined as $\hat{r}^{k,i}_h(x,a) = \phi(x,a)\tran(\thetaKH + \zh^{\kEpoch,i})$ and $\widehat{P}^k_h(\cdot | x,a) = \phi(x,a)\tran \psiKH$. 
    Now decompose the regret as follows.
    \begin{align*}
        \regret
        =
        \sum_{k \in [K]} V^{\piOpt}_1(x_1) - V^{\pi^k}_1(x_1)
        & 
        =
        \underbrace{\sum_{k \in [K]} \hat{V}^{k,i_k}_1(x_1) - V^{\pi^k}_1(x_1)}_{(i)} 
        +
        \underbrace{\sum_{e \in \setEpochs} \sum_{k \in \setEpochsE} \hat{V}^{k,\hatJE}_1(x_1) -  \hat{V}^{k,i_k}_1(x_1)}_{(ii)}
        \\
        & 
        + 
        \underbrace{\sum_{e \in \setEpochs} \sum_{k \in \setEpochsE} \hat{V}^{k,\hatJE,\piOpt}_1(x_1) - \hat{V}^{k,\hatJE}_1(x_1) }_{(iii)}
        +
        \underbrace{\sum_{e \in \setEpochs} \sum_{k \in \setEpochsE} V^\star_1(x_1) - \hat{V}^{k,\hatJE,\piOpt}_1(x_1) }_{(iv)}
        %
        .
    \end{align*}

    For term $(i)$, we use \cref{lemma:cost-of-optimism-PO} as follows.
    \begin{align*}
        (i)
        & 
        \le
        \sum_{e \in \setEpochs} \sum_{k \in \setEpochsE} \EE[P, \piK]\brk[s]*{ 
        \betaR (1 + \sqrt{2}\betaZ) \norm{\phi(x_{1:H},a_{1:H})}_{(\covK)^{-1}}
        +
        \sqrt{2}(\betaP \betaZ + \betaQ \betaPtag) \sum_{h \in [H]} \norm{\phi(x_h, a_h)}_{(\covKH)^{-1}} 
        }
        \\
        \tag{\cref{eq:goodBernsteinPO}}
        &
        \le 
        \betaR (1 + \sqrt{2}\betaZ) \sum_{k \in [K]} \norm{\phi^k}_{(\covK)^{-1}}
        +
        \sqrt{2}(\betaP \betaZ + \betaQ \betaPtag) \sum_{k \in [K]} \sum_{h \in [H]} \norm{\phi^k_h}_{(\covKH)^{-1}}
        \\
        &
        \qquad
        +
        ((1+ \sqrt{2}\betaZ)\betaR + \sqrt{2}(\betaP \betaZ + \betaQ \betaPtag)H) \sqrt{2K \log\frac{7}{\delta}}
        \\
        \tag{\cref{lemma:elliptical-potential}, $\betaZ, \betaPtag \ge 3, \betaP\sqrt{H} \ge \betaR, \betaQ = 2H \betaP \betaZ$}
        &
        \le
        3 \betaP \betaZ H
        \sqrt{ K d  \log(2K) }
        +
        5H^2 \betaP \betaZ \betaPtag \sqrt{K d \log(2K)}
        +
        6 H^2 \betaP \betaZ \betaPtag \sqrt{K \log\frac{7}{\delta}}
        \\
        & 
        \le
        12 H^2 \betaP \betaZ \betaPtag \sqrt{K d \log\frac{7K}{\delta}}
        .
    \end{align*}
    Next, by \cref{lemma:hedge-term-PO-linear,lem:num-epochs} (with our choice of $\etaH$), we have that 
    \begin{align*}
        (ii)
        \le
        2 \betaQ \sqrt{K E \log m} + 2 \betaQ  \sqrt{2K \log\frac{7}{\delta}}
        &
        \le
        2 \betaQ \sqrt{3 K d H \log(2K) \log m} + 2 \betaQ  \sqrt{2K \log\frac{7}{\delta}}
        \\
        &
        \le
        7 \betaQ \sqrt{K d H} \log \brk*{\frac{7 K m}{\delta}}
        .
    \end{align*}
    Next, we decompose term $(iii)$ using \cref{lemma:extended-value-difference} as follows. 
    \begin{align*}
        (iii)
        & 
        =
        \sum_{e \in \setEpochs} \sum_{k \in \setEpochsE} \EE[\widehat{P}^k, \piOpt]\brk[s]*{
        \sum_{h \in [H]} \sum_{a \in \A}
        \hat{Q}^{k,\hatJE}_{h}(x_h, a)  \brk{\piOpt[h](a \mid x_h) - \pi^{k,\hatJE}_h(a \mid x_h)} }
        \\
        & 
        \tag{$\widehat{P}^k$ fixed within epoch}
        =
        \sum_{h \in [H]} \sum_{e \in \setEpochs} \EE[\widehat{P}^k, \piOpt]\brk[s]*{
        \sum_{k \in \setEpochsE} \sum_{a \in \A}
        \hat{Q}^{k,\hatJE}_{h}(x_h, a)  \brk{\piOpt[h](a \mid x_h) - \pi^{k,\hatJE}_h(a \mid x_h)} }
        \\
        \tag{\cref{lemma:omd-term-PO-linear}, $\widehat{P}^k$ is a sub-probability measure}
        &
        \le
        \sum_{h \in [H]} \sum_{e \in \setEpochs} \brk*{
        \frac{\log \abs{\A}}{\etaO} + \etaO \sum_{k \in \setEpochsE} \betaQ^2}
        \\
        & 
        \le
        \frac{H \numEpochs \log \abs{\A}}{\etaO} + \etaO 
        H K \betaQ^2
        \\
        \tag{\cref{lem:num-epochs} and choice of $\etaO$}
        &
        \le
        4 H \betaQ \sqrt{ K d H} \log (2K \abs{\A})
        .
    \end{align*}
    Next, we decompose term $(iv)$ using \cref{lemma:extended-value-difference} as follows.
    \begin{align*}
        (iv)
        =
        \sum_{e \in \setEpochs} \sum_{k \in \setEpochsE}
        \sum_{h \in [H]} \EE[\widehat{P}^k, \piOpt]\brk[s]*{ Q^{\star}_h(x_h,a_h) - \phi(x_h,a_h)\tran \brk{ \thetaKH + \z^{\kEpoch,\hatJE} + \psiKH V^{\star}_{h+1} } }
        \le
        0
        ,
    \end{align*}
    where the second inequality is by \cref{lemma:optimism-PO}.
    Putting everything together, we conclude that
    \begin{align*}
        \regret
        &
        \le
        12 H^2 \betaP \betaZ \betaPtag \sqrt{K d \log\frac{7K}{\delta}}
        +
        7 \betaQ \sqrt{K d H} \log \brk*{\frac{7 K m}{\delta}}
        +
        4 H \betaQ \sqrt{ K d H} \log (2K \abs{\A})
        \\
        &
        \le
        12 H^2 \betaP \betaZ \betaPtag \sqrt{K d \log\frac{7K}{\delta}}
        +
        22 H^2 \betaP \betaZ \sqrt{K d H} \log \brk*{\frac{7 K m \abs{\A}}{\delta}}
        \\
        &
        \le
        24 H^2 \betaP \betaZ \sqrt{5 K d \numStates}\log \frac{20 K H \numStates \abs{\A}}{\delta}
        +
        22 H^2 \betaP \betaZ \sqrt{K d H} \log \brk*{\frac{7 K m \abs{\A}}{\delta}}
        \\
        &
        \le
        76 H^2 \betaP \betaZ \sqrt{K d \max\brk[c]{H, \numStates}}\log \frac{20 K H \numStates \abs{\A} m}{\delta}
        \\
        &
        \le
        1747 \sqrt{ K d^3 H^7 \max\brk[c]{H, \numStates}}\log^2 \frac{20 K H \numStates \abs{\A} m}{\delta}
        \\
        &
        =
        1747 \sqrt{ K \numStates^3 \abs{\A}^3 H^7 \max\brk[c]{H, \numStates}}\log^2 \frac{20 K H \numStates \abs{\A} m}{\delta}
        .
        \qedhere
    \end{align*}
\end{proof}

\subsection{Proofs of good event (REPO in Tabular MDPs)}
\label{sec:good-event-proofs-PO}

\begin{lemma*}[restatement of \cref{lemma:good-event-PO}]
    Consider the following parameter setting:
    \begin{gather*}
        \lambdaP = 1
        ,
        \lambdaR = H
        ,
        m = 9 \log (7K / \delta)
        ,
        \betaR
        =
        2 H \sqrt{2dH \log (14K/\delta)}
        ,
        \betaZ
        =
        \sqrt{11 d H \log (14 m K/\delta)}
        ,
        \\
        \betaP
        =
        4 H \sqrt{3 d \log (14 KH / \delta)}
        ,
        \betaPtag
        =
        2 \sqrt{5 \numStates \log (20 K H \numStates \abs{\A} / \delta)}
        ,
        \betaQ
        =
        2 H \betaP \betaZ
         .
    \end{gather*}
    Then $\Pr[\Egood] \ge 1 - \delta.$
\end{lemma*}
\begin{proof}
    First, by \cref{lemma:reward-error} and our choice of parameters, $E_1$ (\cref{eq:goodRewardPO}) holds with probability at least $1- \delta/7$.
    %
    %
    Next, $E_3, E_4$ (\cref{eq:goodZetaConcentrationPO,eq:goodZetaAntiConcentrationPO}) follow exactly as \cref{eq:goodZetaConcentrationPOLinear,eq:goodZetaAntiConcentrationPOLinear} proved in \cref{lemma:good-proxy-event-PO-linear}. Specifically, the argument for $E_4$ is unchanged since $\widehat{P}^{\kEpoch}$ and therefore $\hat{\phi}^{\kEpoch,\piOpt}$ are determined before drawing $(\z^{\kEpoch,i})_{i \in [m]}$.
    %
    %

    Now, to prove that $E_5$ (\cref{eq:goodTransitionPOuniform}) holds with probability at least $1 - \delta / 7$, we use standard arguments for tabular MDPs.
    Let $\Bar{P}_h^k(x' \mid x,a) = n^k_h(x,a,x') / \max \brk[c]{1 , n^k_h(x,a)}$ be the empirical transition function where $n^k_h(x,a,x')$ is the number of times (until the beginning of episode $k$) that the agent visited state $x$ at step $h$, took action $a$ and transitioned to state $x'$, and $n^k_h(x,a) = \sum_{x'} n^k_h(x,a,x')$.
    By \citet[Lemma 17]{jaksch2010near}, with probability at least $1 - \delta/7$ for all $x \in \X, a \in \A, h \in [H], k \in [K]$ simultaneously 
    \begin{align*}
        \norm{\Bar{P}_h^k(\cdot \mid x,a) - P_h(\cdot \mid x,a)}_1
        \le
        \sqrt{\frac{5 \numStates \log (20 K H \numStates \abs{\A} / \delta)}{\max \brk[c]{1 , n^k_h(x,a)}}}.
    \end{align*}
    Now notice that $\phi(x,a)\tran \psiH = P_h(\cdot \mid x,a)$ and that $\norm{\phi(x,a)}_{(\hatCovKH)^{-1}} = 1 / \sqrt{1 + n^{\kEpoch}_h(x,a)}$.
    In addition, $\phi(x,a)\tran \psiKH = \widehat{P}^{\kEpoch}_h(\cdot \mid x,a)$ when defining $\widehat{P}^{\kEpoch}_h(x' \mid x,a) = n^{\kEpoch}_h(x,a,x') / \brk{1 + n^{\kEpoch}_h(x,a)}$.
    Thus, $E_5$ (\cref{eq:goodTransitionPOuniform}) is given by the triangle inequality together with
    \begin{align*}
        \norm{\widehat{P}_h^{\kEpoch}(\cdot \mid x,a) - \Bar{P}_h^{\kEpoch}(\cdot \mid x,a)}_1
        \le
        \frac{1}{n^{\kEpoch}_h(x,a) + 1}
        .
    \end{align*}

    Next, to prove that $E_2$ (\cref{eq:goodTransitionPO}) holds, first use
    \cref{lemma:dynamics-error-set-v} with a set of value functions that contains only a single value function $V^\star$ to get that with probability at least $1 - \delta / 7$
    \begin{align*}
        \norm{(\psiH - \psiKH){V}_{h+1}^{\star}}_{\hatCovKH} 
        \le
        \betaP
        ,
        \quad
        \forall e \in \setEpochs, k \in \setEpochsE , h \in [H]
        .
    \end{align*}
    Now, suppose that $E_1, E_3$ \cref{eq:goodRewardPO,eq:goodZetaConcentrationPO} hold. Then for all $e \in \setEpochs, k \in \setEpochsE, h \in [H], i \in [m], a \in \A, x \in \X$:
    \begin{align*}
        \abs{\hat{r}^{k,i}_h(x,a)}
        =
        \norm{\thetaKH + \zh^{\kEpoch,i}}
        \tag{triangle inequality, $\norm{\phi(x,a)} \le 1$}
        &
        \le
        \abs{\phi(x,a)\tran\thetaH} + \norm{\thetaK - \theta} + \norm{\zr^{\kEpoch,i}} + \norm{\zph^{\kEpoch,i}}
        \\
        \tag{$r_h(x,a) \in [0,1], \covK,\hatCovK \succeq H I, \hatCovKH \succeq I$}
        &
        \le
        1
        +
        \frac{
        \norm{\thetaK - \theta}_{\covK} + \norm{\zr^{\kEpoch,i}}_{\hatCovK}}
        {\sqrt{H}} 
        +
        \norm{\zp^{\kEpoch,i}}_{\hatCovKH}
        \\
        \tag{\cref{eq:goodRewardPO,eq:goodZetaConcentrationPO}}
        &
        \le
        1
        +
        \betaR(1+\betaZ) / \sqrt{H}
        +
        \betaP \betaZ
        \\
        \tag{$\betaP \ge 2 \betaR / \sqrt{H}, \betaZ \ge 3, \betaP \ge 4$}
        &
        \le
        2\betaP \betaZ
        .
    \end{align*}
    Notice that $\hat{Q}^{k,i}_h(x,a) = \EE[\widehat{P}^k, \pi^{k,i}] \brk[s]*{\sum_{h'=h}^{H} \hat{r}^{k,i}_{h'}(x_{h'},a_{h'}) \mid x_h = x, a_h = a}$. Since $\widehat{P}^k$ is a sub-probability measure, we conclude that
    \begin{align*}
        \norm{\hat{Q}^{k,i}_h}_{\infty}
        \le
        \sum_{h'=h}^{H} \max_{x \in \X, a \in \A} \abs{\hat{r}^{k,i}_h(x,a)}
        \le
        2 H \betaP \betaZ
        =
        \betaQ
        ,
    \end{align*}
    establishing $E_2$ (\cref{eq:goodTransitionPO}).

    Next, notice that $\norm{\phi^k}_{\covK^{-1}}, \norm{\phi_h^k}_{\hatCovKH^{-1}} \le 1$, thus 
    $0 \le Y_k \le \betaR (1 + \sqrt{2}\betaZ) + \sqrt{2}H(\betaP \betaZ + \betaQ \betaPtag)$. Using Azuma's inequality, we conclude that $E_6$ (\cref{eq:goodBernsteinPO}) holds with probability at least $1-\delta/7$.
    %
    %
    Finally, we use Azuma's inequality to get that with probability at least $1-\delta/7$
    \begin{align*}
        \sum_{k \in [K]} \sum_{i \in [m]} p^k(i) \clip_{\betaQ}\brk[s]*{\hat{V}^{k,i}_1(x_1)}
        \le
        \sum_{k \in [K]} \clip_{\betaQ}\brk[s]*{\hat{V}^{k,i_k}_1(x_1)} + 2 \betaQ  \sqrt{2K \log\frac{7}{\delta}}
        .
    \end{align*}
    Since $\norm{\hat{Q}^{k,i}_h}_{\infty} \le \betaQ$, we conclude that $\clip_{\betaQ}\brk[s]*{\hat{V}^{k,i}_1(x_1)} = \hat{V}^{k,i}_1(x_1)$, establishing $E_7$ (\cref{eq:goodAzumaPO}).

    Taking a union bound, all of the events so far hold with probability at least $1-\delta$.
\end{proof}

\newpage
\section{Technical tools}
\label{sec:technical}

\subsection{Online Mirror Descent}
We begin with a standard regret bound for entropy regularized online mirror descent (hedge). See \citet[Lemma 25]{sherman2023rate}.

\begin{lemma}
\label{lem:omd-regret}
    Let $y_1, \dots, y_T \in \RR[A]$ be any sequence of vectors, and $\eta > 0$ such that $\eta y_t(a) \ge -1$ for all $t \in [T],a \in [A]$. Then if $x_t \in \Delta_A$ is given by $x_1(a) = 1/A \ \forall a$, and for $t \ge 1$:
    \begin{align*}
        x_{t+1}(a)
        =
        \frac{x_t(a) e^{- \eta y_t(a)}}{\sum_{a' \in [A]} x_t(a') e^{- \eta y_t(a')}},
    \end{align*}
    then,
    \begin{align*}
        \max_{x \in \Delta_A} \sum_{t=1}^T \sum_{a \in [A]} y_t(a) \brk*{x_t(a) - x(a)}
        \le
        \frac{\log A}{\eta}
        +
        \eta \sum_{t=1}^T \sum_{a \in [A]} x_t(a) y_t(a)^2.
    \end{align*}
\end{lemma}

\subsection{Value difference lemma}
We use the following extended value difference lemma by \citet{shani2020optimistic}. We note that the lemma holds unchanged even for MDP-like structures where the transition kernel $P$ is a sub-stochastic transition kernel, i.e., one with non-negative values that sum to at most one (instead of exactly one).

\begin{lemma}[Extended Value difference Lemma 1 in \cite{shani2020optimistic}]
\label{lemma:extended-value-difference}
    Let $\M$ be an MDP, $\pi, \hat{\pi} \in \piClass$ be two policies, $\hat{Q}_h: \X \times \A \to \RR, h \in [H]$ be arbitrary function, and $\hat{V}_h : \X \to \RR$ be defined as
    $\hat{V}_h(x) = \sum_{a \in \A} \hat{\pi}_h(a \mid x) \hat{Q}_h(x, a).$ Then
    \begin{align*}
        V^\pi_1(x_1) - \hat{V}_1(x_1)
        &
        =
        \EE[P,\pi]
        \brk[s]*{
        \sum_{h \in [H]}\sum_{a \in \A} \hat{Q}_h(x_h,a) (\pi(a \mid x_h) - \hat{\pi}(a \mid x_h))
        }
        \\
        &
        +
        \EE[P,\pi]
        \brk[s]*{
        \sum_{h \in [H]} r_h(x_h, a_h) + \sum_{x' \in \X} P(x' \mid x_h, a_h) \hat{V}_{h+1}(x') - \hat{Q}_h(x_h, a_h)
        }
        .
    \end{align*}
    We note that, in the context of linear MDP $r_h(x_h, a_h) + \sum_{x' \in \X} P(x' \mid x_h, a_h) \hat{V}_{h+1}(x') = \phi(x_h, a_h)\tran(\thetaH + \psiH \hat{V}_{h+1}).$
\end{lemma}

\subsection{Algebraic lemmas}

Next, is a well-known bound on harmonic sums \citep[see, e.g.,][Lemma 13]{cohen2019learning}. This is used to show that the optimistic and true losses are close on the realized predictions. See proof below for completeness.
\begin{lemma}
\label{lemma:elliptical-potential}
    Let $z_t \in \RR[d']$ be a sequence such that $\norm{z_t}^2 \le \lambda$, and define $V_t = \lambda I + \sum_{s=1}^{t-1} z_s z_s\tran$. Then
    \begin{align*}
        \sum_{t=1}^{T} \norm{z_t}_{V_t^{-1}}
        \le
        \sqrt{T \sum_{t=1}^{T} \norm{z_t}_{V_t^{-1}}^{2}}
        \le
        \sqrt{2 T d' \log (T+1)}
        .
    \end{align*}
\end{lemma}
\begin{proof}
    Notice that $0 \le z_t\tran V_t^{-1} z_t \le \norm{z_t}^2 / \lambda \le 1$.
    Next, notice that 
    \begin{align*}
        \det(V_{t+1})
        =
        \det(V_t + z_t z_t\tran)
        =
        \det(V_t)\det(I + V_t^{-1/2} z_t z_t\tran V_t^{-1/2})
        =
        \det(V_t) (1 + z_t\tran V_t^{-1} z_t)
        ,
    \end{align*}
    which follows from the matrix determinant lemma. We thus have that
    \begin{align*}
        \tag{$x \le 2 \log (1+x), \forall x \in [0,1]$}
        z_t\tran V_t^{-1} z_t
        \le
        \log (1 + z_t\tran V_t^{-1} z_t)
        =
        \log \frac{\det(V_{t+1})}{\det(V_t)}
    \end{align*}
    We conclude that
    \begin{align*}
        \sum_{t=1}^{T} z_t\tran V_t^{-1} z_t
        &
        \le
        2 \sum_{t=1}^{T} \log \brk*{\det(V_{t+1}) / \det(V_t)}
        \\
        \tag{telescoping sum}
        &
        =
        2 \log \brk*{\det(V_{T+1}) / \det(\lambda I)}
        \\
        \tag{$\det(V) \le \norm{V}^{d'}$}
        &
        \le
        2 d' \log (\norm{V_{T+1}} / \lambda)
        \\
        \tag{triangle inequality}
        &
        \le
        2 d' \log \brk*{1 + \sum_{t=1}^{T} \norm{z_t}^2 / \lambda}
        \\
        &
        \le
        2 d' \log (T+1)
        .
    \end{align*}
    The proof is concluded by applying the Cauchy-Schwarz inequality to get that
    $\sum_{t=1}^{T} \norm{z_t}_{V_t^{-1}}
        \le
        \sqrt{T \sum_{t=1}^{T} \norm{z_t}_{V_t^{-1}}^{2}}
        .
    $
\end{proof}

Next, we need the following well-known matrix inequality.
\begin{lemma}[\cite{cohen2019learning}, Lemma 27]
\label{lemma:matrix-norm-inequality}
    If $N \succeq M \succ 0$ then for any vector $v$
    \begin{align*}
        \norm{v}_{N}^2
        \le
        \frac{\det{N}}{\det{M}} \norm{v}_{M}^2
    \end{align*}
\end{lemma}

\subsection{Concentration and anti-concentration bounds}
Next, we have an anti-concentration for standard Gaussian random variables.
\begin{lemma}
\label{lemma:gaussian-anti-concentration}
    Let $\sigma, m \ge 0$. Suppose that $g_i \sim \mathcal{N}(0, \sigma^2)$, $i\in[m]$ are i.i.d Gaussian random variables. With probability at least $1 - e^{-m/9}$
    \begin{align*}
        \max_{i \in [m]} g_i
        \ge
        \sigma
        .
    \end{align*}
\end{lemma}
\begin{proof}
    Recall that for a standard Gaussian random variable $G \sim \mathcal{N}(0,1)$ we have that $\Pr[G \ge 1] \ge 1/9$.
    Since $g_i$ are independent, we conclude that
    \begin{align*}
        \Pr[
        \max_{i \in [m]} g_i
        \le
        \sigma
        ]
        =
        \Pr[G \le 1]^m
        \le
        \brk*{8/9}^m
        \le
        e^{-m/9}
        ,
    \end{align*}
    and taking the complement of this event concludes the proof.
\end{proof}

Next, a tail inequality on the norm of a standard Gaussian random vector
\begin{lemma}
\label{lemma:gaussian-norm-concentration}
    Let $g \sim \mathcal{N}(0,I)$ be a $d$ dimensional Gaussian random vector. With probability at least $1-\delta$
    \begin{align*}
        \norm{g} 
        \le
        \sqrt{\frac{3d}{2} + 4 \log \frac{1}{\delta}}
        \le
        \sqrt{\frac{11d}{2} \log \frac{1}{\delta}}
        .
    \end{align*}
\end{lemma}

Next, we state a standard concentration inequality for a self-normalized processes.
\begin{lemma}[Concentration of Self-Normalized Processes \cite{abbasi2011improved}]
\label{lemma:self-normalized}
    Let $\eta_t$ ($t \ge 1$) be a real-valued stochastic process with corresponding filtration $\mathcal{F}_t$. Suppose that $\eta_t \mid \mathcal{F}_{t-1}$ are zero-mean $R$-subGaussian, and let $\phi_t$ ($t \ge 1$) be an $\RR[d]$-valued, $\mathcal{F}_{t-1}$-measurable stochastic process. Assume that $\Lambda_0$ is a $d \times d$ positive definite matrix and let $\Lambda_t = \Lambda_0 + \sum_{s=1}^{t} \phi_s \phi_s\tran$. Then for any $\delta > 0$, with probability at least $1 - \delta$, we have for all $t \ge 0$
    \begin{align*}
        \norm*{
        \sum_{s=1}^{t} \phi_s \eta_s
        }_{\Lambda_t^{-1}}^2
        \le
        2 R^2 \log \brk[s]*{\frac{\det\brk{\Lambda_t}^{1/2} \det\brk{\Lambda_0}^{-1/2}}{\delta}}
    \end{align*}
    Additionally, if $\Lambda_0 = \lambda I$ and $\norm{\phi_t}^2 \le \lambda$, for all $t \ge 1$ then
    \begin{align*}
        \norm*{
        \sum_{s=1}^{t} \phi_s \eta_s
        }_{\Lambda_t^{-1}}^2
        \le
        R^2 \brk[s]{d \log (t+1) + 2\log (1/\delta)}
    \end{align*}
\end{lemma}

\subsection{Reward and dynamics estimation bounds}

We derive the standard guarantee for the rewards least-squares estimate.
\begin{lemma}[reward error bound]
\label{lemma:reward-error}
    Let $\thetaK$ be as in \cref{alg:RE-LSVI} and suppose that $\lambdaR = H$. With probability at least $1 - \delta$, for all $k \ge 1$
    \begin{align*}
        \norm{\theta - \thetaK}_{\covK}
        \le
        2 H \sqrt{2dH \log (2K/\delta)}
    \end{align*}
\end{lemma}
\begin{proof}
    We write $v^\tau = (\phi^\tau)\tran \theta + \eta_\tau$ where 
    $
        \eta_\tau
        =
        \sum_{h \in [H]} r_h^k - r_h(x_h^k, a_h^k)
        .
    $
    Notice that $\abs{\eta_\tau} \le H$, making it $H$-subGaussian, which satisfies the conditions of \cref{lemma:self-normalized} with $R = H$. Next, notice that
    \begin{align*}
        \thetaK
        =
        \covK^{-1} \sum_{\tau=1}^{k-1} \phi^\tau v^\tau
        =
        \covK^{-1} \sum_{\tau=1}^{k-1} \phi^\tau \brk[s]{(\phi^\tau)\tran \theta + \eta_\tau}
        =
        \theta 
        -
        \lambdaR \covK^{-1} \theta 
        +
        \covK^{-1} \sum_{\tau=1}^{k-1} \phi^\tau  \eta_\tau
    \end{align*}
    We conclude that with probability at least $1-\delta/5$, for all $k \ge 1$
    \begin{align*}
        \norm*{\theta - \thetaK}_{\covK}
        &
        \le
        \norm*{\lambdaR (\covK)^{-1} \theta}_{\covK}
        +
        \norm*{(\covK)^{-1} \sum_{\tau=1}^{k-1} \phi^\tau  \eta_\tau}_{\covK}
        \\
        &
        =
        \lambdaR \norm*{\theta}_{(\covK)^{-1}}
        +
        \norm*{\sum_{\tau=1}^{k-1} \phi^\tau  \eta_\tau}_{(\covK)^{-1}}
        \\
        \tag{\cref{lemma:self-normalized}}
        &
        \le
        \lambdaR \norm*{\theta}_{(\covK)^{-1}}
        +
        H \sqrt{dH \log(K+1) + 2 \log (1/\delta)}
        \\
        \tag{$\covK \succeq \lambdaR I$}
        &
        \le
        \sqrt{\lambdaR} \norm*{\theta}
        +
        H \sqrt{dH \log(K+1) + 2 \log (1/\delta)}
        \\
        \tag{$\norm{\theta} \le \sqrt{dH}, \lambdaR = H$}
        &
        \le
        H \sqrt{d}
        +
        H \sqrt{dH \log(K+1) + 2 \log (1/\delta)}
        \\
        &
        \le
        2 H \sqrt{dH \log(K+1) + 2 \log (1/\delta)}
        \\
        &
        \le
        2 H \sqrt{2dH \log (2K/\delta)}
        . \qedhere
    \end{align*}
\end{proof}

Next, we derive a standard error bound for the dynamics approximation.
\begin{lemma}[dynamics error uniform convergence]
\label{lemma:dynamics-error-set-v}
    Suppose that $\lambdaP = 1$.
    For all $h \in [H]$, let $\mathcal{V}_h \subseteq \RR[\X]$ be a set of mappings $V: \X \to \RR$ such that $\norm{V}_\infty \le \beta$ and $\beta \ge 1$. Let $\psiKH : \RR[\X] \to \RR[d]$ be the linear operator defined as
    \begin{align*}
        \psiKH V
        =
        (\covKH)^{-1} \sum_{\tau=1}^{k-1} \phi^\tau_h V(x_{h+1}^\tau)
        .
    \end{align*}
    With probability at least $1-\delta$, for all $h \in [H]$, $V \in \mathcal{V}_{h+1}$ and $k \ge 1$
    \begin{align*}
        \norm{(\psiH - \psiKH)V}_{\covKH}
        \le
        4 \beta \sqrt{d \log (K+1) + 2\log (H \mathcal{N}_\epsilon /\delta)}
        ,
    \end{align*} 
    where $\epsilon \le \beta \sqrt{d}/{2K}$, $\mathcal{N}_\epsilon = \sum_{h \in [H]} \mathcal{N}_{h,\epsilon}$, and $\mathcal{N}_{h,\epsilon}$ is the $\epsilon-$covering number of $\mathcal{V}_h$ with respect to the supremum distance.
\end{lemma}
\begin{proof}
    %
    For any $h \in [H]$ let $\mathcal{V}_{h,\epsilon}$ be a minimal $\epsilon-$cover for $\mathcal{V}_h$. 
    Next, for any $h \in [H]$ and $V \in \mathcal{V}_{h+1}$ define the linear map $\eta_{\tau, h}(V) = V(x_{h+1}^\tau) - (\phi_h^\tau)\tran \psi_h V$. 
    Notice that $\abs{\eta_{\tau, h}(V)} \le 2 \norm{V}_\infty \le 2 \beta$, thus $\eta_{\tau, h}(V)$ satisfies \cref{lemma:self-normalized} with $R = 2 \beta$.  Taking a union bound, we conclude that with probability at least $1 - \delta$
    \begin{align}
    \label{eq:psi-concentration}
        \norm*{\sum_{\tau=1}^{k-1} \phi^\tau_h  \eta_{\tau,h}(\tilde{V})}_{(\covKH)^{-1}}
        \le
        2 \beta \sqrt{d \log (K+1) + 2\log (H \mathcal{N}_\epsilon /\delta)}
        \quad
        ,
        \forall k \in [K], h \in [H], \tilde{V} \in \mathcal{V}_{h+1,\epsilon}
        .
    \end{align}
    We assume that the above event holds for the remainder of the proof.
    Now, notice that
    \begin{align*}
        \psiKH V
        =
        (\covKH)^{-1} \sum_{\tau=1}^{k-1} \phi_h^\tau V(x_{h+1}^\tau)
        =
        \psiH V
        -
        \lambdaP (\covKH)^{-1} \psiH V
        +
        (\covKH)^{-1} \sum_{\tau=1}^{k-1} \phi_h^\tau \eta_{\tau, h}(V)
        .
    \end{align*}
    We are now ready to finish the proof.
    Let $k \in [K], h \in [H], V \in \mathcal{V}_{h+1}$ and let $\tilde{V} \in \mathcal{V}_{h+1,\epsilon}$ be the point in the cover corresponding to $V$. Denoting their residual as $\Delta_V = V - \tilde{V}$ we have that
    \begin{align*}
        \norm*{(\psiH - \psiKH) V}_{\covKH}
        &
        \le
        \norm*{\lambdaP (\covKH)^{-1} \psiH V}_{\covKH}
        +
        \norm*{(\covKH)^{-1} \sum_{\tau=1}^{k-1} \phi^\tau_h  \eta_{\tau,h}(V)}_{\covKH}
        \\
        &
        =
        \lambdaP \norm*{\psiH V}_{(\covKH)^{-1}}
        +
        \norm*{\sum_{\tau=1}^{k-1} \phi^\tau_h  \eta_{\tau,h}(V)}_{(\covKH)^{-1}}
        \\
        &
        \le
        \lambdaP \norm*{\psiH V}_{(\covKH)^{-1}}
        +
        \norm*{\sum_{\tau=1}^{k-1} \phi^\tau_h  \eta_{\tau,h}(\tilde{V})}_{(\covKH)^{-1}}
        +
        \norm*{\sum_{\tau=1}^{k-1} \phi^\tau_h  \eta_{\tau,h}(\Delta_V)}_{(\covKH)^{-1}}
        \\
        \tag{$\eta_{\tau,h}(\Delta_V) \le 2 \norm{\Delta_V}_\infty \le 2\epsilon$}
        &
        \le
        \lambdaP \norm*{\psiH V}_{(\covKH)^{-1}}
        +
        \norm*{\sum_{\tau=1}^{k-1} \phi^\tau_h  \eta_{\tau,h}(\tilde{V})}_{(\covKH)^{-1}}
        +
        2 k \epsilon
        \\
        \tag{$\covKH \succeq \lambdaP I,$ Cauchy-Schwarz}
        &
        \le
        \sqrt{\lambdaP} \norm*{\abs{\psiH}(\X)} \norm{V}_\infty
        +
        \norm*{\sum_{\tau=1}^{k-1} \phi^\tau_h  \eta_{\tau,h}(\tilde{V})}_{(\covKH)^{-1}}
        +
        2 k \epsilon
        \\
        \tag{$\norm{V}_\infty \le \beta, \norm{\abs{\psiH}(\X)} \le \sqrt{d}$}
        &
        \le
        \beta \sqrt{d \lambdaP}
        +
        \norm*{\sum_{\tau=1}^{k-1} \phi^\tau_h  \eta_{\tau,h}(\tilde{V})}_{(\covKH)^{-1}}
        +
        2 k \epsilon
        \\
        \tag{\cref{eq:psi-concentration}}
        &
        \le
        \beta \sqrt{d \lambdaP}
        +
        2 \beta \sqrt{d \log (K+1) + 2\log (H \mathcal{N}_\epsilon /\delta)}
        +
        2 k \epsilon
        \\
        \tag{$\epsilon \le {\beta \sqrt{d}}/{2 k}, \lambdaP = 1$}
        &
        \le
        4 \beta \sqrt{d \log (K+1) + 2\log (H \mathcal{N}_\epsilon /\delta)}
        ,
    \end{align*}
    thus concluding the proof.
\end{proof}

\subsection{Covering numbers}

The following results are (mostly) standard bounds on the covering number of several function classes.

\begin{lemma}
\label{lemma:ball-cover}
    For any $\epsilon > 0$, the $\epsilon$-covering of the Euclidean ball in $\RR[d]$ with radius $R \ge 0$ is upper bounded by $(1 + 2R/\epsilon)^d$.
\end{lemma}

\begin{lemma}[Lemma D.6 of \cite{jin2020provably}]
\label{lemma:covering-number}
    Let $\mathcal{V}$ denote a class of functions $V : \mathcal{X} \to \RR$ with the following parametric form
    \begin{align*}
        V(\cdot)
        =
        \clip_R\brk[s]*{\max_a \theta\tran \phi(\cdot,a) + \beta \sqrt{\phi(\cdot, a)\tran \Lambda^{-1} \phi(\cdot, a)}}
        ,
    \end{align*}
    where $R \ge 0$ is a constant and $(\theta,\beta,\Lambda, R)$ are parameters satisfying $\norm{\theta} \le L, \beta \in [0, B], \lambda_{\min}(\Lambda) \ge \lambda$ where $\lambda_{\min}$ denotes the minimal eigenvalue. Assume $\norm{\phi(x,a)} \le 1$ for all $(x,a)$ pairs, and let $\mathcal{N}_\epsilon$ be the $\epsilon-$covering number of $\mathcal{V}$ with respect to the supremum distance. Then
    \begin{align*}
        \log \mathcal{N}_\epsilon
        \le
        d \log(1 + 4 L / \epsilon)
        +
        d^2 \log \brk[s]{1 + 8 \sqrt{d} B^2 / (\lambda \epsilon^2)}
        .
    \end{align*}
\end{lemma}

\begin{lemma}
\label{lemma:lipschitz-cover}
    Let $\mathcal{V} = \brk[c]{V(\cdot; \theta) : \norm{\theta} \le W}$ denote a class of functions $V : \mathcal{X} \to \RR$. Suppose that any $V \in \mathcal{V}$ is $L$-Lipschitz with respect to $\theta$ and supremum distance, i.e.,
    \begin{align*}
        \norm{V(\cdot; \theta_1) - V(\cdot; \theta_2)}_\infty
        \le
        L \norm{\theta_1 - \theta_2}
        ,
        \quad
        \norm{\theta_1}, \norm{\theta_2} \le W
        .
    \end{align*}
    Let $\mathcal{N}_\epsilon$ be the $\epsilon-$covering number of $\mathcal{V}$ with respect to the supremum distance. Then
    \begin{align*}
        \log \mathcal{N}_\epsilon
        \le
        d \log(1 + 2 W L / \epsilon)
    \end{align*}
\end{lemma}
\begin{proof}
    Let $\Theta_{\epsilon/L}$ be an $(\epsilon/L)$-covering of the Euclidean ball in $\RR[d]$ with radius $W$. Define $\mathcal{V}_\epsilon = \brk[c]{V(\cdot; \theta) : \theta \in \Theta_{\epsilon / L}}$.
    By \cref{lemma:ball-cover} we have that $\log\abs{\mathcal{V}_\epsilon} \le d \log(1 + 2 W L / \epsilon)$. We show that $\mathcal{V}_\epsilon$ is an $\epsilon$-cover of $\mathcal{V}_{\epsilon}$, thus concluding the proof.
    Let $V \in \mathcal{V}$ and $\theta$ be its associated parameter. Let $\theta' \in \Theta_{\epsilon / L}$ be the point in the cover nearest to $\theta$ and $V' \in \mathcal{V}$ its associated function. Then we have that 
    \begin{align*}
        \norm{V(\cdot) - V'(\cdot)}_\infty
        =
        \norm{V(\cdot; \theta) - V(\cdot; \theta')}_\infty
        \le
        L \norm{\theta - \theta'}
        \le
        L (\epsilon / L) 
        = 
        \epsilon
        .
        &
        \qedhere
    \end{align*}
\end{proof}

\end{document}